\pdfoutput=1
\documentclass[11pt]{article}

\usepackage[final]{acl}

\usepackage{times}
\usepackage{latexsym}

\usepackage[T1]{fontenc}
\usepackage[utf8]{inputenc}

\usepackage{microtype}

\usepackage{inconsolata}

\usepackage{graphicx}

\usepackage{amsmath}
\usepackage{amsfonts}
\usepackage{mathtools}
\usepackage{comment}
\usepackage{enumitem}
\usepackage{subcaption}
\usepackage[most]{tcolorbox}
\usepackage{colortbl}
\usepackage{booktabs}
\usepackage{multirow}
\usepackage{rotating}
\usepackage{amssymb}

\newcommand{\diagmethod}{\texttt{DiaVLo}}
\newcommand{\sk}{\texttt{SHOULD-KNOW}}
\newcommand{\skshort}{\texttt{SK}}
\newcommand{\rk}{\texttt{REALLY-KNOW}}
\newcommand{\rkshort}{\texttt{RK}}

\title{\diagmethod{}: Diagnosing Behaviours of Vision-Language Models}

\author{
    Lorenzo Corti\thanks{~Corresponding author.}\\
    Delft University of Technology \\
    l.corti@tudelft.nl \And
    Jie Yang \\
    Delft University of Technology \\
    j.yang-3@tudelft.nl
}

\begin{document}
\maketitle
\begin{abstract}
    % To self: trying to keep the abstract concise (< 150 words).
    Vision-language models (VLMs) rely on storing and transferring appropriate information across their sub-components.
    Verifying that the VLMs exhibit desired behaviours, while avoiding harmful ones, is central to their reliable deployment.
    Yet, methods that identify VLM behaviours remain scarce.
    We present \diagmethod{}, a diagnostic framework that leverages human curation and VLMs' generation capabilities to construct specifications of desired and observed VLM behaviours, surfacing potential misalignments.
    Beyond this, \diagmethod{} also provides causal estimates to identify the most influential concepts steering VLM behaviours.
    We evaluate \diagmethod{} on several open-source VLMs under both classification and generation conditions.
    Our experiments show that \diagmethod{} produces behaviour labels that correlate with model performance and provide context for measured performance.
    \diagmethod{} surfaced behaviours that are clearly aligned and misaligned, alongside patterns in how VLMs perceive, organise, and prioritise concepts. % when responding to users' requests.
\end{abstract}

\section{Introduction}

% == == Background (Exciting & Imporant): What is the big goal here?
% Unifying data from different modalities is regarded as a prominent path in advancing artificial intelligence. Recent advances in large language models (LLMs) have demonstrated that these models can serve as general-purpose interfaces and bridge data modalities \citep{Hao2022LLMInterfaces}.
The success of current vision-language models (VLMs) hinges on the interplay of three primary modules: a visual encoder (e.g., CLIP \citep{Radford2021CLIP}), a projection middle-layer, and a decoder-only LLM (e.g., Vicuna \citep{Chiang2023Vicuna}).
The combination of these three elements, either composite \citep{liu2024llavanext} or trained end-to-end \citep{Bai2025Qwen25VL}, has proven effective across multimodal tasks in both offline benchmarks and downstream applications \citep{Hao2022LLMInterfaces, Li2025VLMSurvey}
% LLMs, in particular, have boosted the capabilities of large visual encoders, enabling systems that can tackle multimodal tasks effectively \citep{Hao2022LLMInterfaces}.
% This integrative approach has proven effective across both offline benchmarks and general downstream applications \citep{Li2025VLMSurvey}.
% However, because of this integrative approach to model building, pinpointing how and where visual and language information is stored and transferred remains an open question \citep{Samyadeep2024TransferMultimodal}.
% Most of VLMs' capabilities seem to stem from the knowledge LLMs encode rather than from visual perception.
% Prior work, specifically, found discrepancies between the performance of VLMs and the visual encoders they use \citep{Fu2025VLMFailures} and highlighted critical blind spots of the latter \citep{Tong2024EyesWideShut}.
%
However, VLMs' performance appears to stem from the language module, as recent work has uncovered critical blind spots in visual perception \citep{Tong2024EyesWideShut} and performance discrepancies on vision-centric tasks \citep{Fu2025VLMFailures}.
These findings point to unpredictable, under-specified VLM behaviours that hinder our understanding of these models and their broader applicability across domains \citep{Shu2025VLMExpl}.
% These findings highlight reliability issues (e.g., hallucinations
% % \citep{Ji2023HallucinationsSurvey}
% and brittleness to prompt perturbations
% % \citep{Zhu2023PromptBench}, and real-world harms \citep{Hamidieh2024BiasVLM}
% ) that limit the application of VLMs to domains with stricter acceptability criteria.

\begin{figure}[t]
  \includegraphics[width=0.95\columnwidth]{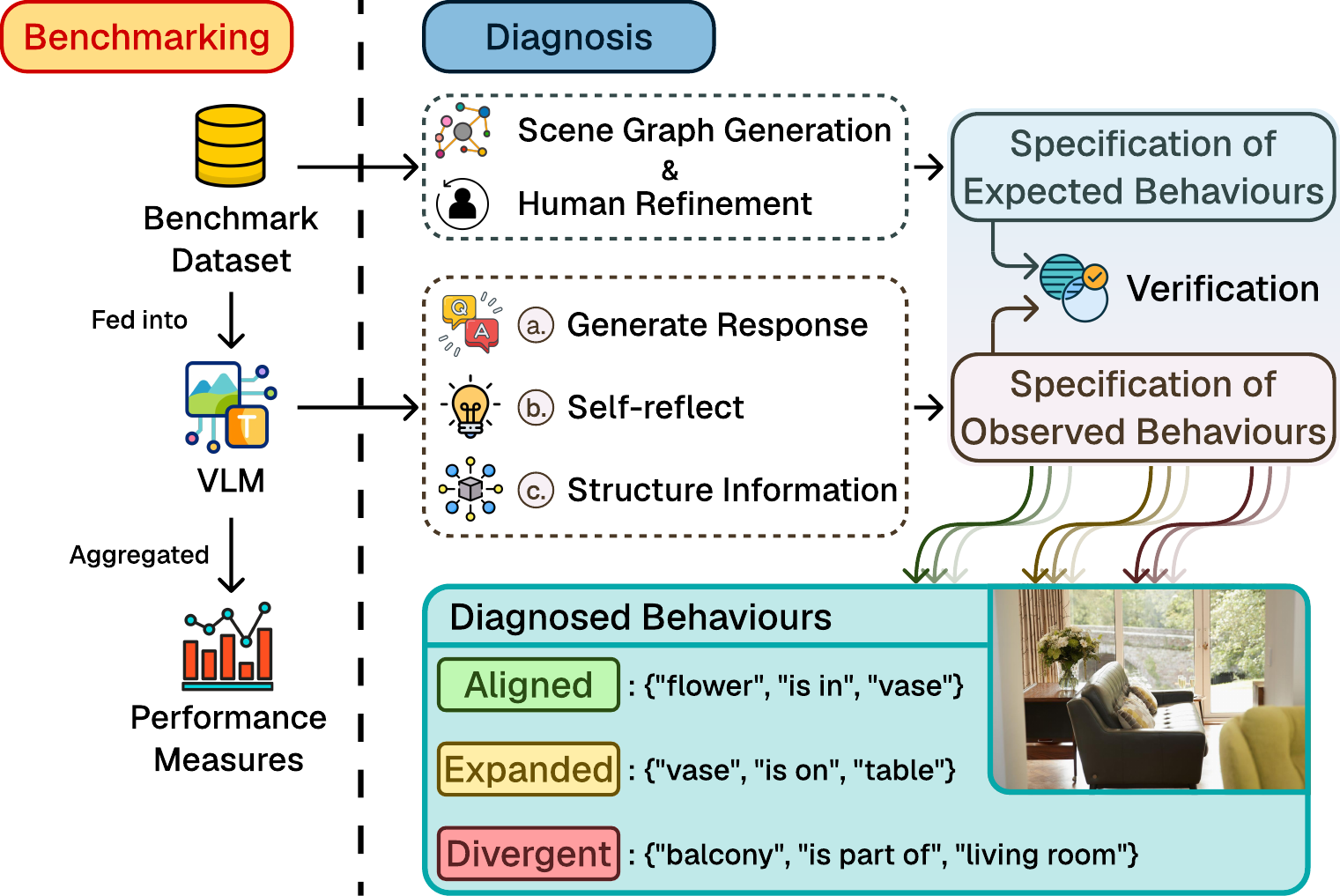}
  \caption{With \diagmethod{}, we frame VLM diagnosis as an addition to benchmarking, enabling the definition and verification of observed model behaviours with respect to specifications of desired behaviours.
  % We first compile model-agnostic specifications of desired behaviours (1). Then, we compile specifications of observed model behaviours for individual model outputs (2). Finally, faulty and benign behaviours are identified and characterised by estimating the causal effects of concepts that influenced the model (3).
  }
  \label{fig:teaser}
\end{figure}

% == == Complication (Deep)
In this work, we address the problem of \textit{diagnosing VLM behaviours}, i.e., verifying that these models exhibit the desired output regularities while avoiding unwanted ones \citep{Holtzman2025ModelBehaviours, Zhang2022MLTesting, Cabrera2023AIDescriptions}. % is key to their reliable deployment.
% We hypothesise this could also inform which data to use to improve VLMs, which otherwise require exponential amounts of data to achieve linear gains \citep{Udandarao2024ZeroShot}.
Prior work sought to analyse VLMs through \textit{benchmarking} and \textit{interpretability}.
% these studies do not provide \textit{desired model behaviours} but instead rely on proxies that do not naturally map to real-world settings where humans might be actively using VLMs.
On the one hand, benchmarks provide aggregated assessments of VLMs' task performance, e.g., using embedding-based metrics \citep{Zhang2020BERTScore} or end-to-end ones \citep{Liu2023GEval}.
% On the one hand, benchmarks provide only an aggregated, high-level overview of VLMs by using established metrics such as embedding-based ones \citep{Zhang2020BERTScore} or end-to-end \citep{Liu2023GEval} ones.
Here, VLMs are commonly evaluated on diverse, human-interpretable capabilities such as visual perception and reasoning.
For instance, SEED-Bench-2, organises 24k datapoints across 27 dimensions \citep{Li2023SEEDBench}.
On the other hand, interpretability methods (e.g., \citet{Achtibat2024LXT, Parcalabescu2024Selfconsistency}) describe salient visual and text features that VLMs may use to produce a response.
These are generally classified as black-box methods (i.e., requiring query access to the model) or white-box methods (i.e., requiring full model access) \citep{Resck2025MLLMExpl}.
Yet, these crucial works often rely on proxies that do not naturally map to real-world settings where humans might be actively using VLMs \citep{Xiao2023EvaluatingNLG, Sokol2020ExplFactSheets}.
% descriptions of VLM outputs that are often hard to interpret without a specification of desired VLM behaviours, despite striving to aid human decision-making \citep{Guidotti2022Counterfactuals, Sokol2020ExplFactSheets}.

%For instance, methods relying on model gradients \citep{Sundararajan2017IntegratedGradients, Achtibat2024LXT} and input perturbations \citep{Parcalabescu2024Selfconsistency} produce token-level attributions that are virtually equal without specifying the desired VLM behaviours.

% However, diagnosing model behaviour is challenging as (1) it encompasses practitioners' heuristic insights and learning processes \citep{Araki1991GeneralDebugging, Lertvittayakumjorn2021HumanXAINLPSurvey}; and (2) the definition of ``failure case'' is context-dependent and assumes the existence of specifications of desired model behaviours.\footnote{``Failure'': when desired and observed model behaviours differ. ``Error'': when model outputs and ground truth differ.}

% Yet, existing approaches \citep{Zhao2024XAILLMs} often generate overly simplistic representations of model behaviour and neglect the (possible) causal relationships between input features and model outputs \citep{Guidotti2022Counterfactuals}. While both perspectives shed light on model behaviours in different ways, their limitations undermine our ability to adequately \textit{characterise} and \textit{foresee} real-world model failures.

% == == Solution (New)
Given these limitations, we introduce \diagmethod{}, a diagnostic framework 
enabling the definition of desired behaviours and the extraction and classification of observed VLM behaviours in response to the former (\autoref{fig:teaser}).
% with three main steps: defining desired behaviours, extracting observed behaviours, and classifying observed behaviours in response to the desired ones.
% Such behaviours are defined at the conceptual level, where real-world entities are put into relationships, e.g., \texttt{(dog, near, tree)}.
First, desired behaviours are defined by creating symbolic representations of visual inputs and having humans verify their correctness and relevance to the task.
Second, VLMs are prompted to verbalise the structured rationales underlying their responses, thereby producing self-explanatory representations of their own behaviours.
Finally, desired and observed behaviours are semantically compared to identify behavioural alignment.
In \diagmethod{}, we leverage causal modelling to quantify the influence of visual concepts on VLM outputs and mitigate potential inconsistencies in self-explained behaviours.
For this, we synthetically generate counterfactual observed behaviours and estimate concept-level effects with a Double Machine Learning estimator \citep{Chernozhukov2018DoubleML}.

% By drawing from literature in software engineering and explanation-based debugging \citep{Araki1991GeneralDebugging, Lertvittayakumjorn2021HumanXAINLPSurvey}, we formalise the diagnosis problem as the verification of observed model behaviours against a set of model-agnostic specifications of desired behaviours \ju{I would explain this early}.

We evaluate \diagmethod{} across four open-source VLMs and four public visual question answering datasets, in both classification and generation settings. 
Our results show that the behaviour labels assigned by \diagmethod{} correlate with measured model performance, indicating that \diagmethod{} can provide context to standard evaluation metrics.
Additionally, \diagmethod{} surfaced behaviours that are clearly aligned and misaligned, while allowing us to uncover patterns related to (1) how VLMs perceive concepts (through hypernyms and hyponyms), (2) differences compared to humans in relating concepts together, and (3) prioritisation of certain concepts when responding to users' requests.
% Additionally, beyond behaviours that are clearly aligned or misaligned with the desired ones, \diagmethod{} allowed us to uncover patterns in how the VLMs tested perceive and organise visual concepts that differ from those of humans.
% Finally, the concept-level attributions produced by \diagmethod{} indicate that VLMs might perceive but not attend strongly to certain concepts depending on the request.
% 
In summary, we contribute: a conceptualisation of the problem of diagnosing vision-language models; \diagmethod{}, a framework for surfacing and classifying the desired and observed VLM behaviours; and an extensive analysis of the behaviours of four well-known VLMs surfaced with \diagmethod{}.

% \begin{itemize}[leftmargin=*]
%     \item Conceptualisation of the problem of diagnosing vision-language models;
%     \item \diagmethod{}, a framework for surfacing and classifying the desired and observed VLM behaviours;
%     \item Extensive analysis of the behaviours of four well-known VLMs surfaced with \diagmethod{}.
% \end{itemize}

\section{Related Work}
% \ju{remove subsection and just highlight it to use less space.}
\subsection{Vision-Language Models} \label{sec:related_VLMs}
% Crucially, \citet{Hao2022LLMInterfaces} showed LLM can be used as general-purpose interfaces to handle diverse tasks and data modalities.
Most of the current VLMs are image-and-text-to-text models that combine a visual encoder, a connection module between visual and text modalities, and a decoder-only LLM.
% Oftentimes, the visual encoder is frozen, the connection module is trained (independent of its initialisation), and the textual decoder is fine-tuned.
For the visual encoder, CLIP \citep{Radford2021CLIP} is widely adopted \citep{Laurencon2024BetterVLM}.
As connection modules, prior work used linear layers \citep{Liu2023LLaVa}, cross-attention \citep{Alayrac2022Flamingo}, or ad-hoc methods \citep{Li2023BLIP2}.
Lastly, an LLM (e.g., Vicuna \citep{Chiang2023Vicuna}) handles text generation.
Efforts around visual instruction tuning data improved visual understanding and grounding \citep{liu2024llavanext, peng2023kosmos2}.
%rich responses \citep{chen2023sharegpt4v}

% Besides the architectural choices and benefits of using current generative LLMs \cite{Hao2022LLMInterfaces}, efforts around visual instruction tuning datasets contribute to well-performing VLMs.
% the performance of VLMs also be attributed to efforts around visual instruction tuning datasets. 
% These datasets enable improved understanding of complex prompts and images \cite{liu2024llavanext}, richer responses \cite{chen2023sharegpt4v, Li2023mimicit}, and stronger visual grounding \cite{peng2023kosmos2}.

% \subsection{Benchmarking VLMs}
\paragraph{Benchmarking VLMs}
Several benchmarks have been proposed to assess the capabilities of VLMs, 
% A wide range of benchmarks has been proposed to assess the capabilities of VLMs.
% Oftentimes, these are constructed by combining human curation and other generative models, e.g., GPT-4 \cite{OpenAIChatGPT}, to scale up the volume \cite{Liu2023LLaVa} and the variety of the data \cite{chen2023sharegpt4v}.
% This is achieved either by carefully augmenting existing datasets (e.g., MS COCO \cite{Lin2014MSCOCO}) through LLMs \cite{chen2023sharegpt4v, Liu2023LLaVa} or through human curation targeted at exploiting instruction-following capabilities of VLMs \cite{Wang2023DecodingTrust}.
% 
Several benchmarks have been proposed to assess the capabilities of VLMs, \textit{inter alia}, on visual question answering \citep{Liu2023LLaVa, Goyal2017VQAv2, Li2023SEEDBench}, low-level visual perception \citep{Wu2024QBench}, visual relation grounding \citep{Lu2025CompreCap}, and causal relevance graphs \citep{Pratama2026ViLCar}.
% Most of these focus on tasks traceable back to visual question answering (VQA), e.g., LLaVa-Bench \citep{Liu2023LLaVa}, VQA-v2 \citep{Goyal2017VQAv2}, and SEED-Bench  \citep{Li2023SEEDBench} for general capabilities; Q-Bench \citep{Wu2024QBench} for low-level visual perception, and GAIA \citep{Mialon2023GAIA} or MMBench \citep{Liu2024MMBench} for more challenging scenarios.
Despite their utility, benchmarks do not necessarily help identify or uncover specific behaviours \citep{Holtzman2025ModelBehaviours} or study information transfer across modalities \citep{Samyadeep2024TransferMultimodal}.
% \jie{would be useful if we say a bit about the relationship between the VLM architectures and the behavioural issues, and problems with benchmarking.}

\begin{comment}
% \paragraph{Issues with Benchmarking}
\noindent \textbf{Issues with Benchmarking.}
Despite the ongoing efforts around benchmarking VLMs, these and related metrics (computed in offline settings) often provide a crude impression of the reliability of VLMs and do not actively support diagnosing their behaviours. % -- neither during development nor when deployed.
Furthermore, existing metrics and human evaluation processes might introduce unwanted noise and biases \cite{Xiao2023EvaluatingNLG, Zhou2022DeconstructingNLG}.
In this work, we contribute to current evaluation practices for VLMs with a framework that produces causal characterisations of VLM behaviours (benign and faulty) for model diagnostics purposes.
% We do so by constructing specifications of expected and observed model behaviours rather than relying on aggregate metrics.
\end{comment}

\begin{comment}
    \begin{itemize}
        \item DecodingTrust \cite{Wang2023DecodingTrust} -- trust
        \item Red-Eval \cite{Bhardwaj2023RedteamingLLM} -- safety
        \item HaluEval \cite{Li2023HaluEval} -- hallucinations
        \item Do-Not-Answer \cite{Wang2023DoNotAnswer} -- safeguards
        \item PlanBench \cite{Valmeekam2023PlanBench} -- planning
        \item GPQA \cite{Rein2023GPQA} -- ``raw'' Q\&A abilities
    \end{itemize}
\end{comment}

\subsection{Explainable AI \& Causality} \label{sec:related_xai}

% \noindent \textbf{System-centred Works.}
% Explanations are viewed as means for diverse stakeholders to interpret, evaluate, or contest the output of AI systems.
% Explainable AI research is largely algorithm-centred and focuses on describing the outputs of AI systems and models \cite{BarredoArrieta2020XAISurvey, Guidotti2022Counterfactuals}.
A plethora of Explainable AI (XAI) methods have been proposed to produce \textit{local} (sample-level) or \textit{global} (class-level) output explanations, either in post-hoc (i.e., without altering a model) or self-explaining (i.e., embedded within a model) fashions \citep{Zhang2021ExPred, Resck2025MLLMExpl}.
Common XAI methods report the importance of individual input features \citep{Selvaraju2017GradCAM}, identify influential or prototypical training samples \citep{Koh2017IF, Chen2019ProtoPNet}, or produce concept-based \citep{Balayn2021SECA}, rule-based \citep{Ribeiro2018Anchor}, or counterfactual explanations \citep{Wachter2017Counterfactual}.
Notably, counterfactual explanations are the main category of explanations that incorporate causality despite calls for its broader adoption in XAI research \citep{Guidotti2022Counterfactuals}.

% However, \citet{Rodis2024MultimodalXAI} denote how only a fraction of existing XAI approaches applicable to multimodal tasks (e.g., \cite{Lyu2022DIME, Sundararajan2017IntegratedGradients}) can be used with current generative multimodal models.
%
% -- Removed the part on explanation properties. Does not add much...
% To cope with and evaluate the diverse XAI methods, prior work compiled properties of explanations \cite{carvalho2019machine, Sokol2020ExplFactSheets} which cover both system-specific aspects (e.g., fidelity) and human-specific factors (e.g., comprehensibility).

% -- Swapped Human-centred XAI works for works on self-explantions. Check previous paper version for it.
% Things to add
% - Generative agents
% - Stuff from Radek's thesis
% - Other?

\paragraph{Self-explanations}
Following advances in LLMs, natural language explanations can be easily produced alongside model outputs \citep{Huang2023LLMSelfExplain} to elicit reasoning and potentially surfacing reasoning traces \citep{Wei2022CoT}.
Recent work has found these explanations to be coherent and comparable to those from other XAI methods \citep{Huang2023LLMSelfExplain}.
In parallel, we are also starting to see LLM-specific adaptations of explanation evaluation criteria such as fidelity and self-consistency \citep{Madsen2024SelfFidelity, Parcalabescu2024Selfconsistency}.

\subsection{Model Diagnosis}
Model diagnosis is the process of investigating issues in a model and entails analysing symptoms, formulating hypotheses, and possibly reproducing the issue. % is the process of investigating ``what'' went wrong in a model.
It precedes \textit{model debugging}, which focuses on `how' to resolve issues.
Model diagnosis is not trivial, as issues can stem from the data, the training process, or the model, \textit{inter alia}.
% \footnote{We explicitly denote a difference with \textit{debugging} which concerns ``how'' to fix an existing issue.}
% Yet, it requires experience and heuristic insights to be carried out \cite{Araki1991GeneralDebugging}.
%
Prior work proposed diagnostic frameworks \citep{Ribeiro2020Checklist, Ye2024FLASK}, knowledge probes \citep{Jawahar2019BERTSyntax, May2019SEAT, Jiang2020LPAQA}, and techniques for mechanistic analysis \citep{Bricken2023MechInterpretability}.
Yet, findings from these (fundamental) works do not predictably transfer to observable model behaviours and their analysis \citep{Holtzman2025ModelBehaviours, Sharkey2025MechInterpretabilityChallenges}.
% as well as automatic techniques for analysing a model's internal states and generating synthetic data for fine-tuning \citep{Ma2018MODE, Pei2017DeepXplore}.
% \citet{Werpachowski2019OfflineTest} -- overfitting of existing models on test sets.

\paragraph{Explanation-based Approaches}
Research has also suggested principled approaches to using XAI methods to verify model outputs and identify errors \citep{Kulesza2015XAI, Caruana2015ExplReasoning, Han2021ExplErrorCauses}.
The works by \citet{Balayn2021SECA} and \citet{Sharifi2022Scalpel} are of particular interest to ours as they seek to define desirable and observable model behaviours, respectively. % to spot unknown-unknowns \cite{Attenberg2015UnknownUnknowns}.
% that disclose how their decisions are made and are receptive to corrections. These include the situatedness (i.e., awareness of a user's situation and needs) and soundness (also referred to as \textit{fidelity} \cite{carvalho2019machine}) of explanations as well as proper handling of user corrections and ensuring their reversibility.
We point to \citet{Lertvittayakumjorn2021HumanXAINLPSurvey} for a comprehensive survey of these approaches.
\section{Defining the VLM Diagnosis Problem} \label{sec:problem_statement}

We define the VLM diagnosis problem as the process of analysing the associations that VLMs make across visual and textual inputs to produce their outputs.
% Diagnosing VLMs concerns assessing whether certain operations are carried out regardless of the correctness of the output.
This requires one to construct (or access, if available) specifications of \textit{desired} and \textit{observed} behaviours.
% However, these are context-specific, and there is no agreed-upon way to obtain them.
%
Note that, while observed behaviours can be extracted in various ways (\autoref{sec:related_xai}), desired behaviours often stem from domain requirements \citep{Chen2025LawGrounding} or human expectations \citep{Lucy2024ExpectationsNLG} and may entail additional effort in formulating.
Hereafter, we use \sk{} (\skshort{}; \citet{Balayn2021SECA}) and \rk{} (\rkshort{}; \citet{Sharifi2022Scalpel}) for desired and observed behaviours, respectively.
% Hereafter, we adopt the terminology from \citet{Balayn2021SECA} and \citet{Sharifi2022Scalpel}, and refer to desired behaviours as \sk{} and to observed behaviours as \rk{}.

% \noindent\textbf{Notation.}
We denote $\mathcal{D} = \{(x_0, Y_0), \ldots, (x_{|\mathcal{D}|}, Y_{|\mathcal{D}|})\}$ as the set of a image-text pairs $x_i$ and their ground truths $Y_i$, 
%a set of prompts $\mathcal{P} = \{p_0, p_1, \ldots, p_{|\mathcal{P}|}\}$, 
and a VLM $\mathcal{M}: \mathcal{D} \rightarrow \mathcal{V}$, producing text token sequences $\hat{Y} = [\hat{y}_0, \ldots, \hat{y}_m]$ where $\hat{y}_i \in \mathcal{V}$. % and $\hat{Y} \in \hat{\mathcal{Y}}$.
% We refer to the set of token sequences $\hat{Y}$ as $\hat{\mathcal{Y}}$.
% Note that, depending on the prompt $p_i$, the generated output sequence can be a class label inferred by a model: $Y = [\hat{l}]$.
% 
% \noindent\textbf{Problem.}
To diagnose $\mathcal{M}$, we use \sk{} and \rk{} behavioural specifications $s_i = (c_i, r_{ij}, c_j) \rightarrow \hat{Y}$ which relate human-understandable concepts $c_i \in \mathcal{C}$ through $r_{ij}$ used by $\mathcal{M}$ to produce $\hat{Y}$.
We opt for a triplet-based definition to provide a structured, tractable representation of visual inputs.
In general, \skshort{} specifications are task-specific but model agnostic, while \rkshort{} specifications are specific to both.
With these, we construct the following behaviour types:
% \begin{equation}
%     \exists (c_i, \ldots, c_j) \in \mathcal{C}: (c_i, \ldots, c_j) \rightarrow \hat{Y}
% \end{equation}
\begin{itemize}[leftmargin=*]
    \item \textbf{Aligned}: \rkshort{} $\subseteq$ \skshort{} -- All of the behaviours exhibited by $\mathcal{M}$ meet the desired ones. More specifically, if \rkshort{} = \skshort{}, we say the behaviours are \textit{fully aligned}. Otherwise, if \rkshort{} $\subset$ \skshort{}, we say the behaviours are \textit{partially aligned}.
    % The \rk{} behaviours are a subset of the \sk{} behaviours, thus, $\mathcal{M}$ partially exhibits the desired behaviours.
    \item \textbf{Expanded}: \rkshort{} $\cap$ \skshort{} $\neq \emptyset \land \exists s_i^{\rkshort{}} \notin$ \skshort{} -- A VLM $\mathcal{M}$ partially exhibits the desired behaviours, but additional, unknown behaviours are observed.
    % The \sk{} behaviours are only partially observed in $\mathcal{M}$. However, $\mathcal{M}$ exhibits additional \rk{} behaviours for which no reference exists. 
    \item \textbf{Divergent}: \rkshort{} $\cap$ \skshort{} $= \emptyset$ -- A VLM $\mathcal{M}$ does not exhibit any of the desired behaviours.
    % There is no correspondence between the specified desired behaviours and observed behaviours.
    % This third type of failure is the hardest to classify as the model might be indeed wrong or the specification insufficient. 
    % \jie{even in Type 1 or 2 the specifications can be insufficient. Shall we separate the failure type definition from potential issues in execution?} \lore{So, do we only mention the fact that, in this case, the model is incorrect?}
\end{itemize}

These behaviour definitions can be used to diagnose VLMs at the task level (e.g., visual question answering) to construct articulate model behaviours.
While broadly applicable, we foresee the need to refine these definitions to better cover nuanced VLM behaviours.
% These behaviour definitions are broadly applicable and can be used to diagnose VLMs at the task level (e.g., visual question answering) to construct articulate model behaviours.
% For this reason, we foresee the need for application-specific refinements to our definitions to diagnose nuanced VLM behaviours more effectively, like in the case of unobserved \skshort{} or unexpected \rkshort{} behaviours.

% (1) the desired but unobserved \skshort{} behaviours in the case of partial alignment,
% (2) the newly observed \rkshort{} behaviours in case of expanded model behaviour, and 
% (3) the degree of divergence between \rkshort{} and \skshort{} behaviours.

% We aim to create symbolic specifications of what $\mathcal{M}$ \sk{} and \rk{} that are usable for diagnostic purposes.
% its behaviour and uncovering data instances that cause reliability issues.
\section{The \diagmethod{} Framework} \label{sec:method}

\begin{figure*}[t]
    \centering
    \includegraphics[width=0.98\textwidth, keepaspectratio]{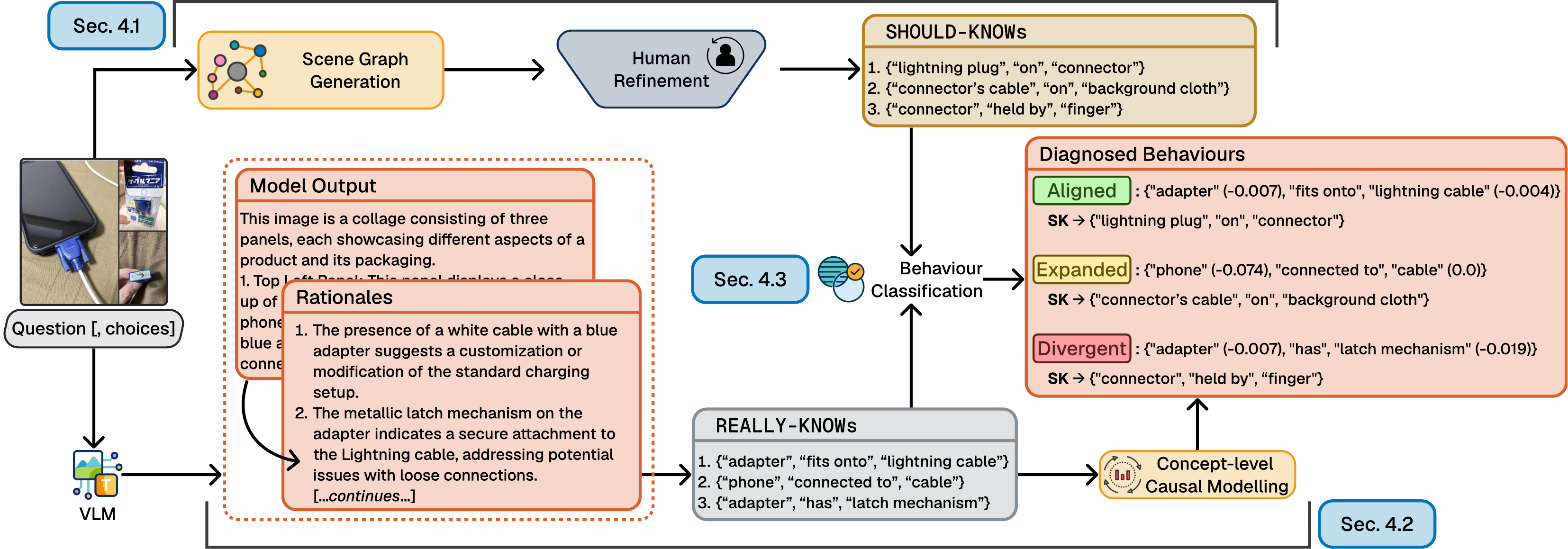}
    \caption{In \diagmethod{}, two parallel processes enable the classification of VLM behaviours. (Top) Scene graph generation and human curation produce \skshort{}s that are relevant for the visual and language inputs. (Bottom) VLMs provide \rkshort{}s by verbalising rationales for their responses. We also compute concept-level causal estimates for deeper insights. (Middle) Finally, \diagmethod{} classifies model behaviours based on lexical and semantic matching.}
    \label{fig:framework}
\end{figure*}

We illustrate the \diagmethod{} diagnostic framework (\autoref{fig:framework}) and its components: the definition of \sk{}s, the extraction of \rk{}s, and the classification of VLM behaviours.

\begin{comment}
We use scene graphs as a reference point for the knowledge that VLMs should have and the feature attribution maps to describe what VLMs do.
However, feature attribution maps (1) do not specify in human-readable terms the concepts a model might be looking at (2) nor the causal effect of those concepts on the model output.
To tackle the first point, we use feature maps to identify subgraphs from the scene graphs by looking at the overlap between bounding boxes from scene graphs and salient patches in feature attribution maps.
Afterwards, to tackle the second point, we apply a controlled perturbation model to generate variations of the scene graphs just filtered (\autoref{sec:pert_model}). This set of \textit{counterfactual} scene graphs is used to compute the causal strengths of individual variables (i.e., concepts and relationships) towards the labels assigned by VLMs (\autoref{sec:causal_effects}).
Given the original and counterfactual scene graphs, we fit a regression model and obtain the causal strengths of individual variables.
\end{comment}

\subsection{Defining \sk{}s} \label{sec:method_sk}
The first step in \diagmethod{} is to construct \skshort{} specifications to serve as a diagnostic reference throughout \diagmethod{} to identify model behaviours and estimate the concept-level causal effects on VLM outputs. % of individual concepts towards a given VLM output.
Concretely, we first use scene graph generation (SGG) to obtain a structured representation of images.
% of the concepts and their relationships in images.
Since these descriptions can be imprecise and generated in a task-agnostic manner (i.e., they do not account for the associated input prompt), we verify and expand them through human annotation to ensure correctness and task relevance. % effectiveness for diagnosis?
% These allow us to specify and verify concept-level representations of the images over which a VLM needs to reason.
% to refine and correct \sk{} specifications $SG_{SK}$ for a model $\mathcal{M}$ and a set of image $\mathcal{I}$.
% Prior human-in-the-loop works showed the effectiveness of incorporating human input to define model behaviour \cite{Sharifi2022Scalpel, Balayn2021SECA}.

\subsubsection{Scene Graph Generation} \label{sec:graph_extract}
Given an image, SGG produces: (1) a set of concepts $\mathcal{C} = \{c_i\} = \{(b_i, l_i)\}$ where each object is localised within the image with a bounding box $b_i$ and belongs to a specific class $l_i$ (e.g., person); and, (2) a set of relational triplets $\mathcal{E} = (c_i, r_{ij}, c_j)$ where $r_{ij}$ relates concepts $c_i$ and $c_j$ (e.g., `held by').
In \diagmethod{}, we use IETrans \citep{Zhang2022IETrans}, a state-of-the-art method for unbiased SGG, which uses 70k distinct concept labels and 1.8k relationship labels.
IETrans offers competitive performance, is lightweight, and provides long-tail coverage, avoiding collapsing to uninformative labels generated by other SGG methods.
Note that while modern VLMs demonstrate open-vocabulary object detection capabilities \citep{Li2024SGGVLM, Wang2024SGGVLM}, in our early tests, we found them noisier and less reliable than IETrans.
Therefore, we did not pursue an empirical investigation of the use of VLMs in \diagmethod{}. This would also introduce a form of evaluation circularity and model self-preference bias \citep{Panickssery2024LLMSelfPreference}.

\subsubsection{Human Curation}
We rely on human contributors to curate the descriptions obtained through SGG, since these can be imprecise and are agnostic to the associated text input.
% Since the descriptions obtained through SGG can be imprecise and do not account for the text input associated with a given image, we rely on human contributors to further curate the symbolic descriptions, ensuring correctness and task relevance.
Despite the scalability of LLM-as-a-judge strategies, we opted for human curation, as LLM judges often require human-written references to combat performance degradation \citep{Krumdick2025LLMJudgesNeedHuman} and can exhibit documented preferential biases \citep{Ye2025LLMJudgesBiases}.
Therefore, we leave the study of LLM judges for diagnosis as future work.
% Furthermore, this step allows us to obtain task-oriented specifications (i.e., related to a given prompt) which would be otherwise not possible with IETrans alone.

For \diagmethod{}, we design and implement two separate human curation tasks to verify and expand the SGG step's output. % \footnote{See \autoref{app:crowd_tasks} for our crowdsourcing setup.}
%
% \noindent \textbf{Verification Task}.
In the \textbf{verification task}, contributors check the correctness and the task relevance (i.e., with respect to the input text and image) of the triplets extracted in the SGG step. Here, we assist them in copy-editing the triplets with a lightweight auto-complete mechanism based on the concept and relationship labels used by IETrans.
Note that contributors are not restricted to these label sets and can provide new ones already at this stage.
%
% \noindent \textbf{Expansion Task}.
Following this, with the \textbf{expansion task}, we seek to obtain additional triplets not identified in the SGG step.
We asked contributors to indicate any relevant but missing triplets needed to extend the previously verified \skshort{} specifications, if any.
Here, annotations comprise concept and relationship labels, as well as concept bounding boxes.

\subsection{Extracting \rk{}s} \label{sec:method_rk}
We compose \rk{} specifications by leveraging self-explanations from VLMs \citep{Madsen2024SelfFidelity}.
% Evidence on the reliability and fidelity of self-explanations is mixed.
While self-explanations (and chain-of-thought traces) may not always be faithful to the model \citep{Madsen2024SelfFidelity, Agarwal2024FaithfulnessSelfExplanations}, they may be comparably plausible to other explanation approaches, particularly when a task is decomposed adequately \citep{Huang2023LLMSelfExplain, Turpin2023CotTFidelity}.
% Through self-explanations, the model can theoretically simulate its own behaviour as the weights are shared between explanation generation and other tasks.\footnote{Also called \textit{self-model capability} \citep{Kadavath2022LLMSelfModel}.}
In \diagmethod{}, we decompose the task into three steps, allowing the model to carry it out effectively \citep{Wang2023PlanSolveLLMs, Randl2025ReliabilitySelfExpl}.
First, we collect the VLM's responses to image-text pairs.
Second, we prompt the VLM to produce rationales that corroborate its responses.
Finally, we ask the VLM to format its rationales, for comparison with the \sk{} specifications.

% This procedure is lightweight and can be used with open-source and closed-source VLMs. %, and can be extended (or repeated) to illuminate model inconsistencies and unfaithful self-explanations.

\paragraph{Localising Concepts}
We recover the bounding boxes of the \rkshort{} concepts by first attempting to match \rkshort{} and \skshort{} triplets. If unsuccessful, we use the open-vocabulary object detection model OWLv2 \citep{Minderer2023OWLv2} and similarity matching as fall-back methods.\footnote{We use \href{https://huggingface.co/ibm-granite/granite-embedding-english-r2}{\texttt{ibm-granite/granite-embedding-english-r2}}.}
This information is also used in our causal modelling approach.

\subsubsection{Causal Modelling of Concept Roles} \label{sec:causal_estimation}
% To identify causal dependencies between image concepts and model outputs, we first generate counterfactual data to feed to the VLMs and the corresponding \textit{alternative} output.\footnote{In \diagmethod{}, we consider the input text as a constant.}
% Then, we estimate concept-level causal effects for each data sample by fitting a new estimator on individualised sample-level, transparent perturbations.

While this procedure is lightweight, a zero-shot approach does not differentiate between the concepts VLMs use \textit{vs} those they simply see.
To quantify the role of different concepts, we leverage causal modelling to estimate their contributions to the observed VLM outputs.
% guarantee that the \rkshort{} specifications accurately reflect the concepts (and relationships) the VLMs use.
% To quantify the role of these concepts, we propose leveraging causal modelling to estimate the effects of individual concepts on the observed model outputs.
Note that, while causal modelling provides a verification layer to self-explanations, this differs from quantifying the fidelity of the \rkshort{}s, which we leave for future work.
Concretely, we model causal relationships in two steps: (1) transparent generation of counterfactual model input-output pairs, and (2) fitting of a double-machine learning (DML) estimator to obtain the concepts' causal effects.

\paragraph{Counterfactual Generation}
To quantify the effect that individual concepts have on model outputs, we first need counterfactual input-output pairs, with alternative concept compositions, and then observe the corresponding model outputs.
Contrary to prior work that leverages neural methods \citep{Alvarez2017CausalSeq2Seq, Xu2021CausalExplRecsys}, we define a transparent, concept-level image perturbation strategy, keeping the input text constant.

% Instead of employing a neural model we adopt a heuristic approach to perturb \rk{} specifications in a controlled manner such that interventions are carefully designed to target specific concepts \cite{woodward2005making}.
% The motivation is two-fold: (1) intervening on individual concepts ensures that possible (un-)observed confounders do not influence the estimated causal effects; (2) controlling the perturbations allows us to focus on naturally occurring $\widetilde{\rkshort{}}$ counterfactuals.

For each \textit{original} data sample, we take its \rkshort{}$^{x_i}_{\mathcal{M}}$ specifications and compute the powerset $\mathbb{P}($\rkshort{}$^{x_i}_{\mathcal{M}})$.
Each element $\mathcal{C} \in \mathbb{P}($\rkshort{}$^{x_i}_{\mathcal{M}})$ represents a combination of concepts $\{c_i, ..., c_j\} \in $ \rkshort{}$^{x_i}_{\mathcal{M}}$.
% Elements $ \in \mathbb{P}($\rkshort{}$^{x_i}_{\mathcal{M}})$ represent the possible combination of concepts from \rkshort{}$^{x_i}_{\mathcal{M}}$.
Based on these combinations, we use the bounding boxes of the concepts to create occlusion masks on the image, thereby obtaining \textit{counterfactual inputs}.
We prevent unwanted interventions on concepts by occluding solely the concepts within a given $\mathcal{C} \in \mathbb{P}($\rkshort{}$^{x_i}_{\mathcal{M}})$, leaving the others visible.
Finally, we generate \textit{counterfactual responses} $\hat{Y}^c$ by simply running inference with the VLM being inspected.
% Given a \rkshort{}$^{x_i}_{\mathcal{M}}$ specification, we first compute its powerset $\mathbb{P}($\rkshort{}$^{x_i}_{\mathcal{M}})$ with respect to its concepts.

% In practice, given a \rk{} specification \rkshort{}$_i$ for a model $\mathcal{M}$ and image, we compute its powerset $\mathbb{P(\cdot)}$ to cover the possible and plausible interventions fully, thus obtaining the set of counterfactual \rk{} specifications $\widetilde{\rkshort{}}_i$.

\paragraph{Causal Effect Estimation}
Given the inherent nonlinearity of VLMs, we opt for a DML approach \citep{Chernozhukov2018DoubleML}.
DML estimators do not rely on parametric assumptions (unlike, e.g., Gaussian mixture density estimators) and are better suited to capture complex relationships in data.
For this, we define a set of covariates $Z_i = \mathcal{C}_{x_i} \setminus {c_i}$, the treatment variable $T_i = c_i$, and the outcome variable $\hat{Y}_{x_i} = \hat{Y}^c_{x_i} \cup {\hat{Y}_{x_i}}$. We then train two ML models: $\hat{Y}_{x_i} \approx \hat{f}(Z_i)$, approximating the outcome given $Z_{x_i}$, and $T_i \approx \hat{g}(Z_i)$, approximating the treatment given $Z_{x_i}$.
Finally, causal effects $\hat{\theta}_{c_i}$ are computed based on residuals $\hat{U}_i$ and $\hat{V}_i$, as:
% Maybe a bit hacky 
% \vspace{\abovedisplayskip}
\noindent
\begin{minipage}{.49\columnwidth}
\begin{equation}
    \hat{U}_i = \hat{Y}_{x_i} - \hat{f}(Z_i)
\end{equation}
\end{minipage}%
\hfill
\begin{minipage}{.49\columnwidth}
\begin{equation}
    \hat{V}_i = \mathcal{C}_{x_i} - \hat{g}(Z_i)
\end{equation}
\end{minipage}
% \vspace{\belowdisplayskip}
\begin{equation}
    \hat{\theta}_{c_i} = (\frac{1}{n} \sum^{n}_{i=1} \hat{V}_i \mathcal{C}_{x_i})^{-1} \cdot \frac{1}{n} \sum^{n}_{i=1} \hat{V}_i \hat{U}_i
\end{equation}

The estimates $\hat{\theta}_{c_i}$ are, in practice, averages of multiple estimates obtained through cross-fitting.
The models $\hat{f}$ and $\hat{g}$ are fitted on $(\phi(T_i,Z_i), \hat{Y}_{x_i})$ where $\phi(T,Z)$ encodes the presence (or absence) of the treatment variable and covariates (i.e., binary vector) based on the combinations in $\mathbb{P}($\rkshort{}$^{x_i}_{\mathcal{M}})$.
% ; and $\hat{Y}^c$ is the corresponding counterfactual model response.
In \diagmethod{}, we use gradient boosted trees.\footnote{Theoretically, any ML model can be used for $\hat{f}$ and $\hat{g}$.}
See \autoref{app:dml_setup} for the implementation details.

Generally, causal estimators rely on a causal graph, i.e., a direct acyclic graph describing relationships between variables.
In \diagmethod{}, we use the \rkshort{}$_\mathcal{M}$ specifications as proxies for causal graphs. We discuss related caveats later in the paper.

\paragraph{Handling Longer Responses}
% The application of causal modelling to longer VLM responses is not straightforward.
In \diagmethod{}, we extract the concepts $\{c_i, ..., c_j\}$ from the original model response $\hat{Y}$ and construct binary vectors encoding the presence or absence of $c_i$ in $\hat{Y}_{x_i}$.
Then, we estimate the causal effects $|\{c_i, ..., c_j\}|$ times given the $Z_i$ covariates, as described above.

% \subsection{Model Behaviour Charactarisation} \label{sec:causal_relations}
% We characterise VLM behaviours by first classifying the behaviour types described in \autoref{sec:problem_statement}, and then applying causal modelling to estimate the effects of individual concepts on the model outputs.
% (1) generating counterfactual \rkshort{} specifications to estimate the causal effects $\theta_i$ of individual triplets in ${RK}$ on the VLM output; and, 
% (2) computing the similarity between \skshort{} and \rkshort{} to identify behaviour drifts.
% The relational triplets are then ranked based on $ce_i$ and aggregated to provide behavioural characterisation for individual \texttt{<image, prompt, output>} triplets.

\subsection{Classification of Behaviours} \label{sec:classify_behaviours}
% One can naively classify model behaviours by directly applying the definitions in \autoref{sec:problem_statement} with comparisons that are purely lexical.
% However, such an approach would not account for the similarities that occur naturally in language and that a VLM might have learnt.
We propose classifying VLM behaviours from a semantic perspective, thereby accounting for naturally occurring language similarities.
In \diagmethod{}, we quantify the cosine similarity $S_c(e($\skshort{}$),e($\rkshort{}$))$ between the embedded \skshort{} and \rkshort{} specifications.\footnote{We use \href{https://huggingface.co/sentence-transformers/all-mpnet-base-v2}{\texttt{sentence-transformers/all-mpnet-base-v2}}. The \href{https://huggingface.co/ibm-granite/granite-embedding-english-r2}{\texttt{ibm-granite/granite-embedding-english-r2}} used before would erroneously assign $S_c>0.8$ to all \skshort{}-\rkshort{} pairs.}
Given the \skshort{} and \rkshort{} behaviours of a data sample, \diagmethod{} iteratively matches the \rkshort{} with the \skshort{} with the largest $S_c$ until either set is exhausted. This allows us to isolate both expected but unobserved behaviours and new ones.
Behaviours can then be classified with flexible thresholding as:
% \textbf{Aligned} if $|\rkshort{}| \leq |\skshort{}|$ and $S_c \geq \tau_a$, \textbf{Expanded} if $|\rkshort{}| > |\skshort{}|$ and $S_c \geq \tau_e$, or \textbf{Divergent} if $S_c < \tau_d$.

\begin{itemize}[leftmargin=*]
    \item \textbf{Aligned}: if $|\rkshort{}| \leq |\skshort{}|$ and $S_c \geq \tau_a$
    \item \textbf{Expanded}: if $|\rkshort{}| > |\skshort{}|$ and $S_c \geq \tau_e$
    \item \textbf{Divergent}: $S_c < \tau_d$
\end{itemize}

\section{Experiments}
We seek to answer two main questions:
(\textbf{Q1}) What common behaviour types are identified in VLMs with \diagmethod{}?; and, (\textbf{Q2}) How informative are the behaviours found with \diagmethod{}?

% How effective is our framework in characterising VLM behaviours?;

% \subsection{Experimental Setup}

\paragraph{Models \& Datasets}
We apply \diagmethod{} on four well-known VLMs: 
InternVL2 8B \citep{Chen2024InternVL},
LLaVa-1.6 7B \citep{liu2024llavanext},
Qwen2.5-VL 7B \citep{Bai2025Qwen25VL}, and
ShareGPT4V 7B \citep{chen2023sharegpt4v}. We test these on four datasets:
% \paragraph{Datasets}
% We use four common VLM benchmark datasets and sample image-question pairs to apply and test \diagmethod{} across various scenarios.
\begin{itemize}[leftmargin=*]
    \item \textit{LLaVa-Bench} \citep{Liu2023LLaVa}: All 60 open-ended, diverse, and challenging questions. We anticipate VLMs to struggle with this data.% over 24 images.
    \item \textit{MMBench} \citep{Liu2024MMBench}: Repurposed captioning questions from the ``Image Scene'' (75) and ``Image Topic'' (64) classes as open-ended. %\footnote{Captioning questions in MMBench require picking the most appropriate caption among the given alternatives.}
    % by filtering captioning questions from the `Image Scene' and `Image Topic' categories and using the ground truth to determine errors, obtaining 255 and 64 image-question-answer triplets respectively.
    \item \textit{SEED-Bench 2} \citep{Li2023SEEDBench}: Multiple-choice questions from the scene understanding (75) and visual reasoning (75) categories.
    % used to assess text and image perception and reasoning. 
    % Scene Understanding (3158), Visual Reasoning (331), Spatial Relation (657; discarded)
    \item \textit{VQA v2} \citep{Goyal2017VQAv2}: Multiple-choice questions from the ``How many people are...'' (75) and ``What is the person...''(75) categories.
    % How many people are... (2005), What is the person... (900).
    % `Is the person', `Is this person', It contains open-ended question-answers pairs for a given image. Ground-truth answers have multiple annotations, often as a single word.
\end{itemize}

% \jie{the datasets are very nice, covering convincing tasks like QA and visual reasoning. good to show a couple of examples how the data/task looks like, perhaps in the appendix.} \lore{Noted, I'll add this!}

% In \autoref{sec:app_data_examples} we report representative examples from each dataset we use in our experiments.

\paragraph{Crowdsourcing}
We recruited 520 workers (avg. wage: 8 GBP/h) on Prolific from English-speaking countries, with an approval rate of $\geq 95\%$. This study received ethics approval from our institution (ID: 4696). Refer to \autoref{app:crowd_tasks} for additional details and screenshots of the task interfaces.

\paragraph{Prompt Templates}
For each model-dataset combination, we fine-tune our prompts to ensure consistent instruction following. See \autoref{app:prompt_templates}.

\paragraph{Implementation Details}
We self-host the VLMs on two NVIDIA A10 GPUs. We use their default parameterisation, limit response length to 256 tokens, and employ greedy decoding for reproducibility. Additional details in \autoref{app:impl_details}.

\paragraph{(Q1) Common Model Behaviour Types}
We use \diagmethod{} to classify VLM behaviours as Aligned, Expanded, and Divergent as proposed in \autoref{sec:classify_behaviours}.
To this end, we also report on our exploration and selection of semantic similarity thresholds.
Finally, we qualitatively inspect 200 samples (covering 952 \rkshort{}s) to surface recurring behaviours.% across behaviour types and in relation to the corresponding \skshort{}s.

% \begin{itemize}
%     \item Classification of behaviours \lore{connect with the para in the method}
%     \item Mention: how did we choose the thresholds?  \lore{brief}
%     \item Manual inspection: how did we do this?
% \end{itemize}

\paragraph{(Q2) Informativeness of Model Behaviours}
We study the informativeness of behaviours from two perspectives.
First, we study how well the similarities between the \skshort{} and \rkshort{} produced by \diagmethod{} can describe model performance.
Even if a VLM performs satisfactorily, it is not a given that the behaviours \diagmethod{} finds align with performance.
This last point would likely require a mechanistic analysis, which is outside the scope of this work.
Since we cannot assume a priori that these two variables are linearly related, we bootstrap Mutual Information (MI).\footnote{Refer to \autoref{app:mi_bootstrapping} for our MI bootstrapping procedure, including significance testing and comparison with randomly shuffled behaviour similarities.}
% We use MI since we cannot assume a priori that model performance and behaviour types are linearly related.
Second, we do model-pairwise comparisons across datasets to explore the distributions of concept-level causal effects.
To mitigate potential biases, we report the median causal estimates across $n=100$ random data splits.
% We compute causal estimates across $n=100$ random data splits and report the medians, mitigating potential biases.

% \begin{itemize}
%     \item use MI to quantify the relationship between behaviour types and model performance. Bootstrapping etc.
%     \item report the causal analysis here? Remember to explain the choice of covariates and treatment variables at each step.
% \end{itemize}
% \input{sections/results}
\section{Results \& Discussion} \label{sec:results}

%% ==== Custom commands for tcoloboxes for examples
\definecolor{llavabenchcolor}{HTML}{DC602E}
\definecolor{mmbenchcolor}{HTML}{D7B49E}
\definecolor{seedbenchcolor}{HTML}{B8D5B8}
\definecolor{vqav2color}{HTML}{05A8AA}
\newcommand{\downrightarrow}{%
    \hspace*{0.3em}\tikz[baseline=-0.5ex,>=stealth]{\draw[->] (0,0.3) -- (0,0) -- (0.3,0);}%
}
%% ==== Custom commands for tcoloboxes for examples

\subsection{Overview of \skshort{} and \rkshort{} specifications} \label{sec:overview_sk_rk}

% \paragraph{Descriptive Statistics}
% \autoref{fig:sk_stats} and \autoref{fig:rk_stats} show the distributions of concepts and relations across the \texttt{SK} and \texttt{RK} specifications. \lore{update figure refs}
% We provide the exact numbers in \autoref{tab:spec_counts}.
% \lore{add a different figure of the distribution?}

\begin{comment}
\begin{table}[t]
\centering
\resizebox{\columnwidth}{!}{%
\begin{tabular}{@{}lllll@{}}
\toprule
 & \textbf{LLaVa-Bench} & \textbf{MMBench} & \textbf{SEED-Bench 2} & \textbf{VQA v2} \\ \midrule
\texttt{SHOULD-KNOW} & 798 (13.3; 5.7) & 2198 (15.8; 6.1) & 2262 (15.1; 2.5) & 2336 (15.6; 2.8) \\ \midrule
InternVL2 & 553 (9.5; 3.3) & 1234 (8.9; 2.8) & 1173 (7.8; 3.1) & 945 (6.3; 3.5) \\
LLaVa-1.6 & 266 (6.0; 3.3) & 776 (5.8; 3.1) & 798 (6.2; 3.5) & 889 (6.3; 3.5) \\
MiniGPT-4 & 497 (9.2; 3.1) & 979 (9.6; 2.8) & 1307 (9.3; 2.9) & 1126 (9.2; 2.8) \\
ShareGPT4V & 260 (5.3; 3.3) & 622 (4.6; 3.1) & 233 (2.1; 2.3) & 43 (2.0; 2.6) \\ \bottomrule
\end{tabular}%
}
\caption{Total number of \texttt{SK} (post human annotation) and \texttt{RK} specifications obtained. In parentheses, we report (mean; standard deviation) across data samples.}
\label{tab:spec_counts}
\end{table}
\end{comment}

\paragraph{Inspecting \sk{}s}

\begin{table}[t]
\centering
\resizebox{\columnwidth}{!}{%
\begin{tabular}{@{}lllll@{}}
\toprule
& \textbf{LLaVa-Bench} & \textbf{MMBench} & \textbf{SEED-Bench 2} & \textbf{VQA v2} \\ \midrule
Initial    & 580 (-)      & 1442 (-)      & 1783 (-)      & 1798 (-)      \\
Validation & 300 (51.7\%) & 839  (58.2\%) & 1042 (58.4\%) & 1051 (58.5\%) \\
Expansion  & 516 (43.4\%) & 1024 (42.1\%) & 1233 (38.7\%) & 1302 (46.0\%) \\ \midrule
Expert     & 546 (11.0\%) & 1094 (5.9\%)  & 1301 (8.4\%)  & 1391 (7.8\%)  \\
M$_\skshort{}$ (IQR) & 9 (5.75, 11) & 7 (5, 10.5)   & 9 (5, 12)     & 8.5 (6,12)    \\ \bottomrule
\end{tabular}%
}
\caption{\skshort{}s collected and percentages of relevant \skshort{}s (Validation), of new \skshort{}s (Expansion), or of expert edits (including median (M$_\skshort{}$) and inter-quartile range (IQR)).}
\label{tab:sk_stats}
\end{table}

Each human annotator worked on 5 data samples, covering a varying number of candidate \skshort{}s. Each sample was shown to a single human annotator. \footnote{See \autoref{app:ietrans_ablation} for ablations on IETrans. We also note that, by first showing contributors triplets produced with IETrans, they may have exhibited anchoring bias.}
Afterwards, with the help of colleagues from our computer science department, we manually verified that the crowdsourced \skshort{}s behaviours (\autoref{tab:sk_stats}) were properly formatted and correctly aligned with the input text and image. 
We found that a maximum of $\approx 11\%$ (\autoref{tab:sk_stats}; Expert row) of behaviours required corrections for trivial mistakes -- mostly typos and incorrectly formatted data.
In a few cases, contributors provided valid triples even when only the concept or relation labels were expected, thereby slightly increasing the final number of \skshort{}s.
% $\sim 11\%$ for LLaVa-Bench, $\sim 5.9$ for MMBench, $\sim 8.4\%$ for SEED-Bench, and $\sim 7.8\%$ for VQA v2.
% In our experiments, we use the latter \skshort{} category.

\paragraph{Inspecting \rk{}s}

\begin{table*}[t]
\centering
\resizebox{\textwidth}{!}{%
\begin{tabular}{@{}lllllllll@{}}
\toprule
           & \multicolumn{2}{c}{\textbf{LLaVa-Bench}}                   & \multicolumn{2}{c}{\textbf{MMBench}}                        & \multicolumn{2}{c}{\textbf{SEED-Bench 2}}                   & \multicolumn{2}{c}{\textbf{VQA v2}}    \\ \midrule
           & N$_s$ (N$_{\rkshort{}}$) & \multicolumn{1}{l|}{M$_{\rkshort{}}$ (IQR)} & N$_s$ (N$_{\rkshort{}}$) & \multicolumn{1}{l|}{M$_{\rkshort{}}$ (IQR)} &  N$_s$ (N$_{\rkshort{}}$) & \multicolumn{1}{l|}{M$_{\rkshort{}}$ (IQR)} & N$_s$ (N$_{\rkshort{}}$) & M$_{\rkshort{}}$ (IQR) \\ \midrule
InternVL2  & \cellcolor{green!20}58 (565)  & \multicolumn{1}{l|}{10 (7, 12)}  & \cellcolor{green!20}139 (1287) & \multicolumn{1}{l|}{9 (7, 11)} & \cellcolor{green!20}150 (1254) & \multicolumn{1}{l|}{8 (6, 10)} & \cellcolor{green!20}150 (1014) & 7 (4, 9) \\
LLaVa-1.6  & \cellcolor{yellow!20}44 (279) & \multicolumn{1}{l|}{7 (3, 8.25)} & \cellcolor{green!20}133 (823)  & \multicolumn{1}{l|}{6 (4, 8)}  & \cellcolor{yellow!20}129 (853) & \multicolumn{1}{l|}{6 (4, 10)} & \cellcolor{green!20}142 (939)  & 6 (4, 9) \\
Qwen2.5-VL & \cellcolor{green!20}56 (504)  & \multicolumn{1}{l|}{9 (6, 12)}   & \cellcolor{green!20}139 (601)  & \multicolumn{1}{l|}{4 (3, 5)}  & \cellcolor{green!20}150 (554)  & \multicolumn{1}{l|}{3 (2, 4)}  & \cellcolor{green!20}150 (336)  & 2 (2, 2) \\
ShareGPT4V & \cellcolor{yellow!20}49 (279) & \multicolumn{1}{l|}{5 (3, 8)}    & \cellcolor{green!20}135 (645)  & \multicolumn{1}{l|}{4 (2, 7)}  & \cellcolor{yellow!20}109 (241) & \multicolumn{1}{l|}{1 (1, 2)}  & \cellcolor{red!20}21 (46)      & 1 (1, 1) \\ \bottomrule
\end{tabular}%
}
\caption{\rkshort{} statistics shown across models and datasets: raw number of \rkshort{} extracted (N), median (M), and inter-quartile range (IQR). We highlight model-dataset pairs based on the number of \rkshort{} successfully extracted: $>90\%$ in green, $70-90\%$ in yellow, and $<70\%$ in red. We include results from ShareGPT4V on VQA v2 for transparency.}
\label{tab:rk_stats}
\end{table*}

% \autoref{tab:rk_stats} gives an overview of the \rkshort{} extracted by \diagmethod{}.
% To obtain the \rk{}s, we iterated on our prompts for specific VLM-dataset combinations.
We extracted from the four VLMs \rkshort{}s for 1754/1996 initial samples, losing $\approx 12\%$ due to empty responses and unstable instruction-following, with ShareGPT4V as the loss leader (\autoref{tab:rk_stats}).
We cleaned up the repeated \rkshort{}s produced by the VLMs (likely due to greedy decoding) by coalescing them into unique ones.
% We discuss this point in \autoref{sec:limitations}.

\subsection{Common Model Behaviour Types (Q1)}

\paragraph{Setting Behaviour Thresholds}

\begin{figure}[t]
    \centering
    \begin{subfigure}[b]{0.95\columnwidth}
        \centering
        \includegraphics[width=0.95\textwidth]{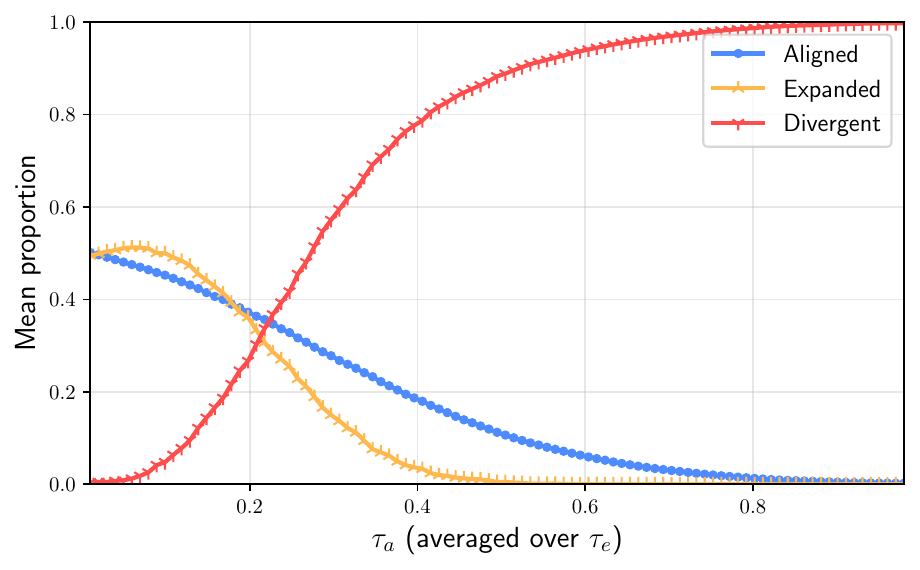}
        % \caption{LLaVa-Bench}
        \label{fig:sensitivity_tau_a}
    \end{subfigure}
    % new line
    \begin{subfigure}[b]{0.95\columnwidth}
        \centering
        \includegraphics[width=0.95\textwidth]{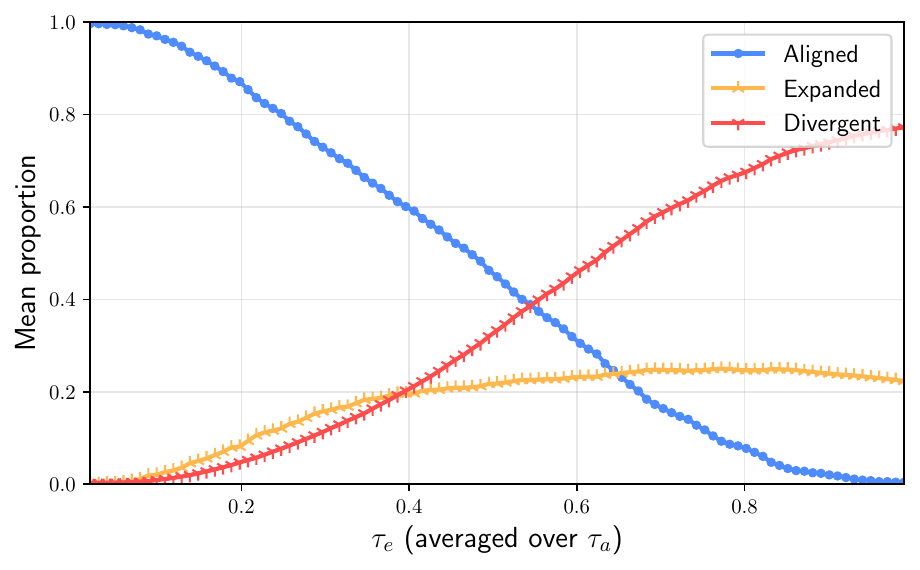}
        % \caption{SEED-Bench-2}
        \label{fig:sensitivity_tau_e}
    \end{subfigure}
    \caption{Behaviour thresholds' sensitivity to $\tau_a$ (top) and $\tau_e$ (bottom). $\tau_d$ is set afterwards, based on $\tau_e$. Additional visualisations are included in \autoref{app:th_sensitivity}.}
    \label{fig:sensitivity_curves}
\end{figure}

We perform a sensitivity analysis to set the behaviour thresholds $\tau_a$, $\tau_e$, and $\tau_d$ (\autoref{fig:sensitivity_curves}).
We found that $\tau_a$ plateaus in $[0.65, 0.8]$, while a $\tau_e \geq  0.4$ leads to an increasing number of incorrectly-classified, Divergent samples.
To corroborate this, we manually inspected 300 aligned \skshort{}-\rkshort{} pairs and 300 divergent \skshort{}-\rkshort{} pairs.\footnote{$S_c$: $\wedge = 0.021$; $\vee = 0.92$; $\overline{S_c} = 0.506$; $\sigma_{S_c} = 0.152$.}
% using a naive 3-way split of the $[0,1]$ range for $\tau_a$, $\tau_e$, and $\tau_d$.\footnote{The measured $S_c$ have $\min = 0.021$; $\max = 0.92$; $\overline{S_c} = 0.506$; and, $\sigma_{S_c} = 0.152$.}
%
With $\tau_a \geq 0.7$, we found $22.31\%$ of the pairs had only one similar concept and were falsely classified as aligned.
With a more conservative $\tau_a \geq 0.75$, we reduced this to $17.70\%$ so that cases with one similar concept and relationship are accounted for.
This also increased the average similarity $\overline{S_c}$ from $0.77$ to $0.84$.
On the other hand, setting $\tau_d < 0.35$ already led to $92.69\%$ of the pairs having dissimilar or distinct concepts and relationships ($\overline{S_c} = 0.21$).
In \diagmethod{}, we thus use: $\tau_a \geq 0.75$; $\tau_e \geq 0.35$; and $\tau_d < 0.35$.

\paragraph{Behaviours Identified}
\begin{figure*}[t]
    \centering
    \includegraphics[width=\textwidth]{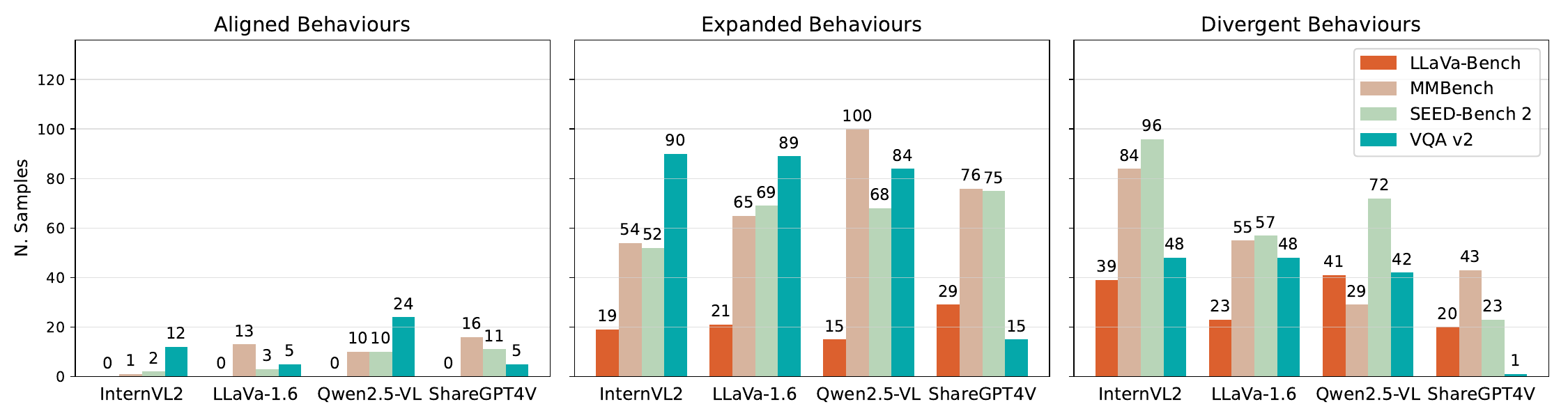}
    \caption{Identified Model Behaviours.}
    \label{fig:behaviour_counts}
\end{figure*}

\autoref{fig:behaviour_counts} shows the distribution of behaviours classified by \diagmethod{}.
Overall, we observe that the VLMs we tested mostly exhibit expanded or divergent behaviours.
Refer to \autoref{app:examples_behaviours} for complete behaviour examples.

\noindent \textbf{(1) Aligned Behaviours ($6.4\%$):}
% Despite finding a low number of aligned behaviours, we believe our constraints are stringent but effective.
\diagmethod{} highlighted a low number of aligned behaviours in the VLMs tested.
Here, the \rkshort{}s are highly similar to the corresponding \skshort{}s: concept and relationship labels are often identical and structured in the same way.
The main qualitative difference 
% (which is accounted for by our semantic checks) 
for aligned behaviours concerns the use of active or passive voice when formulating a \skshort{} or a \rkshort{}, such as:

\begin{tcolorbox}[
    left=1pt,right=1pt,top=1pt,bottom=1pt,
    colframe=seedbenchcolor,
    colback=seedbenchcolor!20,
    colbacktitle=seedbenchcolor!20,
    coltitle=black,
    title=\small\textbf{Qwen2.5-VL on SEED-Bench 2 (Sample ID: 57)}]
\small
\noindent\textbf{\rkshort{}}: \{\texttt{"guitar"}, \texttt{"being played by"}, \texttt{"person"}\}\\
\downrightarrow \noindent\textbf{\skshort{}$_{\texttt{match}}$}: \{\texttt{"player"}, \texttt{"playing"}, \texttt{"guitar"}\}
\end{tcolorbox}

\noindent Predictably, the \skshort{} and \rkshort{} from LLaVa-Bench differ significantly despite comparable performance.
% We note that LLaVa-Bench proved challenging for the VLMs tested -- \skshort{} and \rkshort{} are radically different despite comparable offline performance.

\noindent \textbf{(2) Expanded Behaviours ($52.5\%$):}
Overall, the expanded behaviours cover \skshort{}-\rkshort{} pairs that are thematically related (e.g., clothing) and exhibit semantic or structural differences:
% but differ at the conceptual and relational levels.
% Here, we highlight two common phenomena.
\begin{enumerate}[leftmargin=*]
    \item \textit{Describing over Composing}: We found \rkshort{}s to be largely descriptive (e.g., the properties of objects) rather than compositional (e.g., spatial locations of objects).
    This pattern aligns with prior research of VLMs' blind spots and gaps in their visual capabilities \citep{Fu2025VLMFailures}.
    \item \textit{Organisation of Concepts}: While similar (or identical) concepts are covered in both \skshort{}s and \rkshort{}s, the relationships and concept combinations differ. This might hint at concept-level, latent prioritizations applied by VLMs that, under the same circumstances and input data, do not align with those of humans. We note that these do not necessarily constitute faulty behaviours, but should be identified and assessed nonetheless.
\end{enumerate}

% \begin{tcolorbox}[
%     left=1pt,right=1pt,top=1pt,bottom=1pt
%     ]
% \small
% % \textbf{\texttt{Qwen2.5-VL - SEED-Bench - ID: 57}}
% \noindent\textbf{\skshort{}\texttt{\_1}}: \{\texttt{"flowers"}, \texttt{"on"}, \texttt{"table"}\}\\
% \noindent\textbf{\skshort{}\texttt{\_2}}: \{\texttt{"flowers"}, \texttt{"blooming in"}, \texttt{"vase"}\}
% \tcbline
% \noindent\textbf{\rkshort{}\texttt{\_1}}: \{\texttt{"vase"}, \texttt{"is on"}, \texttt{"table"}\}\\
% \noindent\textbf{\rkshort{}\texttt{\_2}}: \{\texttt{"flowers"}, \texttt{"is in"}, \texttt{"vase"}\}
% \end{tcolorbox}

\begin{tcolorbox}[
    left=1pt,right=1pt,top=1pt,bottom=1pt,
    colframe=vqav2color,
    colback=vqav2color!20,
    colbacktitle=vqav2color!20,
    coltitle=black,
    title=\small\textbf{InternVL2 on VQAv2 (Sample ID: 26130006)}]
\small
\noindent\textbf{\rkshort{}}: \{\texttt{"person"}, \texttt{"near"}, \texttt{"wave"}\}\\
\downrightarrow \noindent\textbf{\skshort{}$_{\texttt{match}}$}: \{\texttt{"man"}, \texttt{"going for"}, \texttt{"wave"}\}
\tcbline
\noindent\textbf{\rkshort{}}: \{\texttt{"surfboard"}, \texttt{"part of"}, \texttt{"person"}\}\\
\downrightarrow \noindent\textbf{\skshort{}$_{\texttt{match}}$}: \{\texttt{"man"}, \texttt{"riding"}, \texttt{"surfboard"}\}
\end{tcolorbox}

\noindent \textbf{(3) Divergent Behaviours ($41.1\%$):}
Here, we have cases in which the VLMs rely on very different concepts and relations compared to the \skshort{}s, or organise those concepts in very different ways, more so than Expanded behaviours.
Interestingly, divergent behaviours are also those with a larger number of \rkshort{}s.
Here, the VLMs may be more uncertain about their responses and may attempt to make their output self-consistent. Predictably, the \skshort{}s and \rkshort{}s related to LLaVa-Bench differ significantly despite comparable measured performance.

\begin{tcolorbox}[
    left=1pt,right=1pt,top=1pt,bottom=1pt,
    colframe=llavabenchcolor,
    colback=llavabenchcolor!20,
    colbacktitle=llavabenchcolor!20,
    coltitle=black,
    title=\small\textbf{ShareGPT4V on LLaVa-Bench (Sample ID: 33)}]
\small
\noindent\textbf{\rkshort{}}: \{\texttt{"mug"}, \texttt{"white"}, \texttt{"ceramic"}\}\\
\downrightarrow \noindent\textbf{\skshort{}$_{\texttt{match}}$}: \{\texttt{"letter"}, \texttt{"written on"}, \texttt{"mug"}\}
\tcbline
\noindent\textbf{\rkshort{}}: \{\texttt{"mug"}, \texttt{"sligthly tilted"}, \texttt{"left"}\}\\
\downrightarrow \noindent\textbf{\skshort{}$_{\texttt{match}}$}: \{\texttt{"cap"}, \texttt{"covering"}, \texttt{"head"}\}
\end{tcolorbox}

\subsection{Informativeness of Model Behaviours (Q2)}

\paragraph{Performance-Behaviours Relationship}
Overall, \skshort{}s and \rkshort{}s from \diagmethod{} are good indicators of VLM performance (\autoref{tab:effectiveness_acc}).
Note that their differing magnitudes are due to how performance is computed.
For LLaVa-Bench and MMBench, performance is continuous within $[0, 1]$, and MI is theoretically unbounded.
% For SEED-Bench-2 and VQA v2, instead, performance is discrete in $\{0,1\}$, and MI $\leq ln(2) \simeq 0.693$.
For the other datasets, instead, performance is discrete in $\{0,1\}$, and MI $\leq ln(2) \simeq 0.693$.
Model-wise, we found that InternVL2 and LLaVa-1.6 exhibit behaviours more closely related to their performance.
Qwen2.5-VL behaved consistently only on LLaVa-Bench, while falling off on the rest.
Finally, ShareGPT4V showed better consistency on open-ended data, likely due to its focus on image captioning.
% appears to be more consistent with open-ended questions than with multiple-choice ones. This might be due to the large number of captions in its pre-training data compared to other image-text formats.
%

\begin{table}[t]
\centering
\resizebox{\columnwidth}{!}{%
\begin{tabular}{lcccccccc}
\toprule
\multirow{2}[3]{*}{\textbf{VLM}} & \multicolumn{2}{c}{\textbf{LLaVa-Bench}} & \multicolumn{2}{c}{\textbf{MMBench}} & \multicolumn{2}{c}{\textbf{SEED-Bench 2}} & \multicolumn{2}{c}{\textbf{VQA v2}} \\
\cmidrule(lr){2-3} \cmidrule(lr){4-5} \cmidrule(lr){6-7} \cmidrule(lr){8-9}
               & \multicolumn{1}{c}{A} & \multicolumn{1}{c}{MI} & \multicolumn{1}{c}{A} & \multicolumn{1}{c}{MI} & \multicolumn{1}{c}{A} & \multicolumn{1}{c}{MI} & \multicolumn{1}{c}{A} & \multicolumn{1}{c}{MI} \\ \midrule
InternVL2      & 0.673 & \textbf{3.367} & 0.796 & \textbf{3.200} & 0.787 & 0.379 & 0.667 & 0.476 \\
LLaVa-1.6      & 0.657 & 2.389 & 0.792 & 2.512 & 0.667 & \textbf{0.438} & 0.585 & 0.481 \\
Qwen2.5-VL     & 0.670 & 3.008 & 0.743 & 1.561 & 0.833 & 0.188 & 0.700 & 0.147 \\
ShareGPT4V     & 0.687 & 2.058 & 0.705 & 2.116 & 0.661 & 0.292 & 0.810 & \textbf{0.558}$^\dagger$ \\ \bottomrule
\end{tabular}%
}
\caption{Measured accuracy (A) vs Mutual Information MI$(A, S_c)$. $\dagger$: Very small sample size: see \autoref{tab:rk_stats}.}
\label{tab:effectiveness_acc}
\end{table}

\begin{figure}[t]
    \centering
    \begin{subfigure}[b]{0.49\columnwidth}
        \centering
        \includegraphics[width=0.95\textwidth]{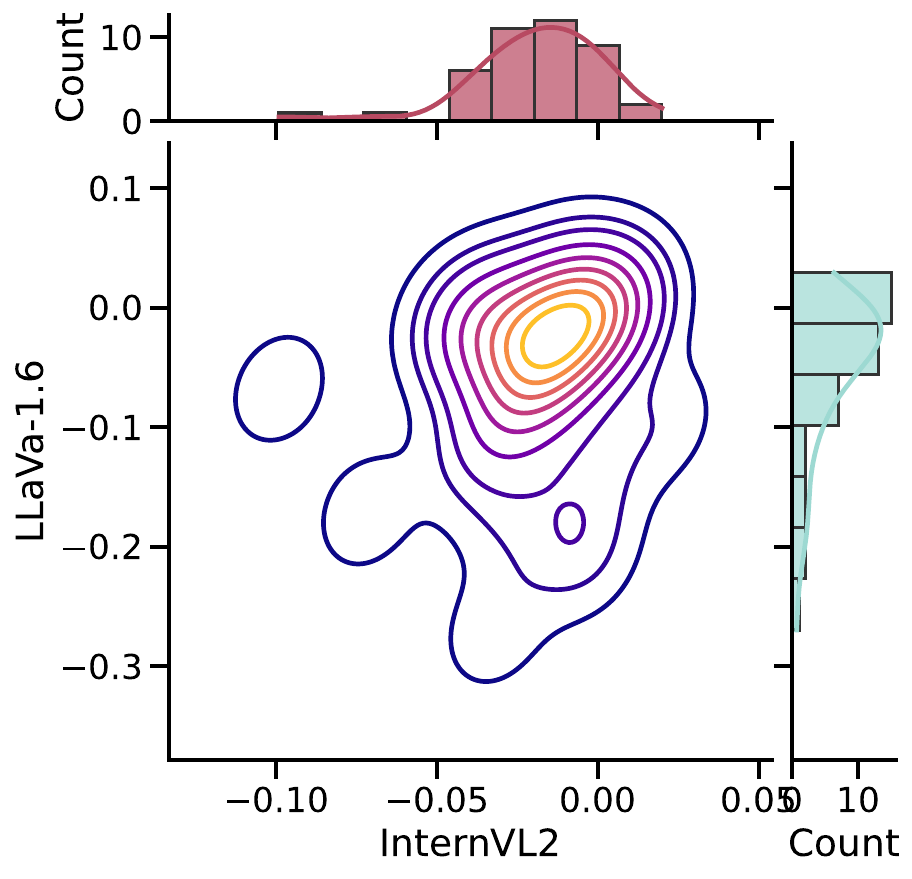}
        \caption{LLaVa-Bench}
        \label{fig:llavabench_effects_examples}
    \end{subfigure}
    \hfill
    \begin{subfigure}[b]{0.49\columnwidth}
        \centering
        \includegraphics[width=0.95\textwidth]{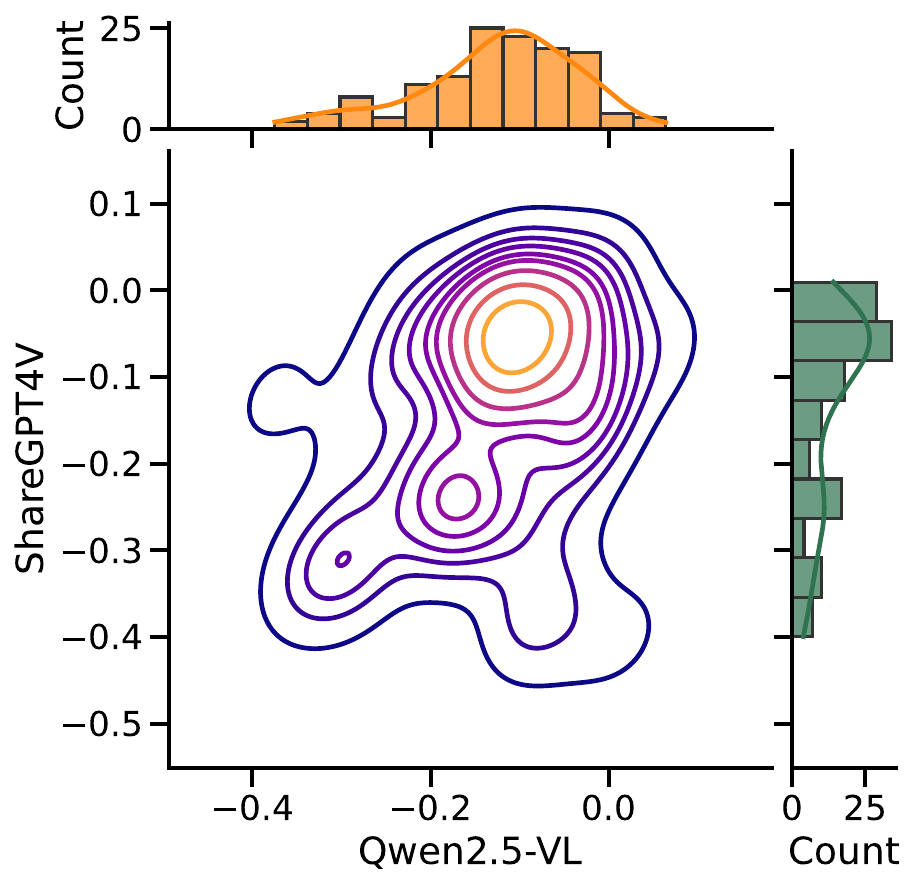}
        \caption{MMBench}
        \label{fig:mmbench_effects_examples}
    \end{subfigure}
    % new line
    \begin{subfigure}[b]{0.49\columnwidth}
        \centering
        \includegraphics[width=0.95\textwidth]{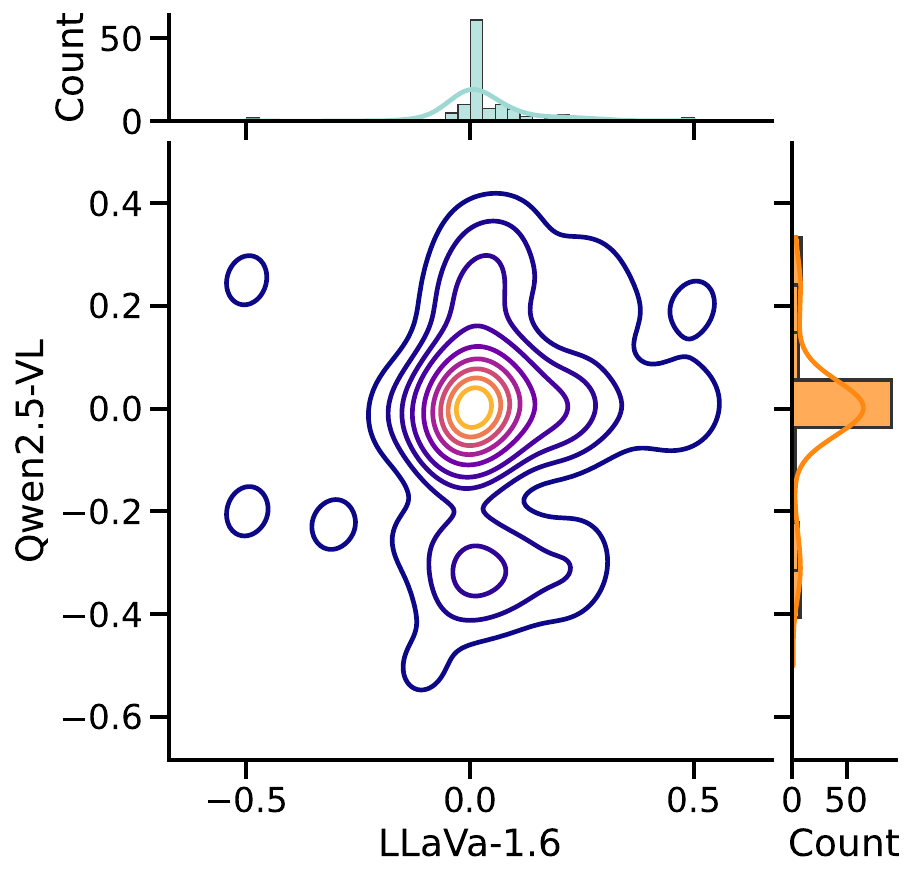}
        \caption{SEED-Bench-2}
        \label{fig:seed_effects_examples}
    \end{subfigure}
    \hfill
    \begin{subfigure}[b]{0.49\columnwidth}
        \centering
        \includegraphics[width=0.95\textwidth]{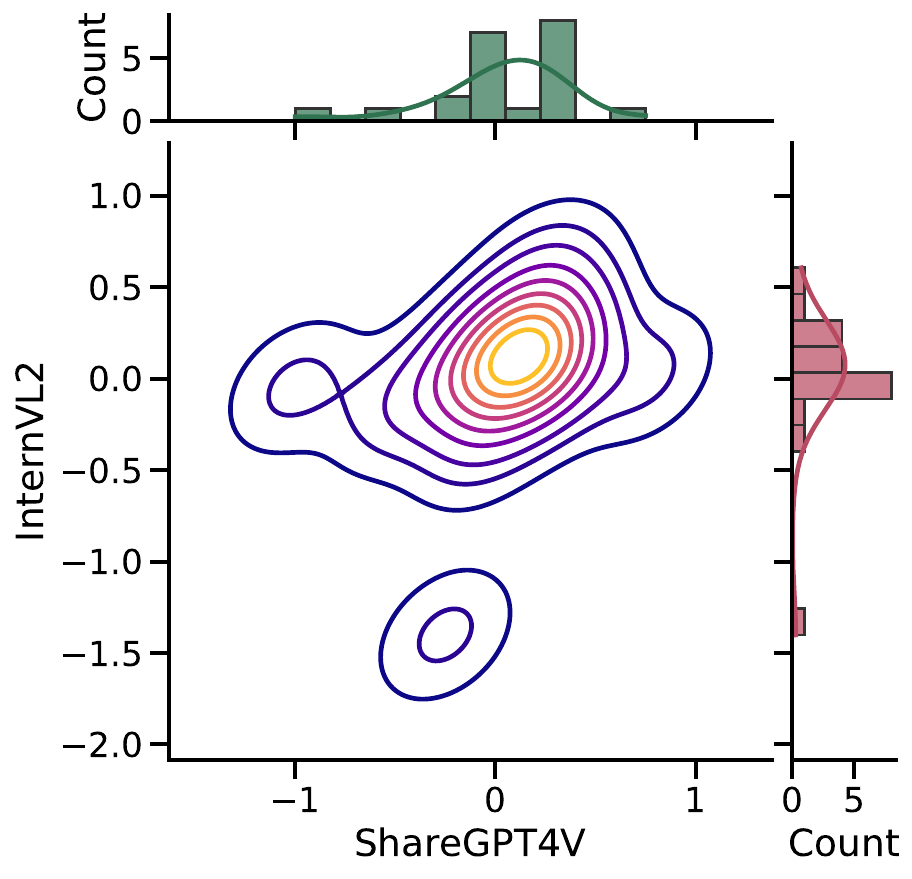}
        \caption{VQA v2}
        \label{fig:vqa_effects_examples}
    \end{subfigure}
    \caption{Example distributions of causal effects. Full comparisons for all VLMs and dataset combinations can be found in \autoref{app:distr_causal_effects}.}
    \label{fig:effects_examples}
\end{figure}

\paragraph{Causal Effect of Concepts}
We find that the causal estimates produced by \diagmethod{} concentrate generally around 0.0, with selected spots offset from it (\autoref{fig:effects_examples}).
Specifically, estimates computed on open-ended datasets span multiple concepts and may disperse across the range of output tokens (\autoref{fig:mmbench_effects_examples}).
Instead, results from multiple-choice datasets show a higher concentration of zero-valued estimates, possibly indicating that VLMs activate on concepts irrelevant to specific questions (\autoref{fig:seed_effects_examples}).
These results indicate that \diagmethod{} identifies concepts that are more likely to influence specific answers.
% This could signal that, while certain concepts may have been used by the VLMs during inference, only a subset of them might actually be driving the models towards specific answers.
Note that, despite carefully occluding images to avoid interference across visual concepts, fine-grained masking approaches \citep{Pawlowski2020DSCAM, Melistas2024BenchCounterfactualImageGen, Rasal2025DCSA} could cause VLMs to produce different counterfactual responses and, in turn, different causal estimates. \footnote{See \autoref{app:owlv2_ablation} for ablations on OWLv2, which we used to locate visual concepts from \rkshort{}s not found directly (lexical match) in the \skshort{} specifications.}
Therefore, definitive conclusions will likely require triangulation with multimodal interpretability research \citep{Liu2025MechanisticVLM}.

\paragraph{\rk{}s as Causal Graphs}
We checked the correctness of the \rkshort{} graphs and the identifiability of the estimands. First, when creating the causal model from the graph structure, we used structural refutation tests and obtained mixed results.
\rkshort{}s were not falsified when sufficient counterfactual samples could be produced ($\gtrapprox 100$ in our data), and the \rkshort{}s were sufficiently informative and passed baseline permutation tests. Otherwise, we consider the results likely inconclusive due to low test power.
Second, we identified estimands based on the assumed \rkshort{} structures. Given the \rkshort{} collected, this step did not surface issues. 
We ran additional placebo refutation tests on the estimation process, which were also inconclusive.
Nonetheless, further research is required to fully characterise the use of \rk{}s as causal graphs.
\section{Conclusion}
We introduced \diagmethod{}, a framework to diagnose the behaviours of VLMs.
By constructing specifications of expected and observed VLM behaviour, \diagmethod{} surfaces potential misalignments and quantifies the effects of the concepts used by VLMs for inference.
Beyond behaviours that are clearly aligned or misaligned, our results show that VLMs can exhibit behaviours that are more nuanced: VLMs can be selective in the concepts they use, and relate them differently from how humans do, potentially favouring broad descriptions over compositional aspects of visual inputs.
Works like \diagmethod{} provide tools for identifying VLM behaviours, supporting inquiries into composite, system-level behaviours, and potentially informing mitigation techniques for curbing unwanted behaviours.

\section*{Limitations}

\paragraph{Faithfulness of Self-explanations}
Our work relies on VLM-generated self-explanations.
Despite adding a verification layer through causal modelling, this is not equivalent to assessing the faithfulness of the self-explanations produced by the VLMs.
Even if less reliable than other explainable AI approaches in some cases \citep{Huang2023LLMSelfExplain, Randl2025ReliabilitySelfExpl}, these explanation methods are still susceptible to biases and spurious correlations. They may treat input modalities separately \citep{Kazmierczak2025ExplainabilityVLMSurvey}.
Possible approaches to mitigate this, and improve the faithfulness of VLMs' self-explanations, could use factored decomposition \citep{Radhakrishnan2023FactoredDecompositionCoT} to avoid spurious information from the original question.
Yet, to the best of our knowledge, these may still provide limited faithfulness for VLMs \citep{Li2026VisualReasonDecomposition, Lee2026VisDoT, Uppaal2026VisualFaithfulness}.
% While previous research has investigated similar decomposition strategies for VLMs, to the best of our knowledge, these may still provide limited faithfulness \citep{Li2026VisualReasonDecomposition, Lee2026VisDoT, Uppaal2026VisualFaithfulness}.
Nonetheless, as VLMs' capabilities and limitations are actively researched \cite{Liu2024SurveyHallucinationVLM, Fu2025VLMFailures}, self-explanations offer a practical means of generating \rk{} specifications for VLMs.

\paragraph{Human Dependency and Scalability}
\diagmethod{} currently employs human curation to refine candidate \skshort{}s and ensure that they are relevant to the corresponding request.
Yet, this dependency poses per-dataset scalability hurdles.
While we believe in the growing need for human curation, more scalable \sk{} curation pipelines could draw on established crowdsourcing workflows \citep{GrundeMcLaughlin2025CrowdsourcingWorkflows}.
For example, one could implement a Find-Fix-Verify workflow \citep{Bernstein2010FindFixVerify} leveraging the increasing capabilities of VLMs.
First, in the \textit{Find} step, \skshort{} candidates that are repeated, likely vague, incorrect, or likely irrelevant to the request are identified.
Then, in the \textit{Fix} step, probable issues can be assigned to another VLM or to humans, depending on their severity, following a Map-Reduce pattern \citep{Kittur2011MapReduce}.
Finally, in the \textit{Verify} step, humans verify the previously applied fixes.
The human dependency, in the average case, would therefore be restricted to a subset of the \skshort{}s, smaller than what we had contributors work on in \diagmethod{}.
A similar pipeline would require balancing precision and recall at the \textit{Find} step, to avoid missing mistakes, ensure that complex and nuanced behaviours are robustly assessed, and tune the router in the \textit{Fix} step to keep the cost of querying VLM judges and human contributions under budget constraints.

\paragraph{Unstable Instruction-following}
We found that the VLMs tested struggled with the questions in the datasets we used, sometimes returning empty strings.
This happened across different datasets and with different decoding strategies (greedy, beam search, and sampling).
% Future research should investigate how the individual components within VLMs interact with each other to identify the root cause of this phenomenon.
Similarly, the VLMs tested struggled to follow the given template when asked to structure their rationales.
We suspect this shortcoming is a by-product of the training data (predominantly conversational) and the model size.
% Additional fine-tuning could better equip these models with information-structuring capabilities.

\paragraph{Not Testing on Larger Models}
In this work, we dealt with VLMs with parameter counts in the 7B-8B range.
Larger VLMs (seem to) ``fill in more blanks'' compared to smaller ones, often resulting in stronger raw performance. Given our results and related work on VLM blindspots and idiosyncrasies between raw performance and model internals \citep{Tong2024EyesWideShut, Fu2025VLMFailures}, we are inclined to believe that similar patterns will be exhibited by larger models as well, despite what leaderboards report.
We acknowledge that applying our framework to larger models may yield conclusions different from those we obtained.

\paragraph{Not Testing on Closed-source Models}
Closed-source models, e.g., GPT-5, are intermittently updated without public notice.
While it is important to analyse these models, given their widespread use, we refrained from using them in our experiments, as doing so would undermine the reproducibility of our results and conclusions.

\section*{Ethical Considerations}

\paragraph{Perpetuating Harmful Behaviours}
While \diagmethod{} is meant to improve VLMs by helping identify misaligned behaviours, this information can also be used to further steer models in harmful directions.
A malicious actor could use \diagmethod{} to reduce appropriate behaviours and obtain a VLM that perpetuates stereotypical or hateful outputs.
To the best of our knowledge, we did not encounter any harmful model outputs in our experiments.
Bridging behavioural analysis of VLMs with red teaming and vulnerability disclosure practices could help mitigate the risk of misuse.
Practitioners working in these areas have protocols for disclosing undesirable or harmful model behaviours to model creators and labs, for architecture- and provider-specific verifications, and for determining the level of model access needed. 
We believe that (1) a more coordinated approach would be needed for reporting concerning behaviours (e.g., as in \citet{Longpre2026FlawReportingAI}) and (2) research at the intersection of interpretability and model steering is crucial to build methods and toolkits to both understand (e.g., through interpretability) and act (e.g., through RL) on model behaviours. Other solutions may only provide temporary, model-specific patches (e.g., through developer-prompt instructions). 

\paragraph{Participant Safety and Consent}
Our data collection to define \sk{} specifications was conducted with ethics approval from our institution (ID: 4696).
All participants from Prolific provided explicit consent before participation and could withdraw and have their contribution deleted without explanation or penalty.
The datasets and classes in our experiments were screened to exclude potentially offensive data points.
To the best of our knowledge, this was achieved, as no participants reported problems during the task.

\paragraph{Data Privacy}
All responses were anonymised, and no personally identifiable information was collected.
Data will be released in processed form to ensure participant privacy.

\begin{comment}
\section*{Acknowledgments}
This work used the Dutch national e-infrastructure with the support of the
SURF Cooperative using grant no. EINF-10580.
\end{comment}

\bibliography{references}

% Appendices after references (ARR guidelines)
\appendix
\section{Primer on Causal Inference} \label{sec:app_causality}
% \jie{not sure if readers can understand this. good to add to the appendix, very briefly, a causality 101 text.}

Causal inference is the ``discipline that considers the assumptions, study designs, and estimation strategies that allow researchers to draw causal conclusions based on data'' \cite{Hill2015CausalOverview}.
Specifically, causal inferences aim to estimate the effect of one variable (e.g., input feature) on another (e.g., model prediction) \cite{PearlMackenzie2018Causality}.
Randomised Control Trials (RCT) are commonly adopted to estimate such effects.
In RCTs, two groups (e.g., groups of people) are administered a \textit{treatment} to assess its \textit{effect} -- compared to not administering such a treatment -- on the outcome of an experiment.
RCTs are carried out abiding to the \textit{ceteris paribus} principle (i.e., ``all other things being equal"): the treatment is the only variable within a trial.

Unfortunately, carrying out RCTs and obtaining \textit{counterfactual} data that answers to ``what-if'' questions (e.g., `What if we had not administered a medicine?') can be expensive, infeasible, or unethical to be carried out.
When access to counterfactual data is not possible, Causal Discovery can be leveraged to find causal relationships and form a so-called causal graph.

In the following, we describe these two concepts.

\subsection{Causal Graphs}
Causal graphs are modelling tools that show the relations and effects a set of (possibly) interrelated independent variables may have on the final outcome $Y$ (i.e., the dependant variable) through a directed acyclic graph (DAG) \citep{PearlMackenzie2018Causality}.
Causal graphs are particularly useful for studying different interventions (i.e., the treatments) without performing a real trial.
Here, we briefly introduce terms that describe the role that individual variables can take within a causal graph.

\noindent \textbf{Confounder}: a variable $Z$ which has an effect on other variables, e.g., $X$ and $Y$, such that $X$ and $Y$ show correlation despite not being causally related.
Confounders need to be accounted for when studying the relationship between $X$ and $Y$.
\begin{equation}
    X \leftarrow Z \rightarrow Y   
\end{equation}

\noindent \textbf{Mediator}: a variable $M$ causally related to an independent variable $X$ causing an \textit{indirect} effect on the outcome $Y$.
\begin{equation}
    X \rightarrow M \rightarrow Y   
\end{equation}

\noindent \textbf{Collider}: a variable $C$ that is influenced by two or more variables $X$ and $Y$.
\begin{equation}
    X \rightarrow C \leftarrow Y  
\end{equation}

When dealing with such variables, one is usually interested in estimating the \textit{Average Treatment Effect} (ATE). That is, the average difference between administering a treatment and not administering it across the population being studied.

\subsection{Causal Discovery}
Generally, creating a causal graph requires domain expertise and is disentangled from the experimental hypotheses.
Still, researchers have proposed techniques to infer causal structures from observational data by relying on statistical independence tests: These fall under the umbrella of Causal Discovery.
Here, we report the main algorithms available. Refer to \cite{Glymour2019CDMethods} for a complete categorisation of causal discovery methods.
First, constraint-based causal discovery algorithms, like Peter-Clark (PC) and Fast Causal Inference (FCI) \cite{Spirtes2000Causation}, are based on a complete and undirected graph including all the variables involved and use statistical (conditional) independence tests to prune the edges.
Second, score-based models like Greedy Equivalence Score (GES) \cite{Chickering2002Optimal} start with an empty graph and add edges as long as the scoring function (e.g., Bayesian Information Criterion) increases. Then, edge removal operations are applied to test whether the score can be further increased.
Finally, pairwise approaches aim to define causal relations between any two variables by evaluating the fitness of the data to an additive noise model \citep{Hoyer2008CD}, by bidirectionally comparing the standard deviation of the rescaled values of one variable with respect to the other one in the pair \citep{Fonollosa2016CD}, or by leveraging asymmetries \citep{Daniusis2012CausalRelations}.

Despite the breadth of techniques available, it is important to note statistical dependence does not imply causal dependence, i.e., the causal graphs that can be obtained might not be complete nor unique \cite{Zhang2011CausalIndTesting, Weinberger2018CausalCoincidences}.
Thus, incorporating domain-specific or task-specific assumptions is fundamental to have satisfactory graphs \cite{Hyvarinen1999NonLinear, Zhang2015Estimation}.
\section{Crowdsourcing Setup} \label{app:crowd_tasks}
Here, we outline our crowdsourcing setup for curating candidate \sk{} obtained with scene graph generation (\autoref{sec:method_sk}).
We recruited 520 workers (avg. wage: 8 GBP/h) on Prolific from English-speaking countries, with an approval rate of $\geq 95\%$.
This study received ethics approval from our institution (ID: 4696).

Before carrying out the tasks, participants were required to provide explicit consent (which they could revoke at any time during the study without penalty) and to complete a brief tutorial explaining the task and the requirements.

\subsection{Verification Task}
\begin{figure*}[t]
    \centering
    \includegraphics[width=0.9\textwidth]{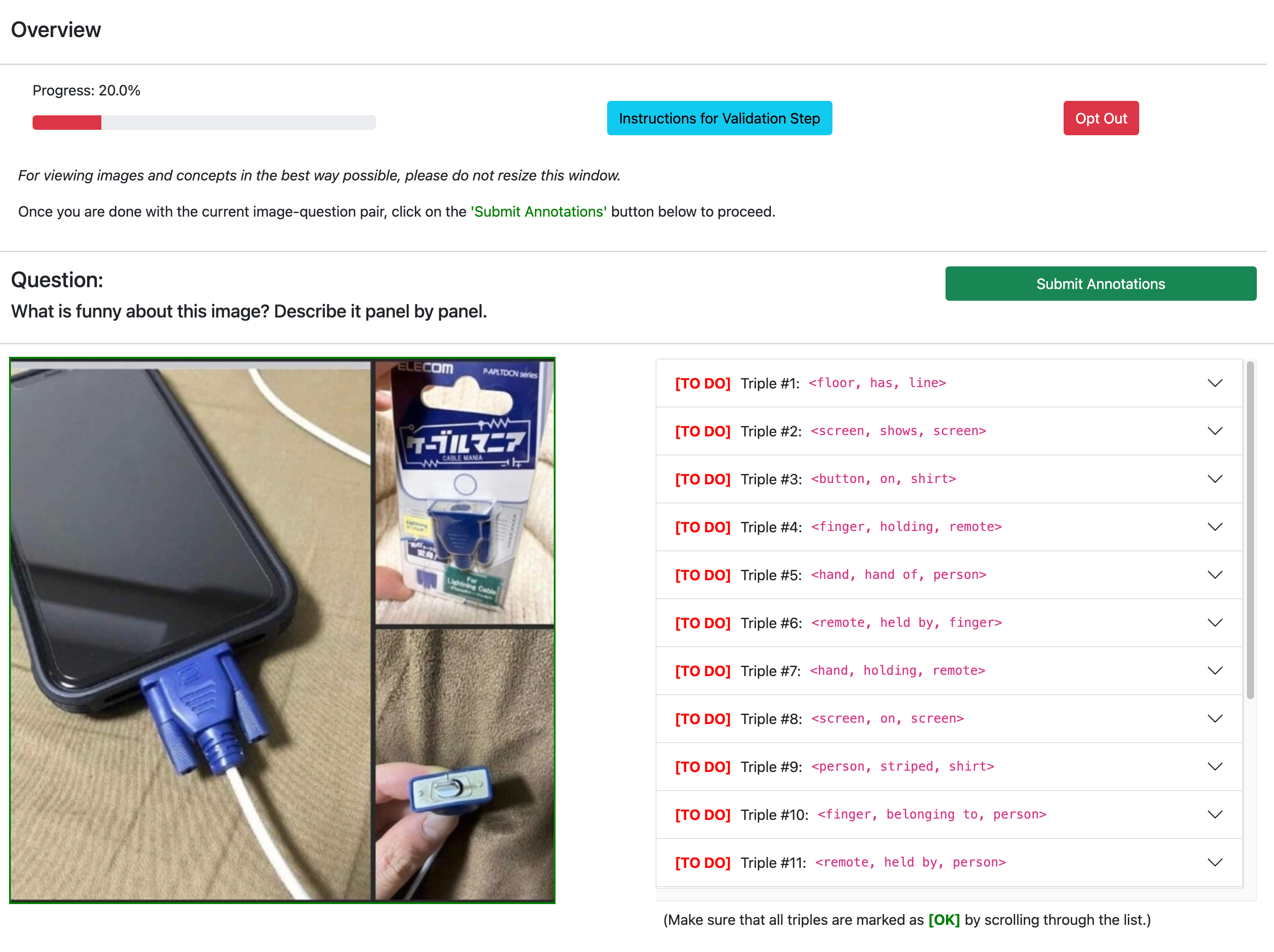}
    \caption{Verification Task}
    \label{fig:verification_task}
\end{figure*}

\begin{figure}[t]
    \centering
    \includegraphics[width=0.8\linewidth]{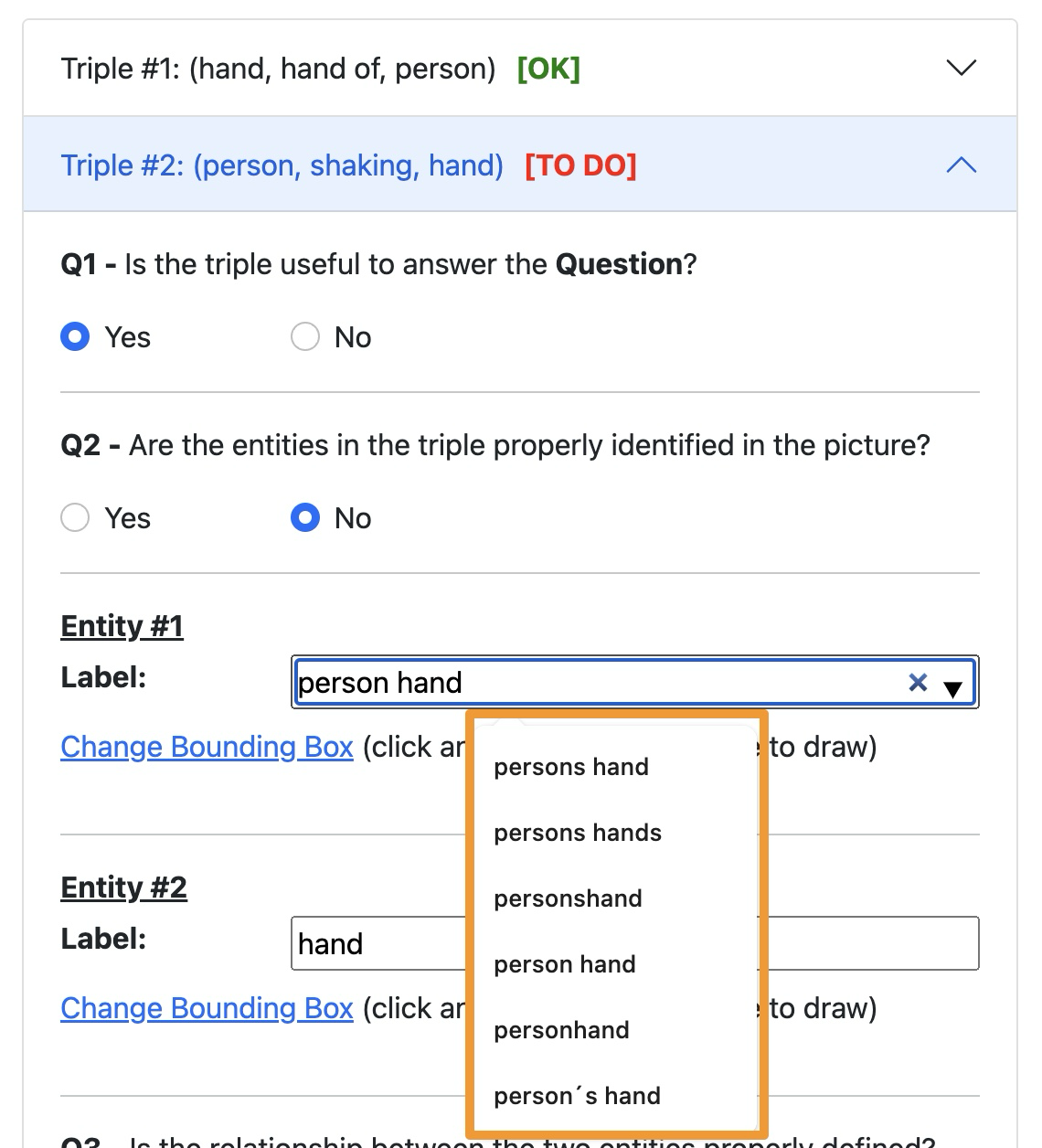}
    \caption{Form to edit triple data with the auto-complete suggestions visible (highlighted).}
    \label{fig:example_editing}
\end{figure}

For this task, participants check the correctness and the relevance of the triplets extracted in the SGG step with respect to the input image and text (\autoref{fig:verification_task}).
Participants can confirm or edit concept labels, concept locations, and relationship labels.
They are assisted in copy-editing the triplets with a lightweight auto-complete mechanism based on the concept and relationship labels used by IETrans.
Note that participants were not limited to these labels and could provide new ones (\autoref{fig:example_editing}).

\subsection{Expansion Task}
\begin{figure*}[t]
    \centering
    \includegraphics[width=0.9\textwidth]{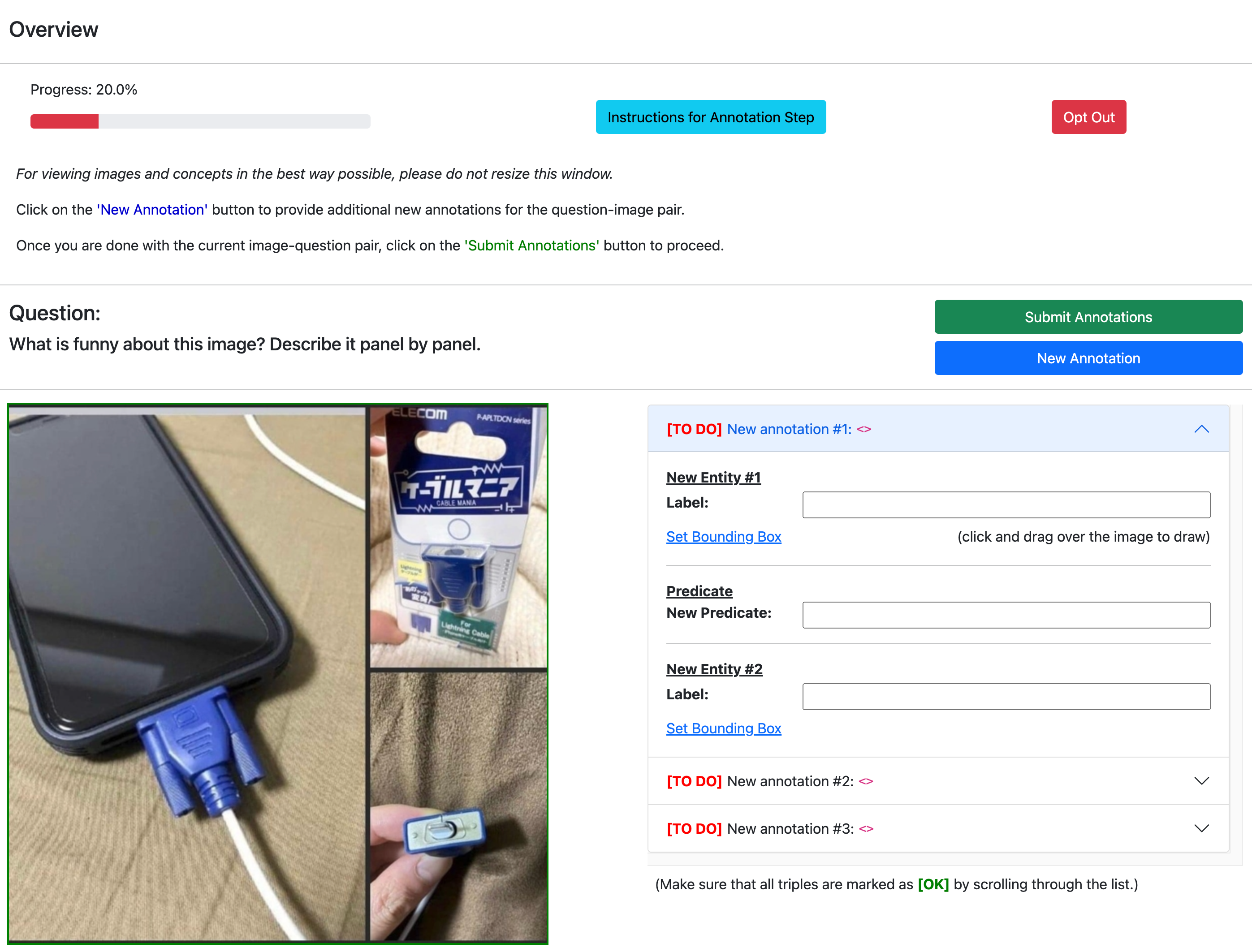}
    \caption{Verification Task}
    \label{fig:annotation_task}
\end{figure*}

In this task, we seek to obtain additional triplets that were not identified in the SGG step.
We asked participants to indicate these relevant but missing triplets to extend the previously verified \skshort{} specifications, if needed (\autoref{fig:annotation_task}).
Here, human annotations comprise concept labels, bounding boxes, and relationship labels.
\section{Prompt Templates} \label{app:prompt_templates}
In this section, we outline the prompt templates designed to obtain responses and concomitant self-explanations from VLMs (\autoref{sec:method_rk}).
These result from several refinements using ChatGPT and manual checking to ensure their correctness, adherence to the task, and lack of hallucinations.

We provide the prompt templates we used for
\begin{itemize}[leftmargin=*]
    \item Collecting model responses: \autoref{app_fig:self_expl_response}
    \item Producing rationales: \autoref{app_fig:self_expl_rationale}
    \item Structuring rationales: \autoref{app_fig:self_expl_struct}
\end{itemize}

% == == == ==

\begin{figure*}[t] % The placement specifier can be [htbp]
\centering
\scalebox{0.95}{
\begin{tcolorbox}[title=Collecting model responses]
\small\noindent
\textbf{LLaVa-Bench} (same for all models) \\

\{\{ \textit{image and question from dataset} \}\}
\tcbline

\textbf{MMBench} (same for all models) \\

\texttt{Provide a concise and descriptive caption for this image.} \\
\{\{ \textit{image and question from dataset} \}\}
\tcbline

\textbf{SEED-Bench 2} (for InternVL2, LLaVa-1.6, and Qwen2.5-VL) \\

\texttt{Based on the provided image, analyze the given question and choose the most accurate option from the listed alternatives. Your response should consist solely of the letter corresponding to the correct answer (A, B, C, or D), without any additional explanation.} \\
\{\{ \textit{image and question from dataset} \}\} \\

\textbf{SEED-Bench 2} (for ShareGPT4V)\\

\texttt{Based on the provided image, analyze the given question and choose the most accurate option from the listed alternatives. Your response \underline{must} consist solely of \underline{the answer you believe is correct}, without any additional explanation.} \\
\{\{ \textit{image and question from dataset} \}\}
\tcbline

\textbf{VQA v2} (same for all models) \\

\texttt{Based on the given image, answer the question using only one word. Provide no additional text or explanation.} \\
\{\{ \textit{image and question from dataset} \}\}

\end{tcolorbox}
}
\captionof{figure}{Prompt templates for getting model responses. Model-specific changes are \underline{underlined}.}\label{app_fig:self_expl_response}
\end{figure*}

% == == == ==

\begin{figure*}[t] % The placement specifier can be [htbp]
\centering
\scalebox{0.95}{
\begin{tcolorbox}[title=Producing rationales]
\small\noindent

\texttt{Review the given answer and carefully analyze the image. Identify and describe the key visual elements and concepts in the image that directly support your reasoning and answer. For each point, clearly explain how that visual element contributes to the answer, and ensure every point is directly linked to the reasoning behind the answer. \\
Present your rationale in a structured, bullet-point format with the following guidelines: \\
- Each bullet point must begin with the * character. \\
- Each point should clearly explain the connection between a specific visual element and the given answer. \\
- Avoid general descriptions of the image; focus on visual details that are relevant to the answer. \\
- Keep each point concise and ensure it directly addresses the reasoning behind the answer.}
\end{tcolorbox}
}
\captionof{figure}{Prompt templates for generating model rationales. Same for all models.}\label{app_fig:self_expl_rationale}
\end{figure*}

% == == == ==

\begin{figure*}[t] % The placement specifier can be [htbp]
\centering
\scalebox{0.95}{
\begin{tcolorbox}[title=Structuring rationales]
\small\noindent
\textbf{InternVL2 and Qwen2.5-VL} \\

\texttt{Carefully analyze each point in the list above and extract distinct pairs of entities and their relationships. For each point, identify exactly two distinct entities at a time, along with a detailed and context-specific relationship between them (e.g., action, spatial position, interaction, causality, part-whole relationship, etc.). For each extraction, you must use the following format exactly: (Entity: [entity\_1], Relationship: [relationship], Entity: [entity\_2]). \\
Additional guidelines: \\
- Both entities must always be present: Ensure that each extraction contains two distinct entities in the format provided. \\
- Do not include references to the image itself (e.g., `image,' `picture,' `photo') as an entity. \\
- Avoid overly simplistic relationships (e.g., `located in' or `visible in'). Focus on specific interactions or connections, such as `forms,' `comprises,' `is part of,' `affects,' or `is near.' \\
- Focus on distinct entity pairs, avoiding duplicates across all points. \\
- If a point contains multiple entities or relationships, break them down into separate pairs. \\
- Entities can refer to physical objects, people, animals, abstract concepts, etc. \\
- Relationships should reflect more complex interactions or spatial/temporal arrangements. \\
- Ensure the relationship clearly reflects how the entities are connected in context. Consider relationships like causality, containment, or part-whole structure. \\
- Prioritize clarity and contextual relevance in your extractions. \\
- Each extraction must contain two distinct entities; avoid incomplete extractions with only one entity. \\
Example: \\
- Original sentence: `A dog is playing with a ball near the tree.' \\
- Output: \\
  - (Entity: dog, Relationship: playing with, Entity: ball) \\
  - (Entity: dog, Relationship: near, Entity: tree) \\
  - (Entity: tree, Relationship: part of, Entity: landscape) }
\tcbline

\textbf{LLaVa-1.6 and ShareGPT4V} \\

\texttt{Analyze the provided content and extract distinct pairs of entities and their relationships using the guidelines below: \\
1. Output Format: Each extraction must follow this format: * (Entity: [entity\_1], Relationship: [relationship], Entity: [entity\_2]) \\
2. Entity Guidelines: \\
   - Extract two distinct entities for each pair. \\
   - Exclude references to the image itself (e.g., ``image," ``picture"). \\
   - Entities can be physical objects, people, animals, or abstract concepts. \\
3. Relationship Guidelines: \\
   - Identify meaningful, context-specific relationships (e.g., forms, comprises, is part of, affects, is near). \\
   - Relationships should describe interactions, connections, or arrangements between the entities, such as causality, spatial/temporal proximity, or part-whole structures. \\
4. Rules for Clarity and Consistency: \\
   - Each extraction must include exactly two entities and one relationship. Avoid incomplete or ambiguous pairs. \\
   - Break down complex points into multiple pairs if needed. \\
   - Avoid duplicates across extractions. \\
5. Relevance and Accuracy: \\
   - Ensure extractions are contextually accurate and relevant. \\
   - Prioritize clear, precise descriptions of entity relationships. \\
Example: \\
 - Original Sentence: ``A dog is playing with a ball near the tree." \\
 - Extractions: \\
   - (Entity: dog, Relationship: playing with, Entity: ball) \\
   - (Entity: dog, Relationship: near, Entity: tree)}

\end{tcolorbox}
}
\captionof{figure}{Prompt templates for structuring VLM rationales. Changes are mostly related to how instructions are organised and their clarity.}\label{app_fig:self_expl_struct}
\end{figure*}
\section{Additional Implementation Details} \label{app:impl_details}

\subsection{BERT Score Configuration}
We followed the indications of the \citet{Zhang2020BERTScore} and used DeBERTa-XLarge to compute the BERT Score measures, given its better correlation with human evaluators.
We report the hashcode to show the settings we used for BERT Score \cite{Zhang2020BERTScore}.
\newline\newline
\noindent \texttt{microsoft/deberta-xlarge-mnli\_L40\\\_no-idf\_version=0.3.12(hug\_trans=4.30.0)}

\subsection{Double ML Setup} \label{app:dml_setup}
For our DML in \diagmethod{}, we use the implementations from \texttt{DoWhy} and \texttt{EconML}.
These implement the \texttt{DML2} variant from \citep{Chernozhukov2018DoubleML}.
These packages handle sample splitting and cross-fitting, which are crucial in the DML framework.
We use the default number of 2 sample splits.
Finally, we perform 5-fold cross-validation to estimate the causal effects from the residuals.
For the two regressors, we use gradient-boosted trees to estimate $\hat{f}(Z)$ and $\hat{g}(Z)$. We rely on the \texttt{scikit-learn} implementation \texttt{GradientBoostingRegressor}, with the following parameters:

\begin{itemize}[leftmargin=*]
    \item Number of estimators $= 100$
    \item Tree depth $= 3$
    \item Learning rate $= 0.1$
\end{itemize}

\begin{table}[t]
\centering
\resizebox{\columnwidth}{!}{%
\begin{tabular}{lll}
\toprule
\textbf{Dataset}                    & \textbf{$J_{concepts}$} & \textbf{$J_{relations}$} \\ \midrule
LLaVa-Bench                         & 0.81                  & 0.82                   \\
MMBench -- \textit{Image Scene}              & 0.41                  & 0.91                   \\
MMBench -- \textit{Image Topic}              & 0.99                  & 0.90                   \\
SEED-Bench 2 -- \textit{Scene Understanding} & 0.85                  & 0.88                   \\
SEED-Bench 2 -- \textit{Visual Reasoning}    & 0.99                  & 0.94                   \\
VQAv2 -- \textit{How many people are...}     & 0.47                  & 0.89                   \\
VQAv2 -- \textit{What is the person...}      & 0.45                  & 0.87                   \\ \bottomrule
\end{tabular}%
}
\caption{Average Jaccard distance, for each dataset, measured between the sets of unique concepts and relationships obtained with and without \texttt{USE\_VISION} enabled.}
\label{tab:ietrans_abl_jaccard}
\end{table}
\section{Additional Results} \label{app:additional_results}

\subsection{Ablating IETrans} \label{app:ietrans_ablation}
We verified that IETrans appropriately relies on visual inputs rather than on spurious linguistic correlations stemming from transferring data from general predicate labels to specific ones (i.e., internal transfer) or relabelling (i.e., external transfer). See the original work by \citep{Zhang2022IETrans} for the details on IETrans data transfer.

Concretely, we compared IETrans runs when \texttt{PREDICT\_USE\_VISION} is set to \texttt{False} (ablation) and \texttt{True} (our experiments).
For this, we measure the difference in the number of unique concepts detected by IETrans (\autoref{tab:ietrans_abl_counts}), as well as the Jaccard distance between the unique concept and relationship labels extracted in the two settings (\autoref{tab:ietrans_abl_jaccard}).
We see that once the visual signal is turned off, IETrans identifies fewer concepts and assigns very different labels to any given sample across the datasets considered.

\begin{table*}[t]
\centering
\resizebox{\textwidth}{!}{%
\begin{tabular}{p{0.25\linewidth}cccccccc}
\toprule
\multirow{2}[3]{*}{\textbf{Dataset}} & \multicolumn{2}{c}{\textbf{Concept Detected}} & \multicolumn{2}{c}{\textbf{Relations Detected}} \\
\cmidrule(lr){2-3} \cmidrule(lr){4-5}
               & \multicolumn{1}{c}{w/ \texttt{USE\_VISION}} & \multicolumn{1}{c}{w/o \texttt{USE\_VISION}} & \multicolumn{1}{c}{w/ \texttt{USE\_VISION}} & \multicolumn{1}{c}{w/o \texttt{USE\_VISION}} \\ \midrule
LLaVa-Bench  & 746  & 689 ($\downarrow$ 7.64\%) & 720 & 720 (-) \\
MMBench -- \textit{Image Scene} & 7905 & 6393 ($\downarrow$ 19.13\%) & 7650 & 7650 (-) \\
MMBench -- \textit{Image Topic} & 1980 & 179 ($\downarrow$ 90.96\%) & 1920 & 1920 (-) \\
SEED-Bench 2 -- \textit{Scene Understanding} & 100903 & 71883 ($\downarrow$ 28.76\%) & 94740 & 94740 (-) \\
SEED-Bench 2 -- \textit{Visual Reasoning} & 10715 & 1314 ($\downarrow$ 87.74\%) & 9930 & 9930 (-) \\
VQAv2 -- \textit{How many people are...}& 58894 & 40604 ($\downarrow$ 31.06\%) & 55140 & 55110 ($\downarrow$ 0.05\%) \\ 
VQAv2 -- \textit{What is the person...}& 26136 & 19292 ($\downarrow$ 26,19\%) & 25530 & 25530 (-) \\
\bottomrule
\end{tabular}%
}
\caption{IETrans concepts and relationships extracted when the vision signal is used (w/ \texttt{USE\_VISION}) compared to when it is not (w/o \texttt{USE\_VISION}).}
\label{tab:ietrans_abl_counts}
\end{table*}

\subsection{Behaviour Threshold Sensitivity} \label{app:th_sensitivity}
\begin{figure*}[t]
    \centering
    \begin{subfigure}[b]{0.49\textwidth}
        \centering
        \includegraphics[width=0.95\textwidth]{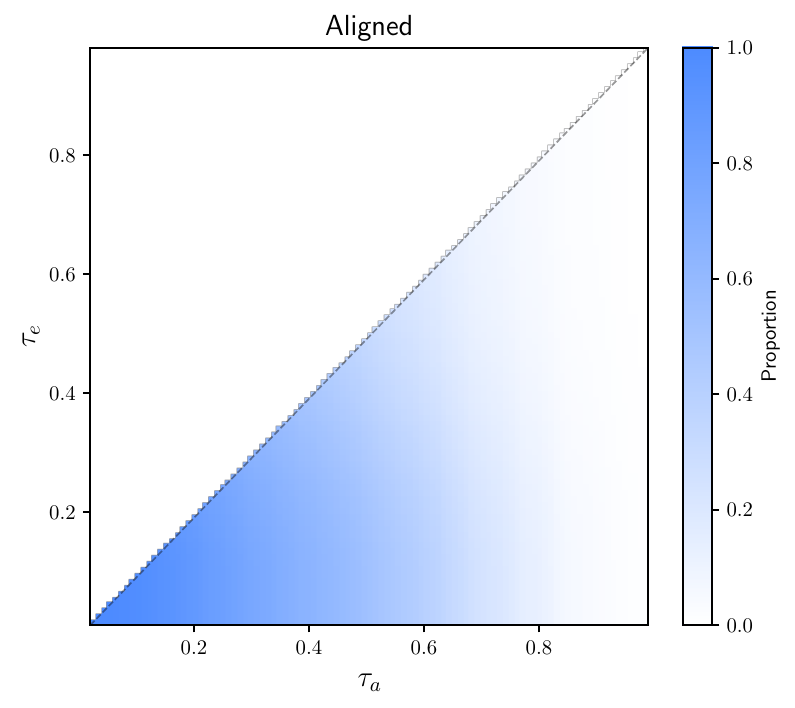}
        \label{fig:heatmap_aligned_100}
    \end{subfigure}
    \hfill
    \begin{subfigure}[b]{0.49\textwidth}
        \centering
        \includegraphics[width=0.95\textwidth]{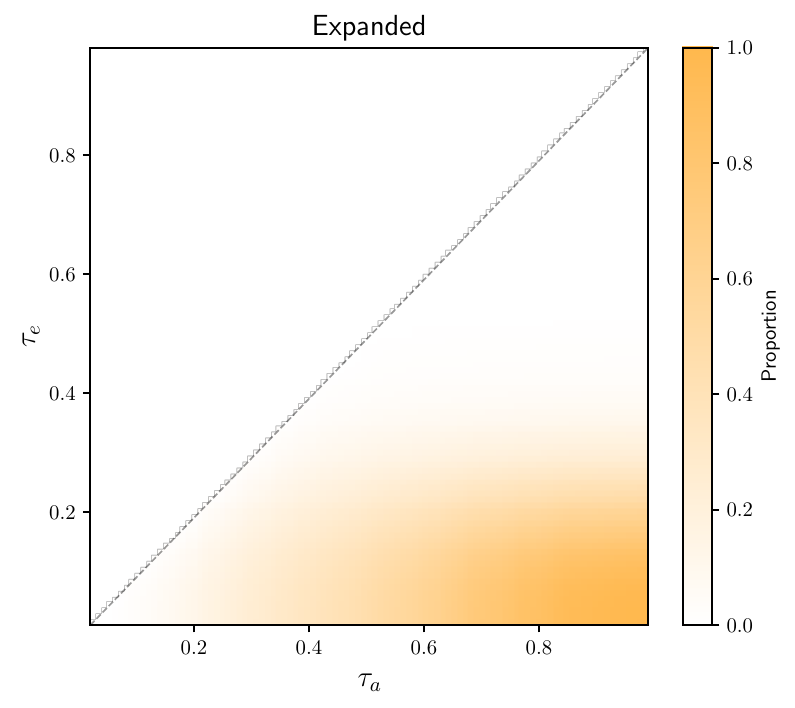}
        \label{fig:heatmap_expanded_100}
    \end{subfigure}
    % new line
    \begin{subfigure}[b]{0.49\textwidth}
        \centering
        \includegraphics[width=0.95\textwidth]{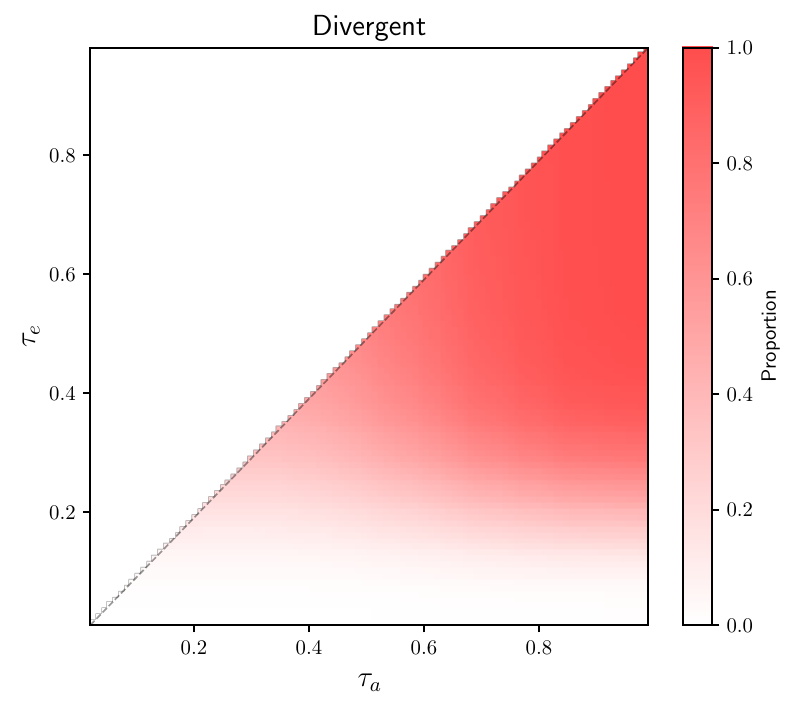}
        \label{fig:heatmap_divergent_100}
    \end{subfigure}
    \hfill
    \begin{subfigure}[b]{0.49\textwidth}
        \centering
        \includegraphics[width=0.95\textwidth]{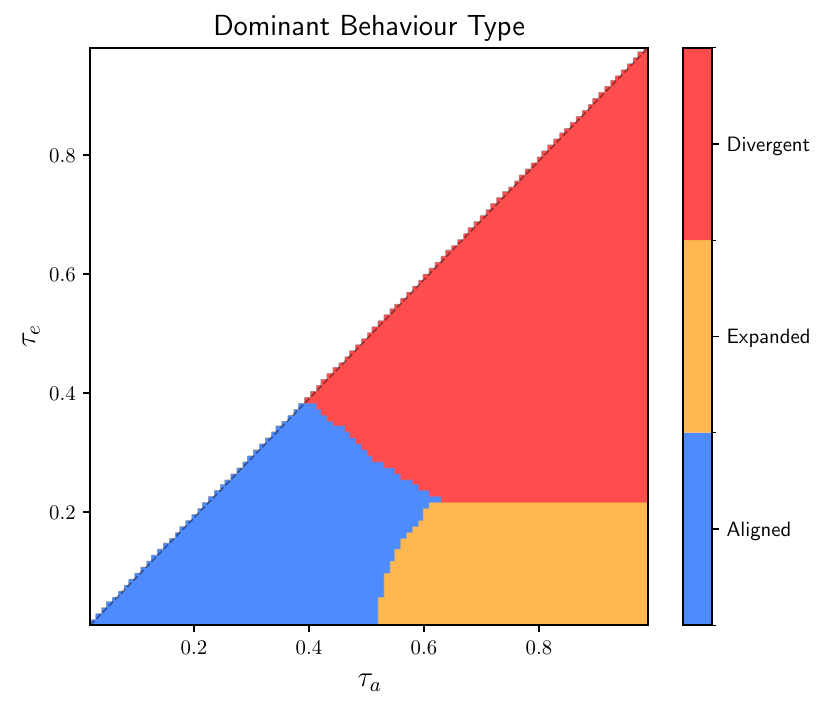}
        \label{fig:dominant_bucket_100}
    \end{subfigure}
    \caption{Proportions of samples classified as Aligned, Expanded, or Divergent, and dominant behaviour type as $\tau_a$ and $\tau_e$ vary.}
    \label{fig:sensitivity_proportions}
\end{figure*}

In addition to \autoref{fig:sensitivity_curves}, we report in \autoref{fig:sensitivity_proportions} the proportions of samples classified as Aligned, Expanded, or Divergent as $\tau_a$ and $\tau_e$ vary.

\subsection{Ablating OWLv2} \label{app:owlv2_ablation}
\begin{sidewaysfigure*}
    \centering
    \includegraphics[width=0.87\textwidth]{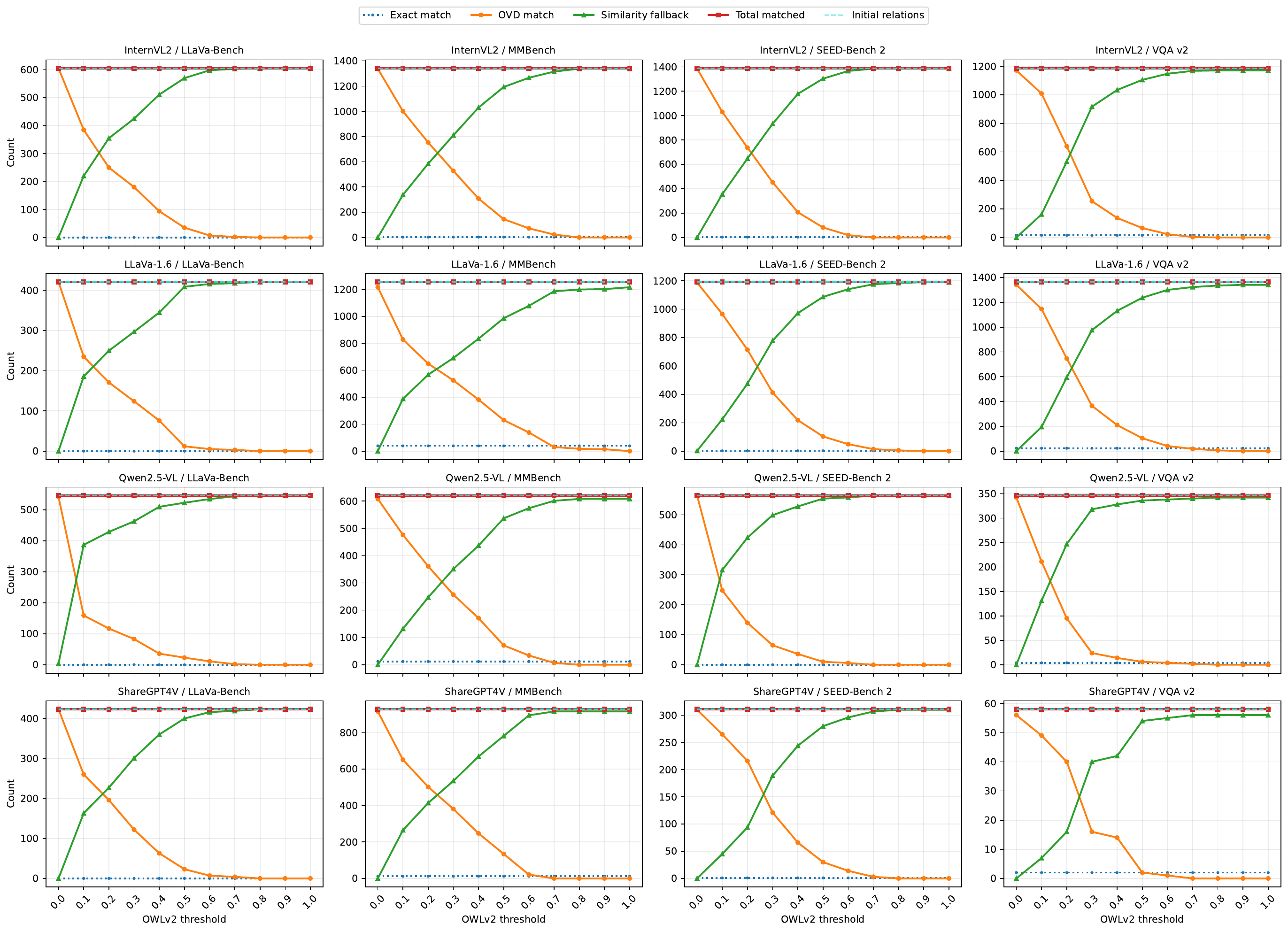}
    \caption{Results from $\tau_{\textnormal{OWLv2}}$ sweep.}
    \label{fig:owlv2_out_th_sweep}
\end{sidewaysfigure*}

\begin{sidewaysfigure*}
    \centering
    \includegraphics[width=0.87\textwidth]{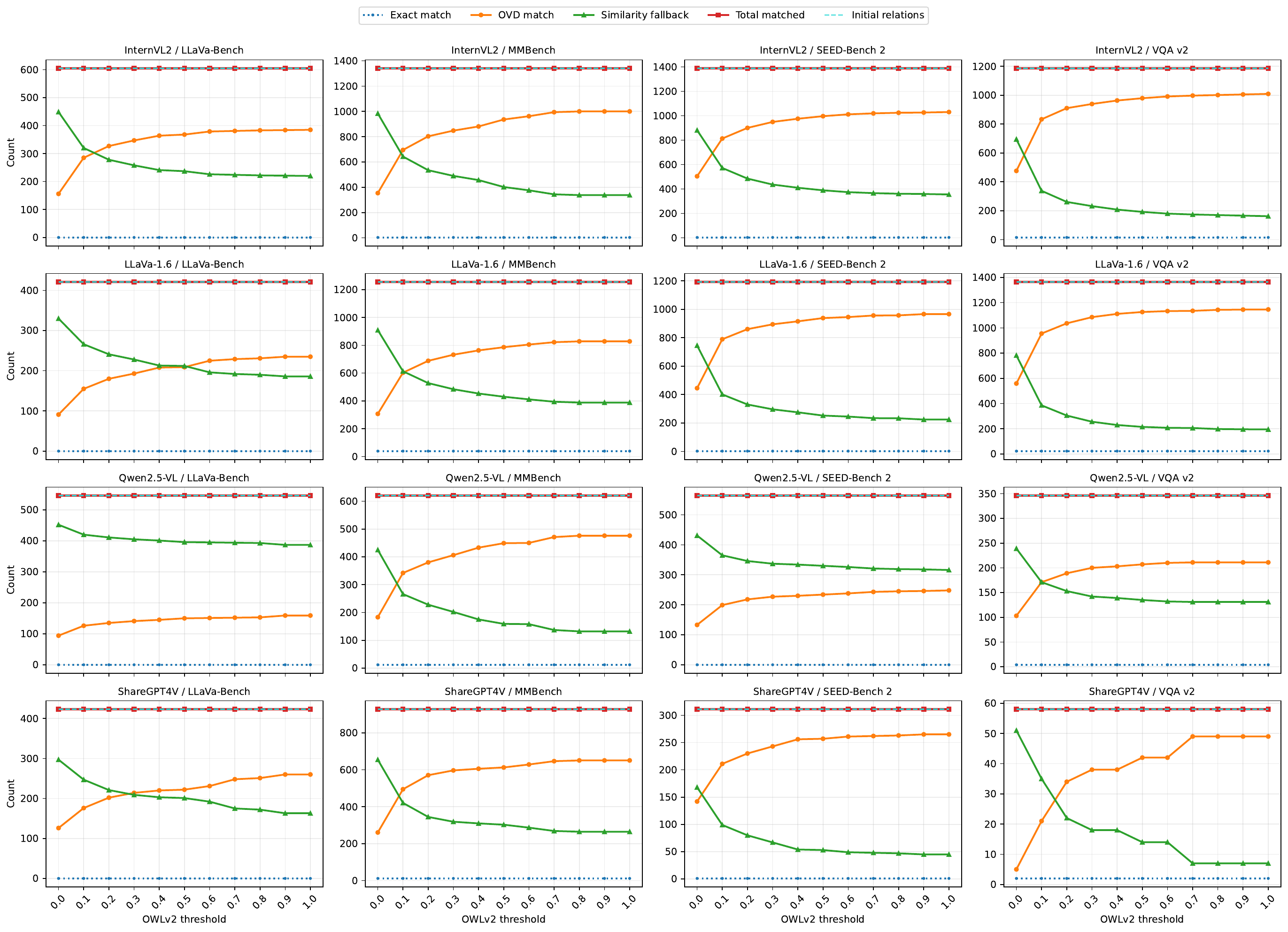}
    \caption{Results from $\tau_{\textnormal{nms}}$ sweep.}
    \label{fig:owlv2_nms_sweep}
\end{sidewaysfigure*}

We applied OWLv2 \citep{Minderer2023OWLv2} zero-shot with the default parametrisation to help us match the \skshort{} and \rkshort{} specifications and derive bounding boxes for visual concepts in the self-explained \rkshort{}.
We considered the top-1 bounding box from OWLv2, using the suggested post-processing threshold $\tau_{\textnormal{OWLv2}} = 0.1$ and suppression threshold $\tau_{\textnormal{nms}} = 0.3$ used to collate overlapping bounding boxes.
\autoref{tab:intermediate_matching} shows the intermediate percentages of concepts for which bounding boxes were found throughout that step of \diagmethod{}.
To understand how OWLv2 behaves within \diagmethod{}, we carried out two ablations:
\begin{itemize}
    \item We swept $\tau_{\textnormal{OWLv2}}$ within its range [$0.0, 1.0$].
    As we increase $\tau_{\textnormal{OWLv2}}$, we can see two phenomena (see \autoref{fig:owlv2_out_th_sweep}).
    First, higher $\tau_{\textnormal{OWLv2}}$ values cause more and more \rk{} triplets to be erroneously passed onto the semantic fallback step, i.e., the third and last step of our \skshort{}-\rkshort{} matching setup. This is expected as we match \skshort{} and \rkshort{} incrementally.
    Second, the \skshort{}-\rkshort{} may decrease to a point where OWLv2 matches fewer triplets compared to exact, lexicon-based matching (e.g., for LLaVa-1.6 on MMBench), therefore missing out on relevant triplets. This may be due to poor calibration of OWLv2's own scores.
    \item We swept the internal Non Maximum Suppression (NMS) threshold $\tau_{\textnormal{nms}}$, while fixing $\tau_{\textnormal{OWLv2}} = 0.1$. The results (\autoref{fig:owlv2_nms_sweep}) show that, for the most part, the overall number of \skshort{}-\rkshort{} matches starts to plateau for $\tau_{\textnormal{nms}} > 0.3$. High $\tau_{\textnormal{nms}}$ may also lead to bounding boxes being aggregated too coarsely, thereby losing information. These results confirm that, in our specific case, the default $\tau_{\textnormal{nms}} = 0.3$ is an effective sweet spot.
\end{itemize}

% \usepackage{booktabs}
% \usepackage{graphicx}
\begin{table*}[t]
\centering
\resizebox{\textwidth}{!}{%
\begin{tabular}{@{}llllllll@{}}
\toprule
\textbf{VLM} &
  \textbf{Dataset} &
  \textbf{Initial \rkshort{}} &
  \textbf{Exact Matching} &
  \textbf{OWLv2 Matches} &
  \textbf{Semantic Fallback} &
  \textbf{Matched} &
  \textbf{Not Matched} \\ \midrule
\textbf{InternVL2}     & LLaVa-Bench  & 605   & 0      & 385     & 220     & 605     & 0      \\
                       & MMBench      & 1341  & 3      & 1000    & 338     & 1341    & 0      \\
                       & SEED-Bench 2 & 1388  & 3      & 1030    & 355     & 1388    & 0      \\
                       & VQA v2       & 1186  & 15     & 1009    & 162     & 1186    & 0      \\ \hline
\textbf{LLaVa-1.6}     & LLaVa-Bench  & 421   & 0      & 235     & 186     & 421     & 0      \\
                       & MMBench      & 1255  & 39     & 828     & 388     & 1255    & 0      \\
                       & SEED-Bench 2 & 1192  & 2      & 966     & 224     & 1192    & 0      \\
                       & VQA v2       & 1364  & 23     & 1146    & 195     & 1364    & 0      \\ \hline
\textbf{Qwen2.5-VL}    & LLaVa-Bench  & 546   & 0      & 159     & 387     & 546     & 0      \\
                       & MMBench      & 620   & 12     & 476     & 132     & 620     & 0      \\
                       & SEED-Bench 2 & 564   & 0      & 248     & 316     & 564     & 0      \\
                       & VQA v2       & 346   & 4      & 211     & 131     & 346     & 0      \\ \hline
\textbf{ShareGPT4V}    & LLaVa-Bench  & 423   & 0      & 260     & 163     & 423     & 0      \\
                       & MMBench      & 928   & 12     & 651     & 265     & 928     & 0      \\
                       & SEED-Bench 2 & 311   & 1      & 265     & 35      & 301     & 10     \\
                       & VQA v2       & 58    & 2      & 49      & 7       & 58      & 0      \\ \hline
\textbf{Total Matched} &              & 12548 & 116    & 8918    & 3504    & 12538   & 10     \\
\textbf{\% Matched}    &              &       & 0.92\% & 71.07\% & 27.29\% & 99.92\% & 0.08\% \\ \bottomrule
\end{tabular}%
}
\caption{Intermediate \rkshort{}-\skshort{} matching counts and percentages.}
\label{tab:intermediate_matching}
\end{table*}

\subsection{Mutual Information Bootstrapping} \label{app:mi_bootstrapping}
Given our small sample size, we use bootstrapping to examine MI itself and assess potential biases and overestimates.
Concretely, we bootstrap p-values and $95\%$ confidence intervals and perform permutation checks.
We perform $1000$ resampling draws and correct our MI estimates as
\begin{equation}
    \textnormal{MI}_{corrected} = \textnormal{MI}_{observed} - \textnormal{MI}_{permutation}
\end{equation}

\noindent We report the results in \autoref{tab:effectiveness_acc_full}.

\begin{figure*}[t]
    \centering
    \begin{subfigure}[b]{0.49\textwidth}
        \centering
        \includegraphics[width=0.95\textwidth]{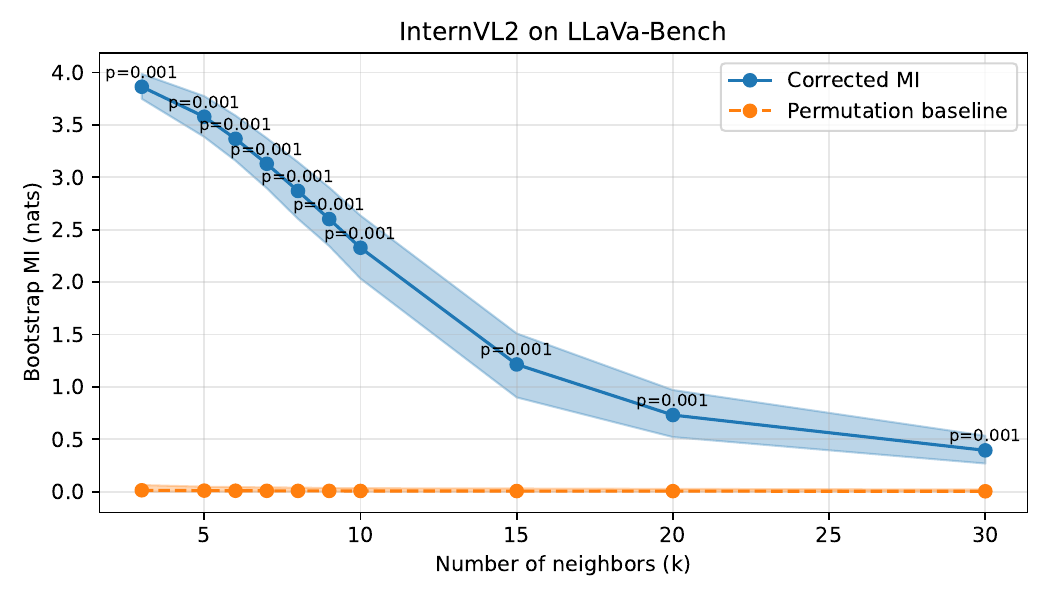}
        % \caption{F1.}
        \label{fig:mi_internvl2_llava-bench}
    \end{subfigure}
    \hfill
    \begin{subfigure}[b]{0.49\textwidth}
        \centering
        \includegraphics[width=0.95\textwidth]{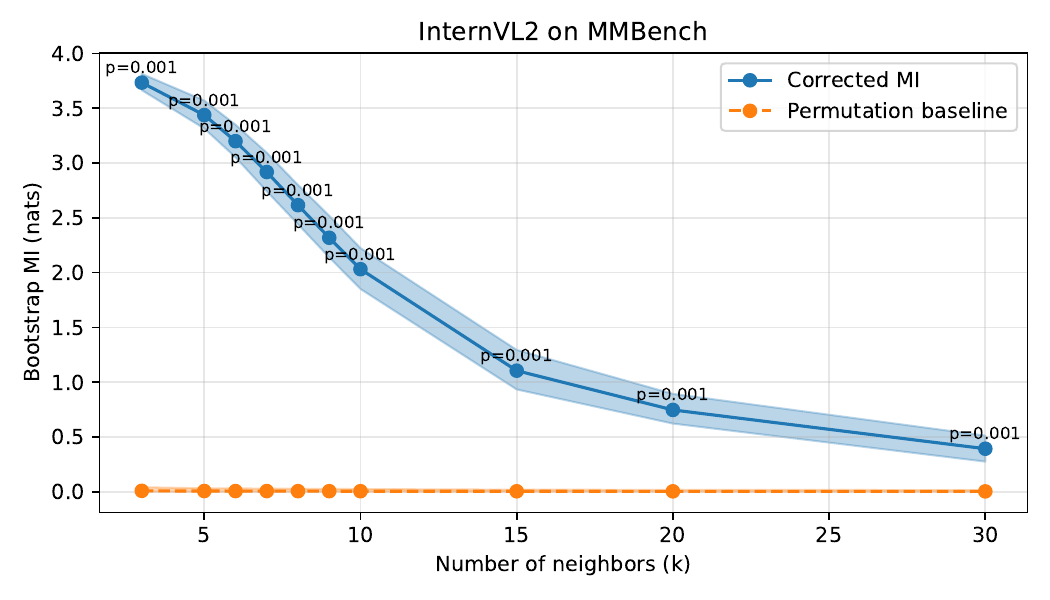}
        % \caption{F1.}
        \label{fig:mi_internvl2_mmbench}
    \end{subfigure}
    % new line
    \begin{subfigure}[b]{0.49\textwidth}
        \centering
        \includegraphics[width=0.95\textwidth]{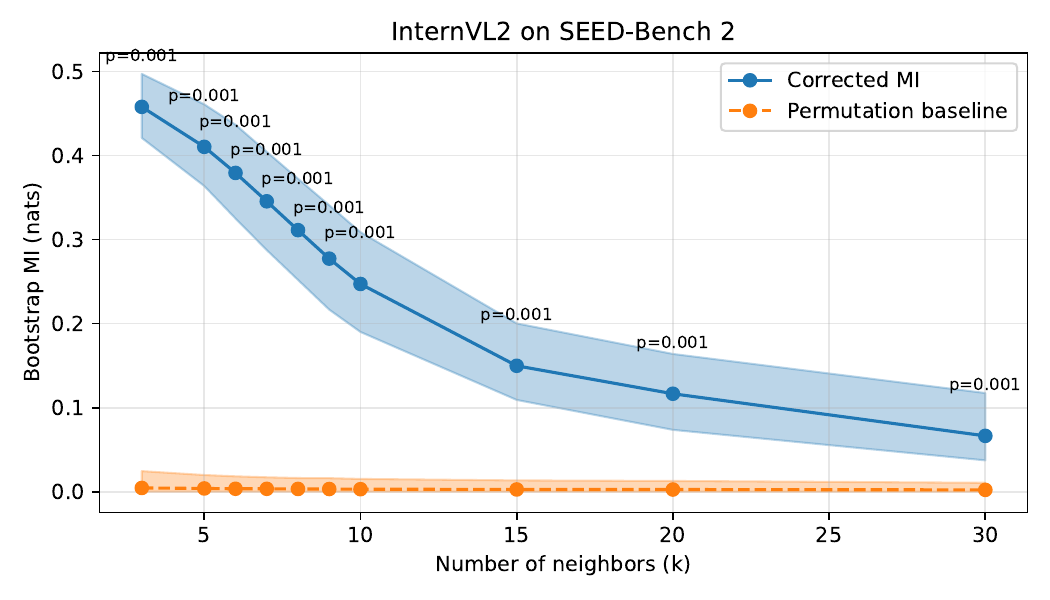}
        % \caption{F1.}
        \label{fig:mi_internvl2_seed}
    \end{subfigure}
    \hfill
    \begin{subfigure}[b]{0.49\textwidth}
        \centering
        \includegraphics[width=0.95\textwidth]{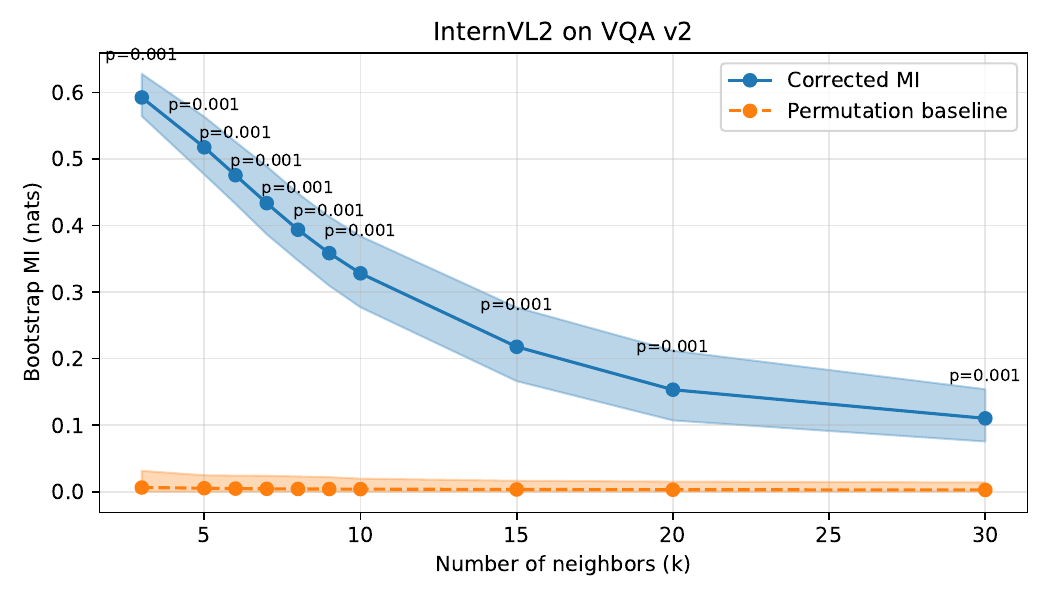}
        % \caption{F1.}
        \label{fig:mi_internvl2_vqav2}
    \end{subfigure}
    \caption{Bootstrapped Mutual Information measures InternVL2 \citep{Chen2024InternVL}.}
    \label{fig:mi_internvl2}
\end{figure*}

\begin{figure*}[t]
    \centering
    \begin{subfigure}[b]{0.49\textwidth}
        \centering
        \includegraphics[width=0.95\textwidth]{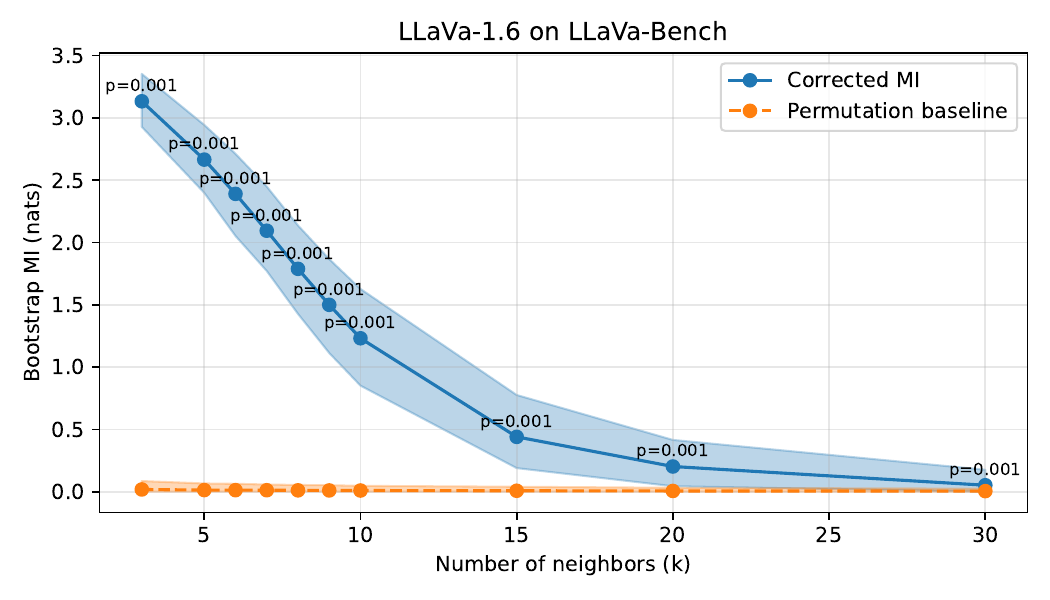}
        % \caption{F1.}
        \label{fig:mi_llava-1.6_llava-bench}
    \end{subfigure}
    \hfill
    \begin{subfigure}[b]{0.49\textwidth}
        \centering
        \includegraphics[width=0.95\textwidth]{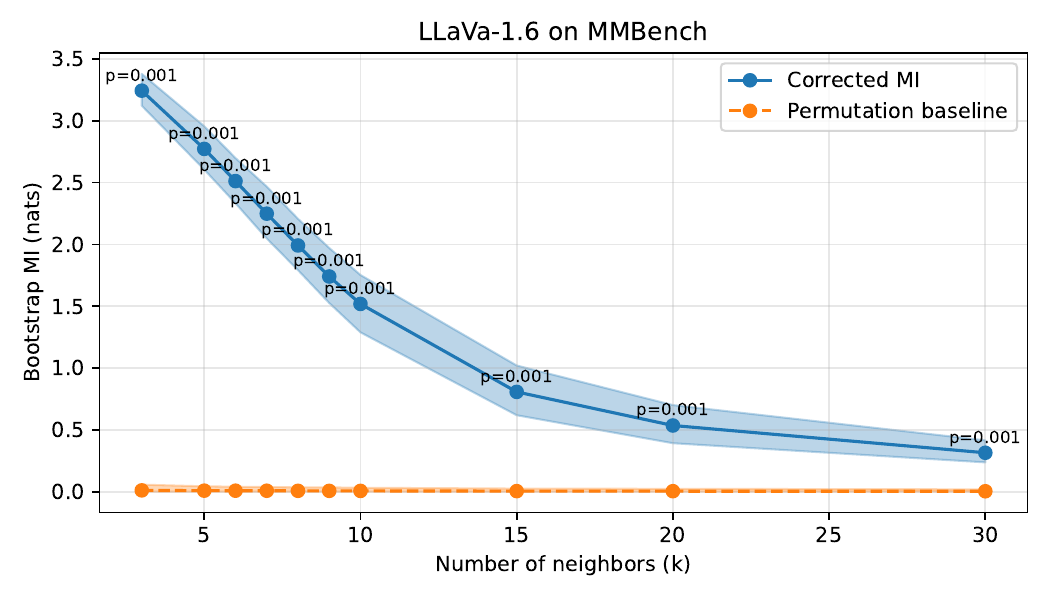}
        % \caption{F1.}
        \label{fig:mi_llava-1.6_mmbench}
    \end{subfigure}
    % Goes to second line
    \begin{subfigure}[b]{0.49\textwidth}
        \centering
        \includegraphics[width=0.95\textwidth]{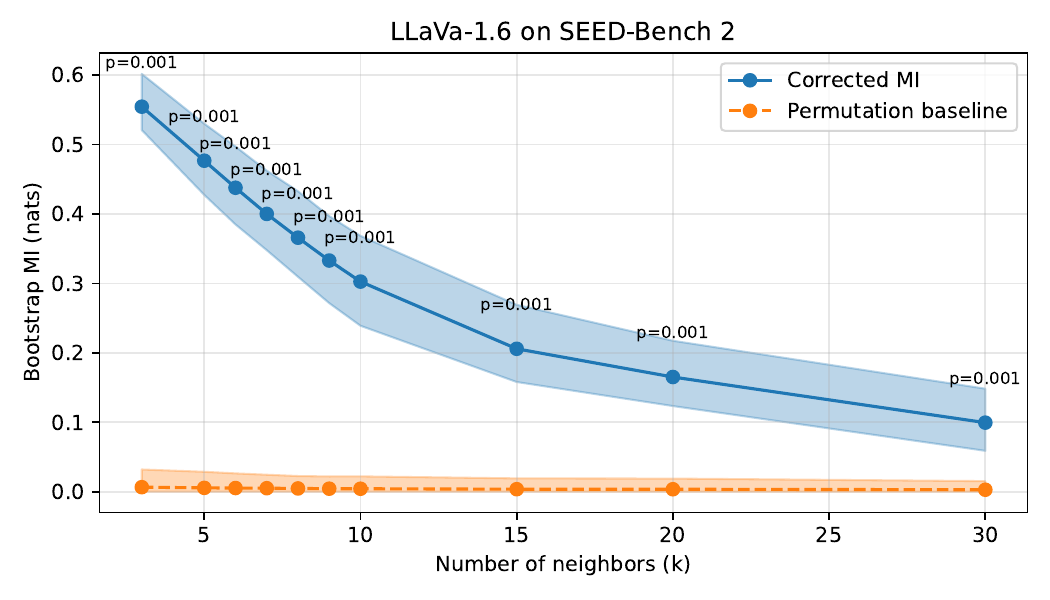}
        % \caption{F1.}
        \label{fig:mi_llava-1.6_seed}
    \end{subfigure}
    \hfill
    \begin{subfigure}[b]{0.49\textwidth}
        \centering
        \includegraphics[width=0.95\textwidth]{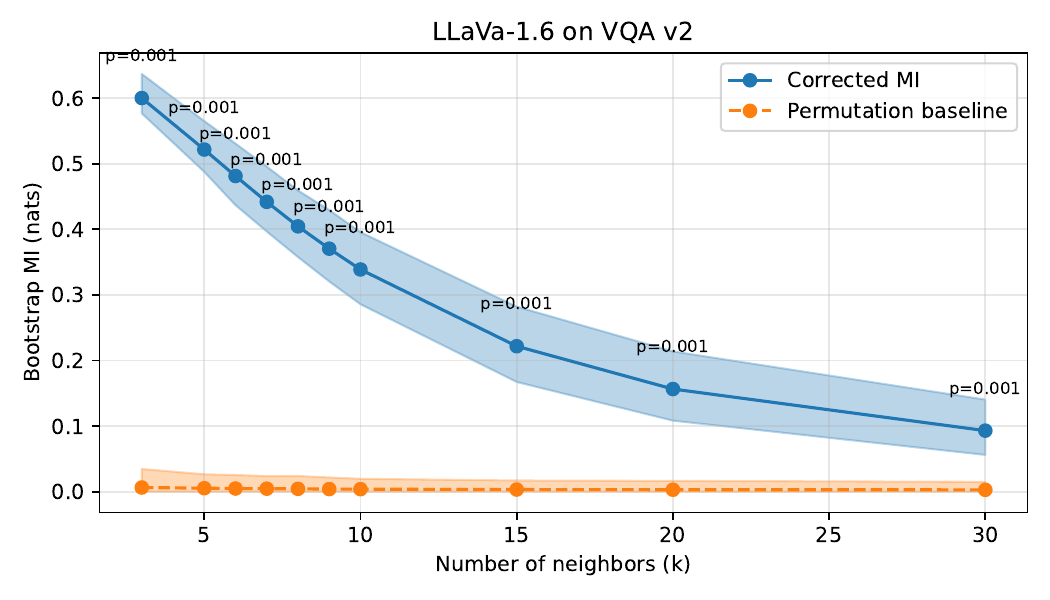}
        % \caption{F1.}
        \label{fig:mi_llava-1.6_vqav2}
    \end{subfigure}
    \caption{Bootstrapped Mutual Information measures LLaVa-1.6 \citep{liu2024llavanext}.}
    \label{fig:mi_llava-1.6}
\end{figure*}

\begin{figure*}[t]
    \centering
    \begin{subfigure}[b]{0.49\textwidth}
        \centering
        \includegraphics[width=0.95\textwidth]{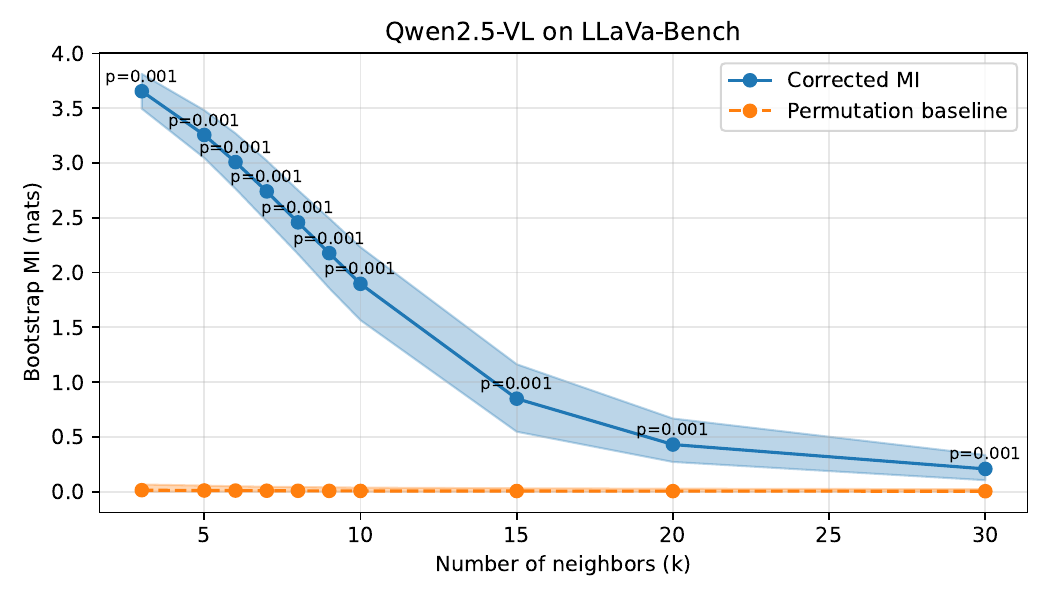}
        % \caption{F1.}
        \label{fig:mi_qwen2_5_vl_llava-bench}
    \end{subfigure}
    \hfill
    \begin{subfigure}[b]{0.49\textwidth}
        \centering
        \includegraphics[width=0.95\textwidth]{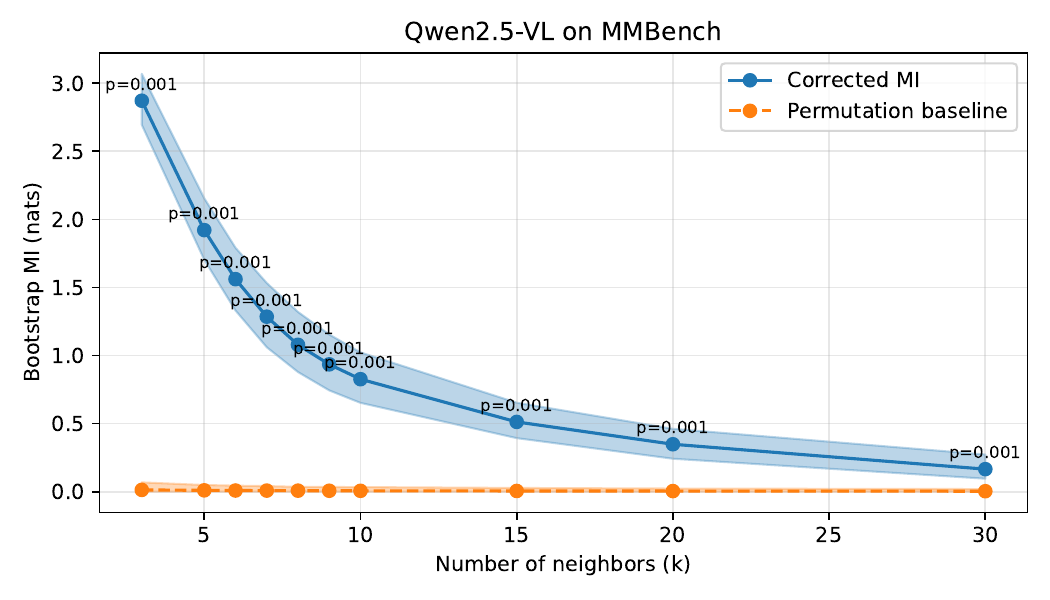}
        % \caption{F1.}
        \label{fig:mi_qwen2_5_vl_mmbench}
    \end{subfigure}
    % Goes to second line
    \begin{subfigure}[b]{0.49\textwidth}
        \centering
        \includegraphics[width=0.95\textwidth]{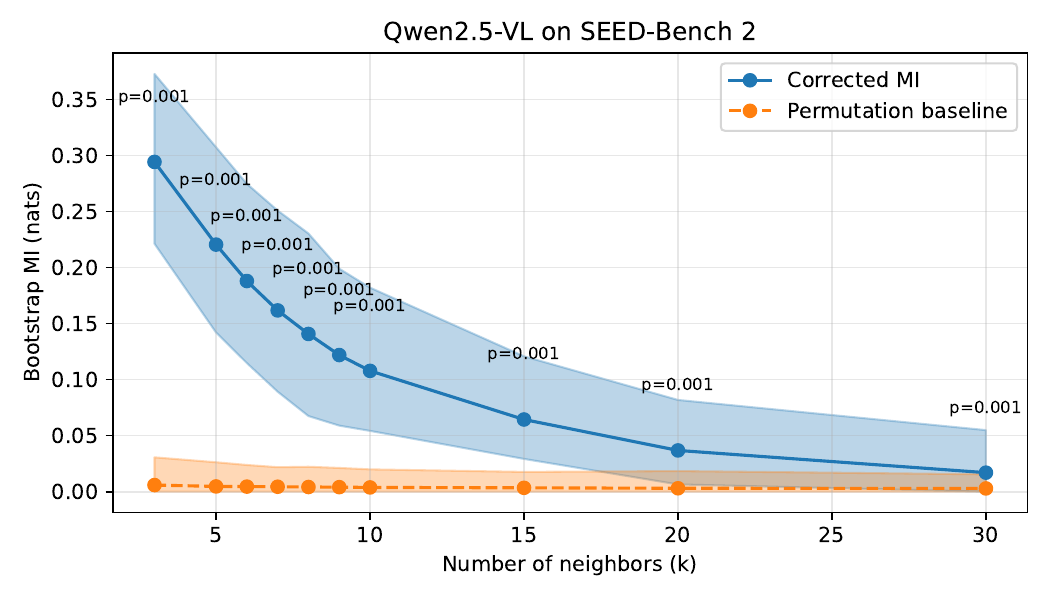}
        % \caption{F1.}
        \label{fig:mi_qwen2_5_vl_seed}
    \end{subfigure}
    \hfill
    \begin{subfigure}[b]{0.49\textwidth}
        \centering
        \includegraphics[width=0.95\textwidth]{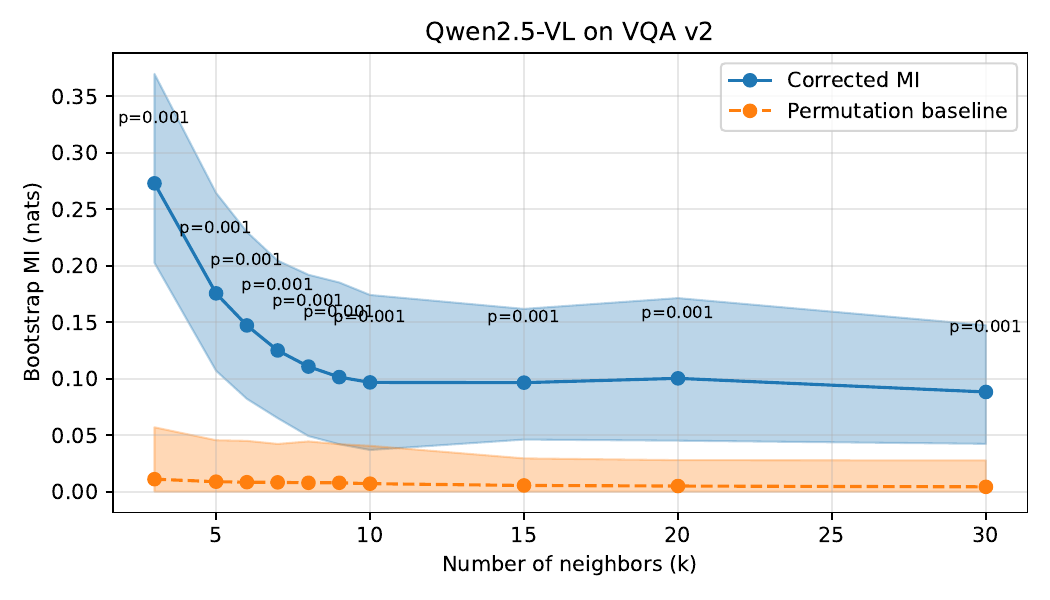}
        % \caption{F1.}
        \label{fig:mi_qwen2_5_vl_vqav2}
    \end{subfigure}
    \caption{Bootstrapped Mutual Information measures Qwen2.5-VL \citep{Bai2025Qwen25VL}.}
    \label{fig:mi_qwen2_5_vl}
\end{figure*}

\begin{figure*}[t]
    \centering
    \begin{subfigure}[b]{0.49\textwidth}
        \centering
        \includegraphics[width=0.95\textwidth]{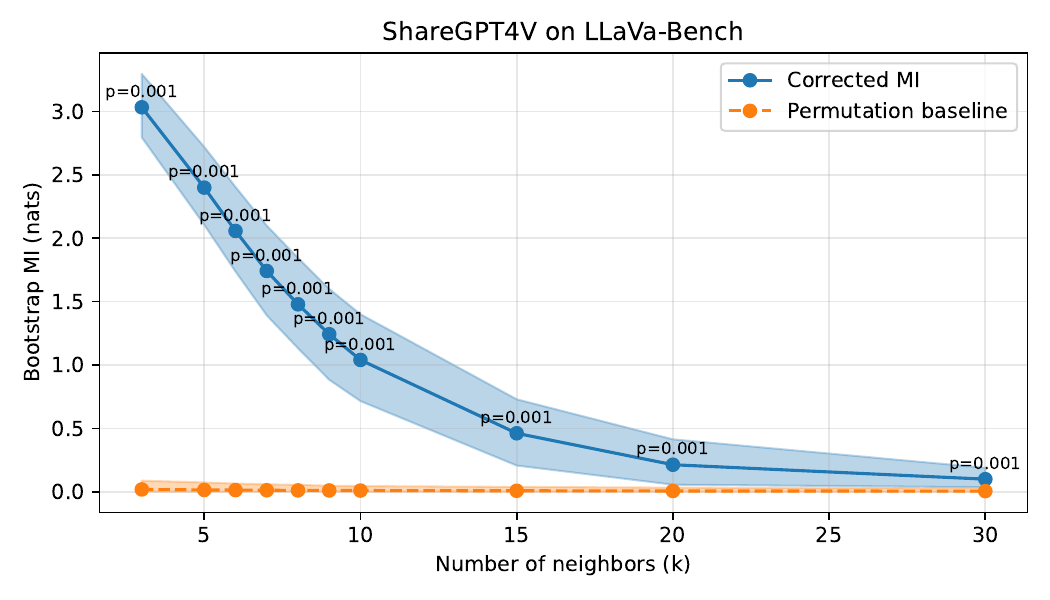}
        % \caption{F1.}
        \label{fig:mi_sharegpt4v_llava-bench}
    \end{subfigure}
    \hfill
    \begin{subfigure}[b]{0.49\textwidth}
        \centering
        \includegraphics[width=0.95\textwidth]{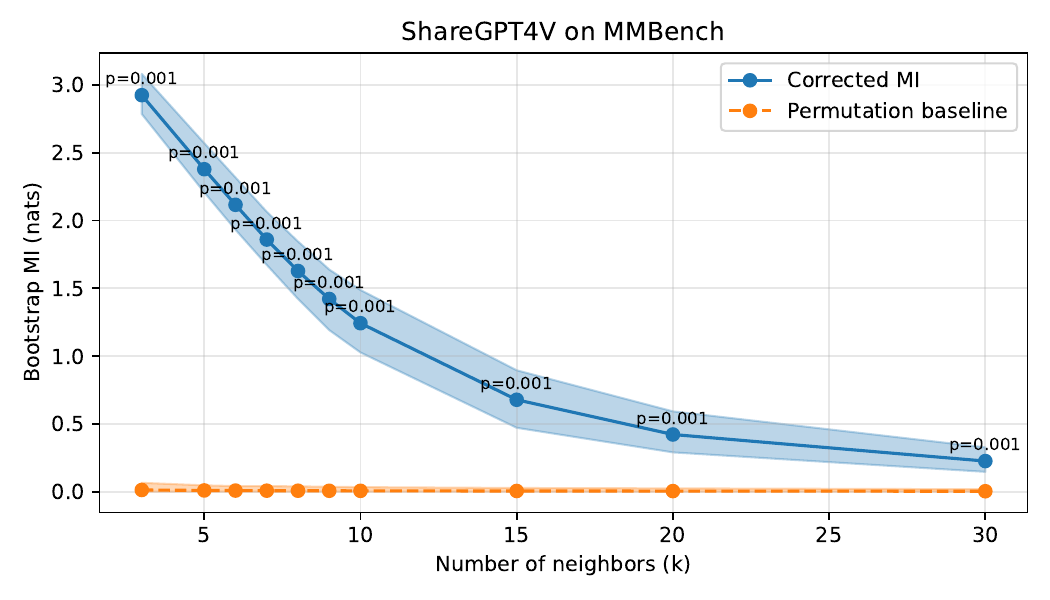}
        % \caption{F1.}
        \label{fig:mi_sharegpt4v_mmbench}
    \end{subfigure}
    % Goes to second line
    \begin{subfigure}[b]{0.49\textwidth}
        \centering
        \includegraphics[width=0.95\textwidth]{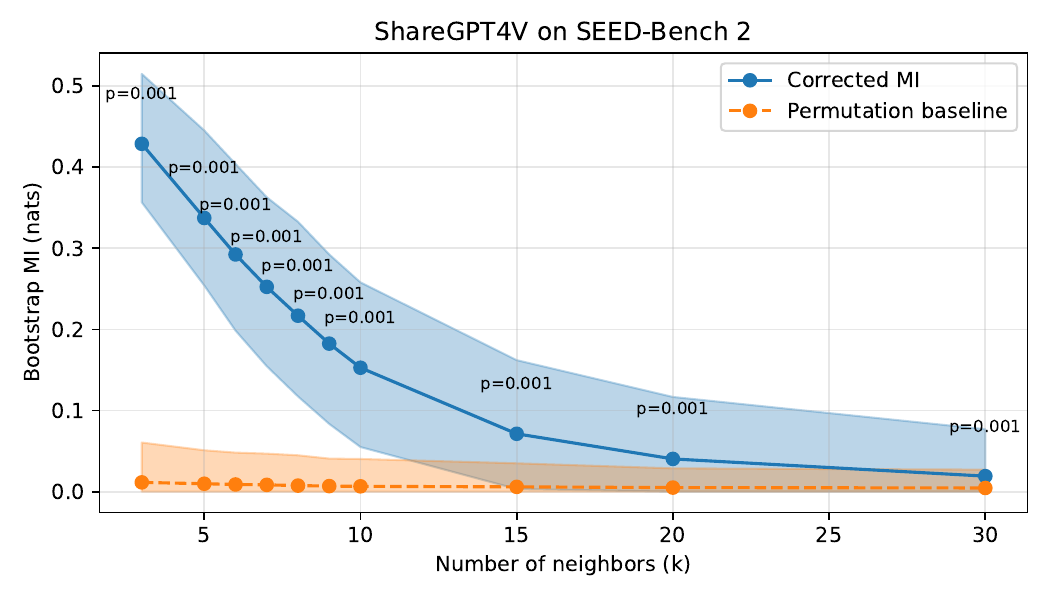}
        % \caption{F1.}
        \label{fig:mi_sharegpt4v_seed}
    \end{subfigure}
    \hfill
    \begin{subfigure}[b]{0.49\textwidth}
        \centering
        \includegraphics[width=0.95\textwidth]{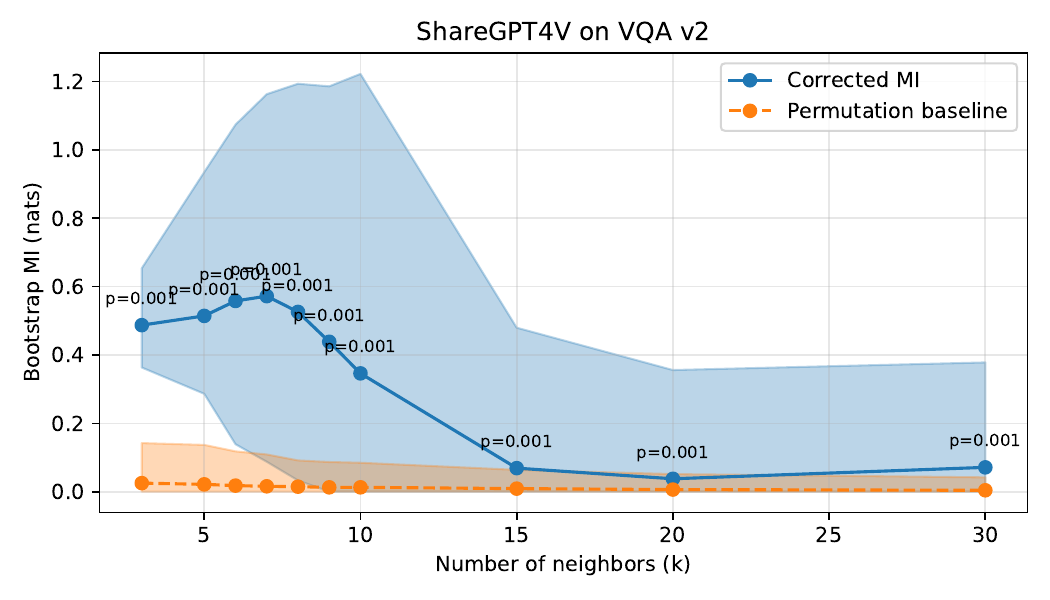}
        % \caption{F1.}
        \label{fig:mi_sharegpt4v_vqav2}
    \end{subfigure}
    \caption{Bootstrapped Mutual Information measures ShareGPT4V \citep{chen2023sharegpt4v}.}
    \label{fig:mi_sharegpt4v}
\end{figure*}

Another important point is that we use the \texttt{scikit-learn} implementation of MI (\texttt{mutual\_information\_regression} and \texttt{mutual\_information\_classif}), which is based on entropy estimation using k-nearest-neighbour distances.
We visually verify the MI estimates for $k = [3, 5, 7, 10, 15, 20, 30]$ to determine a range suitable for robust MI estimation.
From the MI estimates shown in \autoref{fig:mi_internvl2}, \autoref{fig:mi_llava-1.6}, \autoref{fig:mi_qwen2_5_vl}, and \autoref{fig:mi_sharegpt4v}, we can determine that:
\begin{itemize}[leftmargin=*]
    \item Given that MI estimates collapse to $\approx 0$ when we randomly shuffle the data, measured performance is indeed related to the behaviours identified; and, 
    \item For $k \in [5, 10]$, we have a practical and balanced range that is (1) less sensitive to noise and overestimations (low $k$) and (2) less aggressively smoothed (high $k$).
\end{itemize}

To provide robust MI estimates, we take the median MI for $k \in \{5,6,7,8,9,10\}$.

\begin{table*}[t]
\centering
\resizebox{\textwidth}{!}{%
\begin{tabular}{lcccccccc}
\toprule
\multirow{2}[3]{*}{\textbf{Model}} & \multicolumn{2}{c}{\textbf{LLaVa-Bench}} & \multicolumn{2}{c}{\textbf{MMBench}} & \multicolumn{2}{c}{\textbf{SEED-Bench 2}} & \multicolumn{2}{c}{\textbf{VQA v2}} \\
\cmidrule(lr){2-3} \cmidrule(lr){4-5} \cmidrule(lr){6-7} \cmidrule(lr){8-9}
               & \multicolumn{1}{c}{A} & \multicolumn{1}{c}{MI} & \multicolumn{1}{c}{A} & \multicolumn{1}{c}{MI} & \multicolumn{1}{c}{A} & \multicolumn{1}{c}{MI} & \multicolumn{1}{c}{A} & \multicolumn{1}{c}{MI} \\ \midrule
InternVL2      & 0.673 & 3.367 [3.158, 3.589] & 0.796 & 3.200 [3.053, 3.350] & 0.787 & 0.379 [0.325, 0.437] & 0.667 & 0.476 [0.433, 0.526] \\
LLaVa-1.6      & 0.657 & 2.389 [2.052, 2.711] & 0.792 & 2.512 [2.333, 2.699] & 0.667 & 0.438 [0.385, 0.497] & 0.585 & 0.481 [0.437, 0.531] \\
Qwen2.5-VL     & 0.670 & 3.008 [2.764, 3.270] & 0.743 & 1.561 [1.331, 1.792] & 0.833 & 0.188 [0.115, 0.275] & 0.700 & 0.147 [0.082, 0.230] \\
ShareGPT4V     & 0.687 & 2.058 [1.737, 2.405] & 0.705 & 2.116 [1.932, 2.318] & 0.661 & 0.292 [0.198, 0.404] & 0.810 & 0.558 [0.139, 1.074] \\ \bottomrule
\end{tabular}%
}
\caption{VLMs accuracy (A) on the four datasets and Mutual Information (MI) measures between accuracy and model behaviour types. In parentheses, we report the bootstrapped 95\% confidence intervals for the MI measures.}
\label{tab:effectiveness_acc_full}
\end{table*}

\subsection{Examples of Model Behaviours} \label{app:examples_behaviours}
\begin{figure*}[t]
    \centering
    \includegraphics[width=\textwidth]{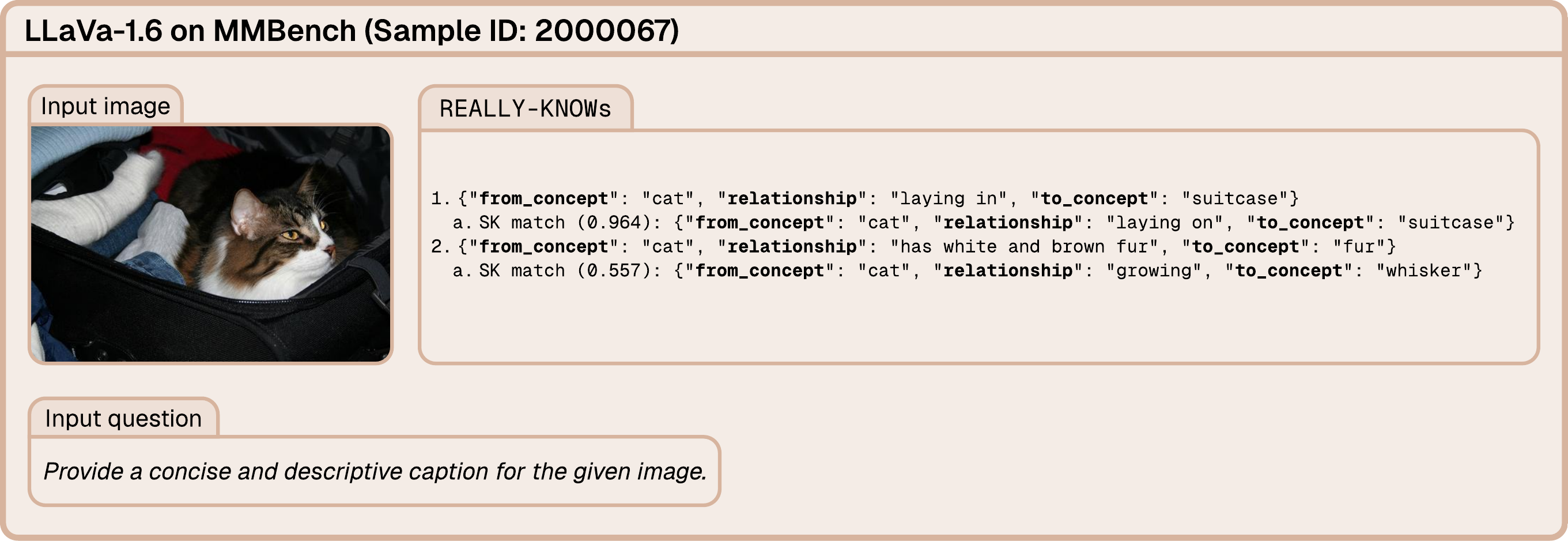}
    \caption{Example of aligned model behaviour from LLaVa-1.6 on MMBench.}
    \label{fig:llava16_mmbench_type1}
\end{figure*}

\begin{figure*}[t]
    \centering
    \includegraphics[width=\textwidth]{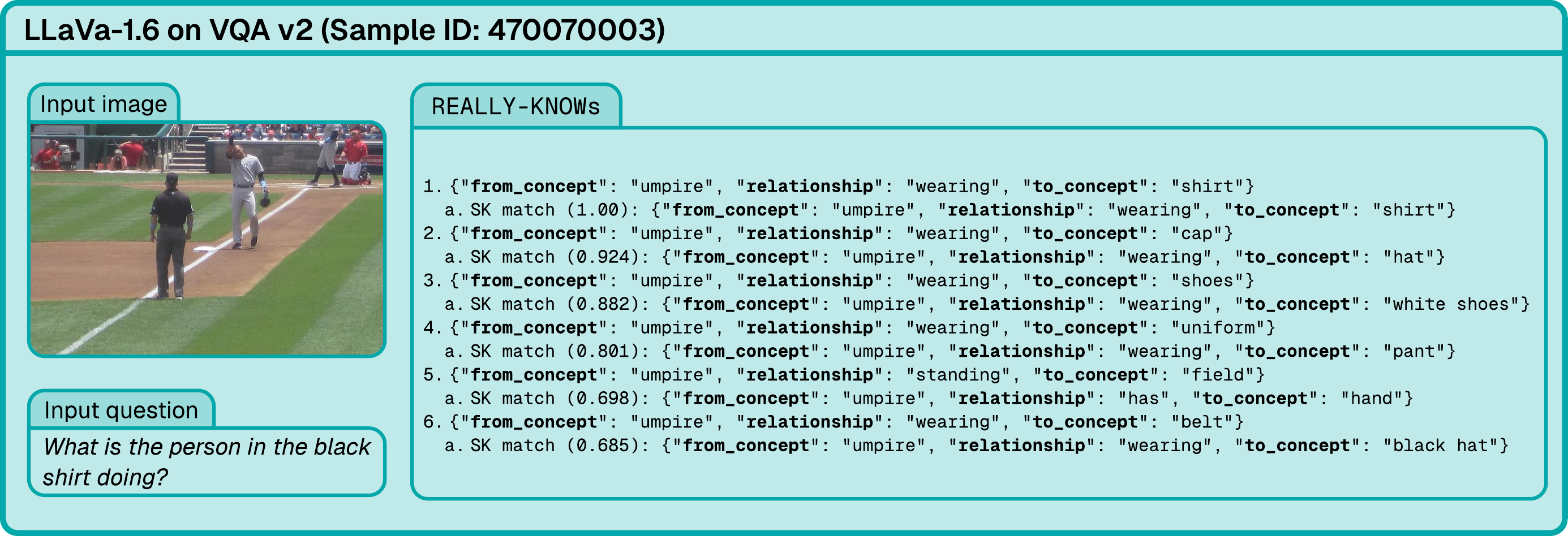}
    \caption{Example of aligned model behaviour from LLaVa-1.6 on VQA v2.}
    \label{fig:llava16_vqav2_type1}
\end{figure*}

\begin{figure*}[t]
    \centering
    \includegraphics[width=\textwidth]{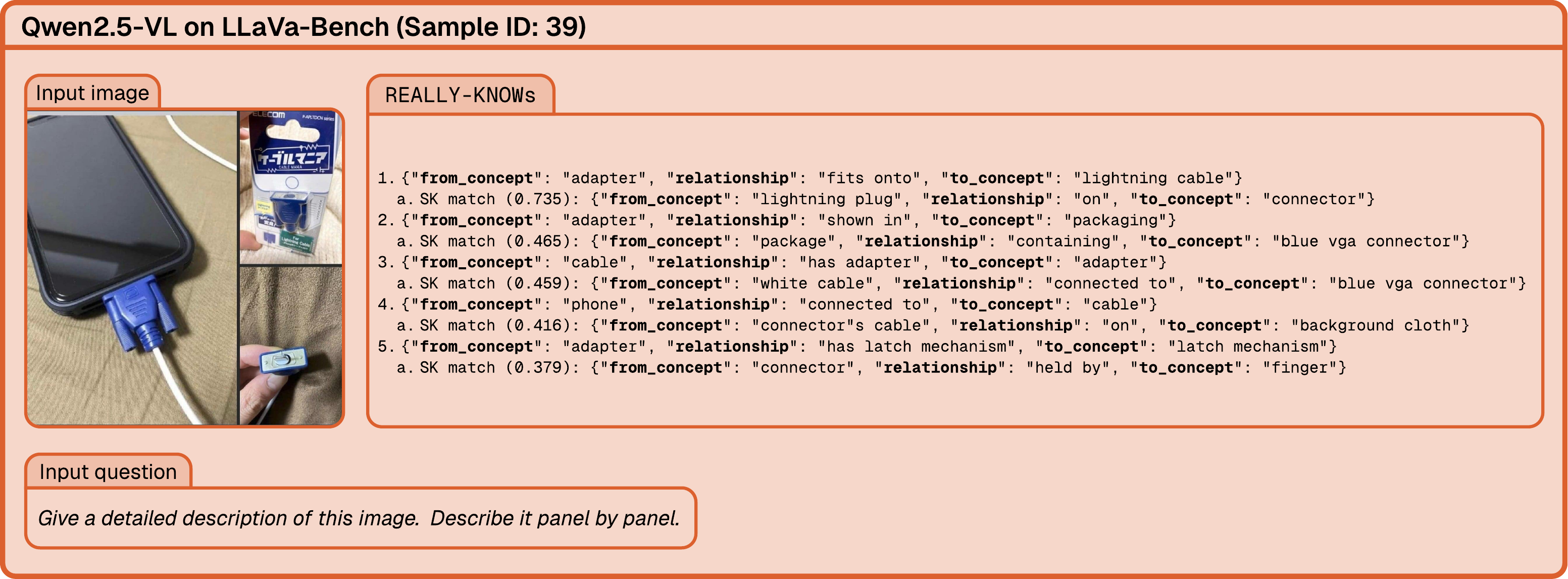}
    \caption{Example of expanded model behaviour from Qwen2.5-VL on LLaVa-Bench.}
    \label{fig:qwen25vl_llavabench_type2}
\end{figure*}

\begin{figure*}[t]
    \centering
    \includegraphics[width=\textwidth]{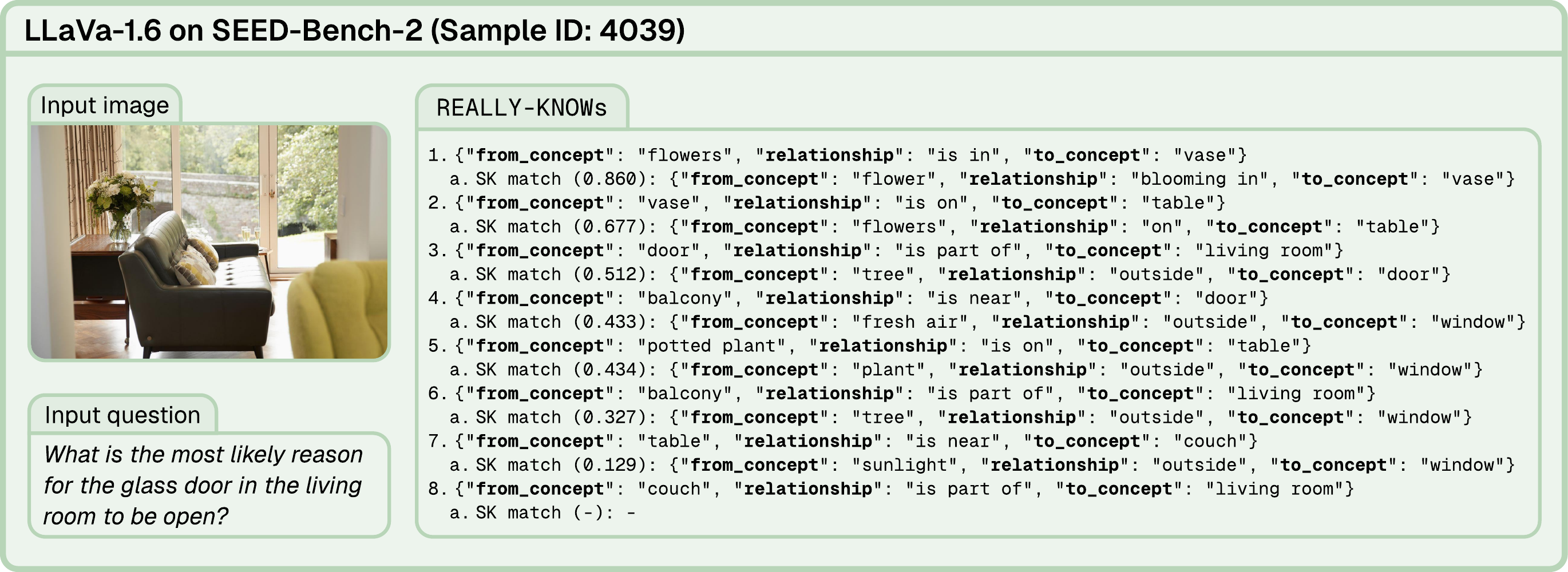}
    \caption{Example of expanded model behaviour from LLaVa-1.6 on SEED-Bench-2.}
    \label{fig:llava16_seed_type2}
\end{figure*}

\begin{figure*}[t]
    \centering
    \includegraphics[width=\textwidth]{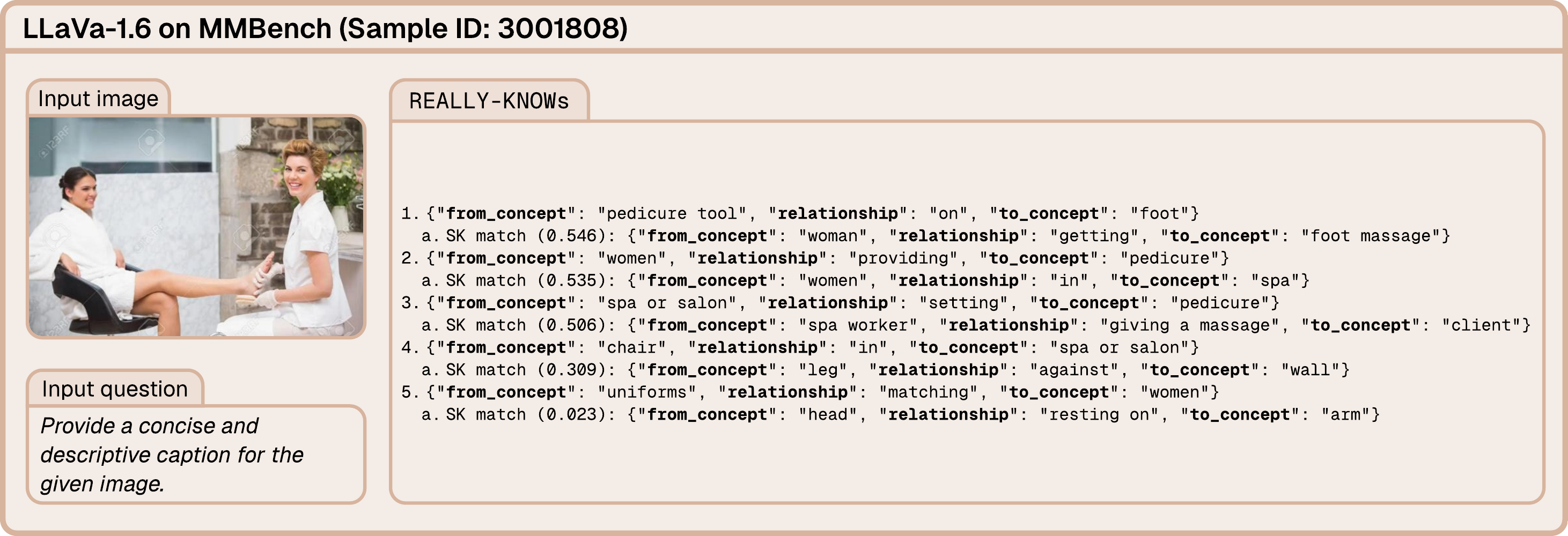}
    \caption{Example of expanded model behaviour from LLaVa-1.6 on MMBench.}
    \label{fig:llava16_mmbench_type2}
\end{figure*}

\begin{figure*}[t]
    \centering
    \includegraphics[width=\textwidth]{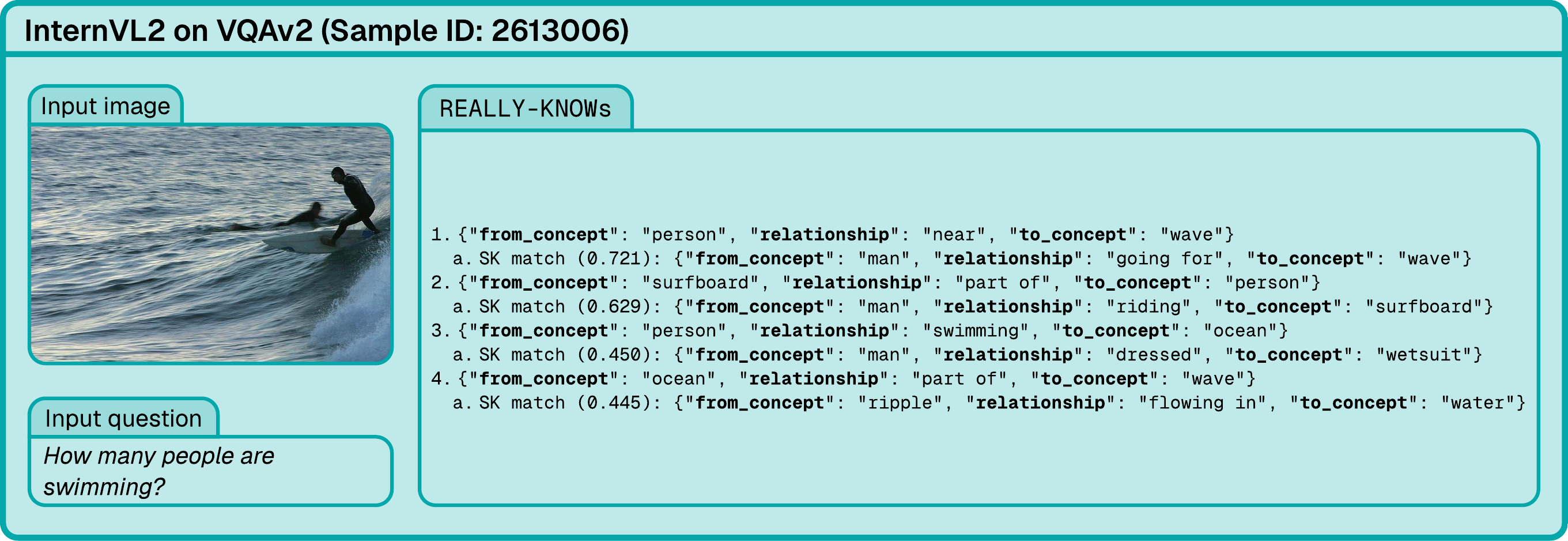}
    \caption{Example of expanded model behaviour from InternVL2 on VQA v2.}
    \label{fig:internvl2_vqav2_type2}
\end{figure*}

\begin{figure*}[t]
    \centering
    \includegraphics[width=\textwidth]{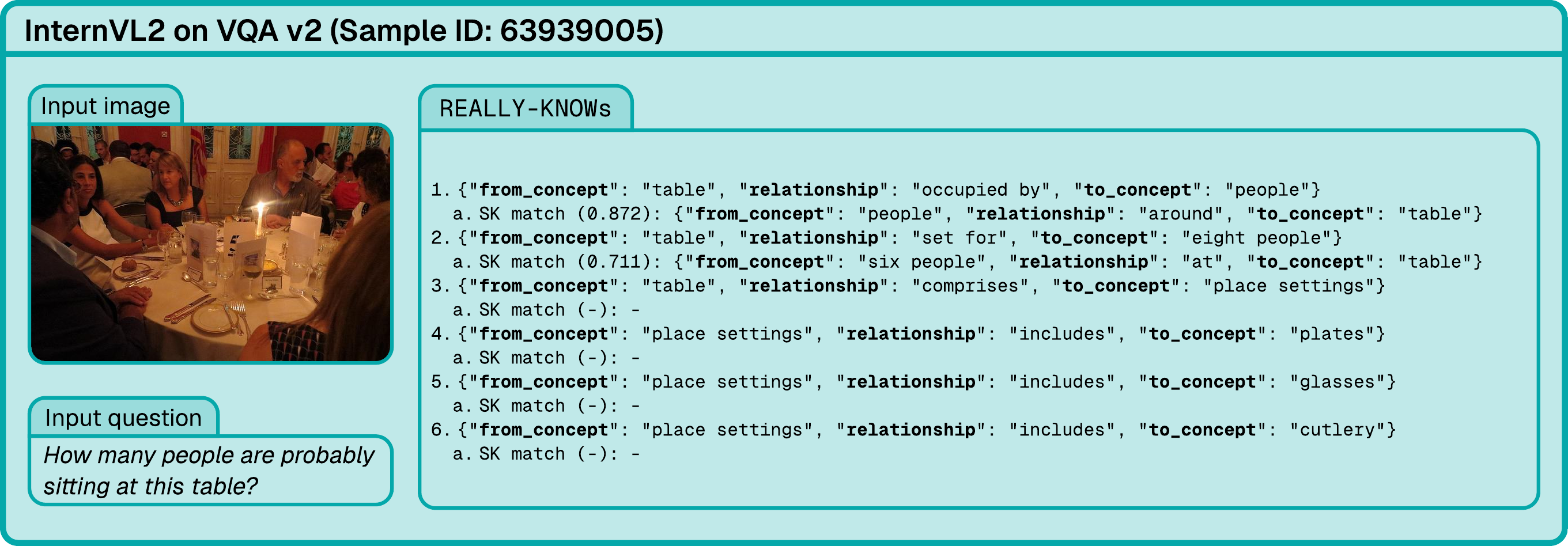}
    \caption{Example of divergent model behaviour from InternVL2 on VQA v2.}
    \label{fig:internvl2_vqav2_type3}
\end{figure*}

\begin{figure*}[t]
    \centering
    \includegraphics[width=\textwidth]{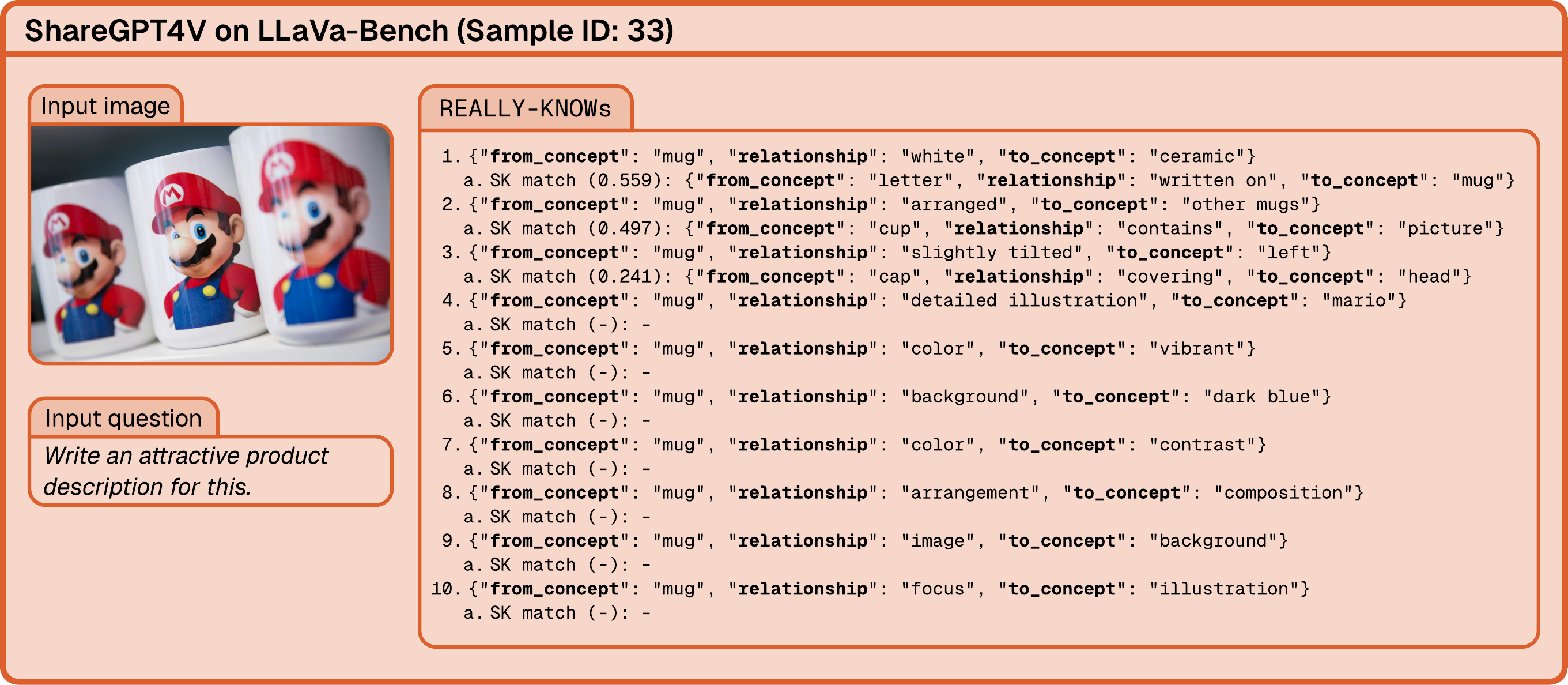}
    \caption{Example of divergent model behaviour from ShareGPT4V on LLaVa-Bench.}
    \label{fig:sharegpt4v_llavabench_type3}
\end{figure*}

\begin{itemize}[leftmargin=*]
    \item Aligned behaviours: \autoref{fig:llava16_mmbench_type1}, \autoref{fig:llava16_vqav2_type1}
    \item Expanded behaviours: \autoref{fig:qwen25vl_llavabench_type2}, \autoref{fig:llava16_seed_type2}, \autoref{fig:llava16_mmbench_type2}, \autoref{fig:internvl2_vqav2_type2}
    \item Divergent behaviours: \autoref{fig:internvl2_vqav2_type3}, \autoref{fig:sharegpt4v_llavabench_type3}
\end{itemize}

% == == == == == == == == ==
% == == Aligned behaviours
% == == == == == == == == ==

% == == == == == == == == ==
% == == Expanded behaviours
% == == == == == == == == ==

% == == == == == == == == ==
% == == Divergent behaviours
% == == == == == == == == ==

\subsection{Model-pairwise Distributions of Causal Effects} \label{app:distr_causal_effects}
\begin{figure*}[t]
    \centering
    \begin{subfigure}[b]{0.245\textwidth}
        \centering
        \includegraphics[width=0.98\textwidth]{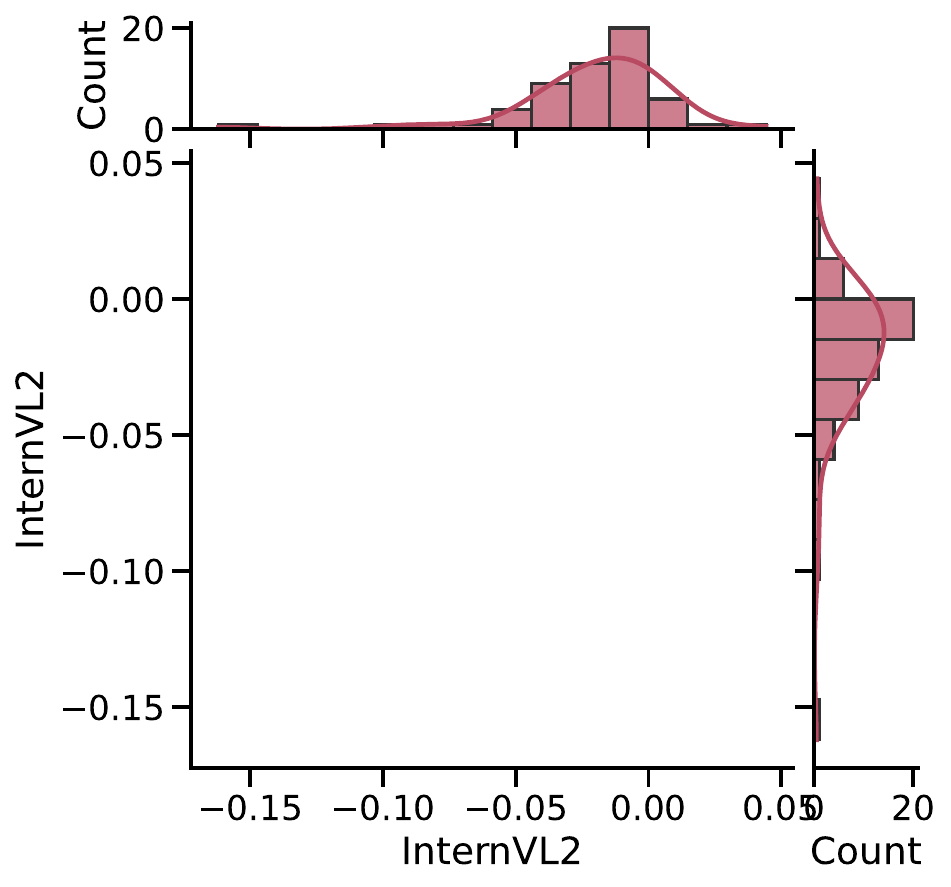}
    \end{subfigure}
    \hfill
    \begin{subfigure}[b]{0.245\textwidth}
        \centering
        \includegraphics[width=0.98\textwidth]{imgs/distr_effects/llava-bench/internvl2_llava-1.6.pdf}
    \end{subfigure}
    \hfill
    \begin{subfigure}[b]{0.245\textwidth}
        \centering
        \includegraphics[width=0.98\textwidth]{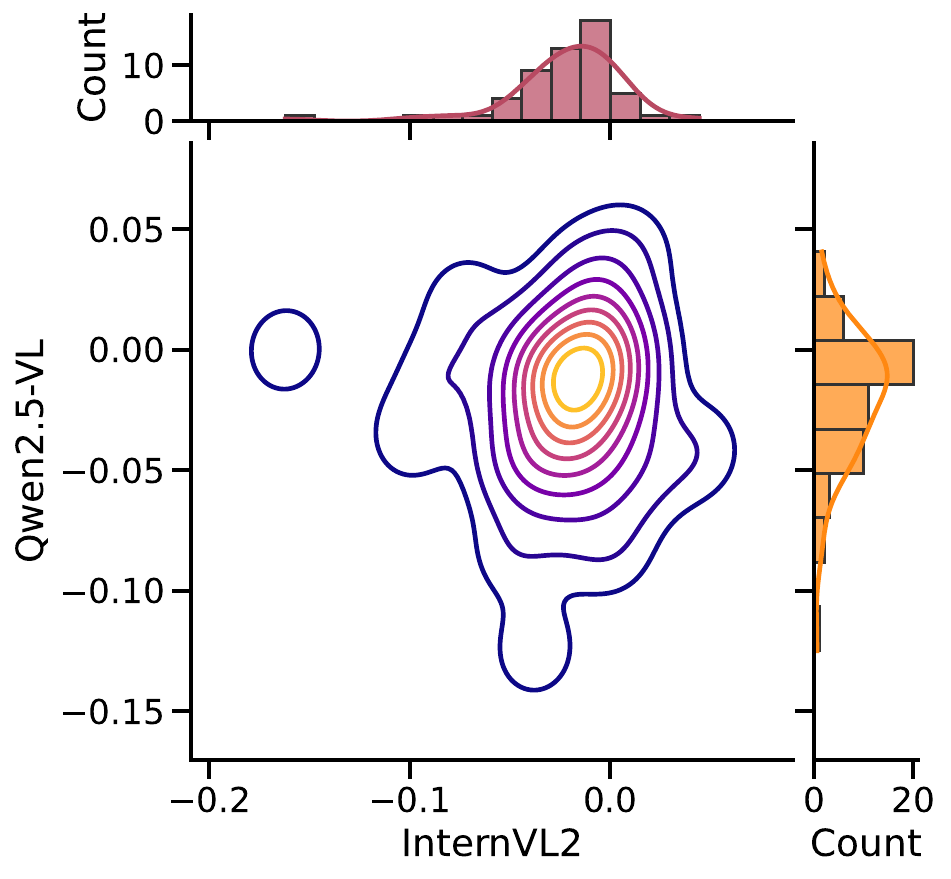}
    \end{subfigure}
    \hfill
    \begin{subfigure}[b]{0.245\textwidth}
        \centering
        \includegraphics[width=0.98\textwidth]{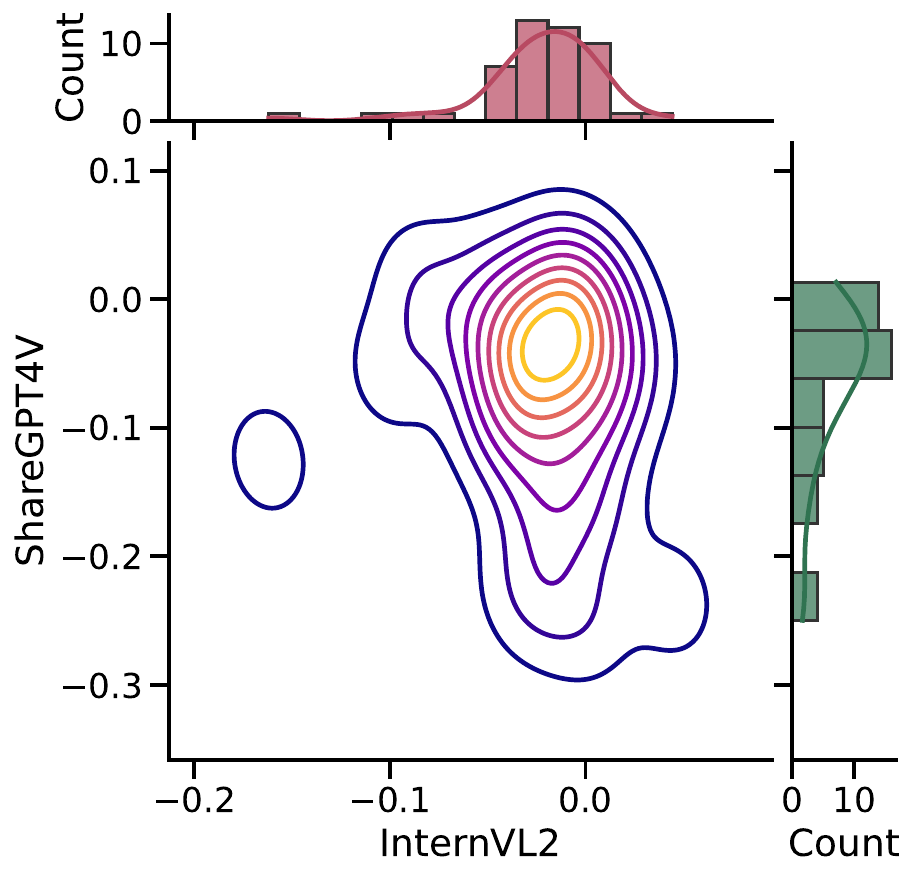}
    \end{subfigure}
    % new line
    \begin{subfigure}[b]{0.245\textwidth}
        \centering
        \includegraphics[width=0.98\textwidth]{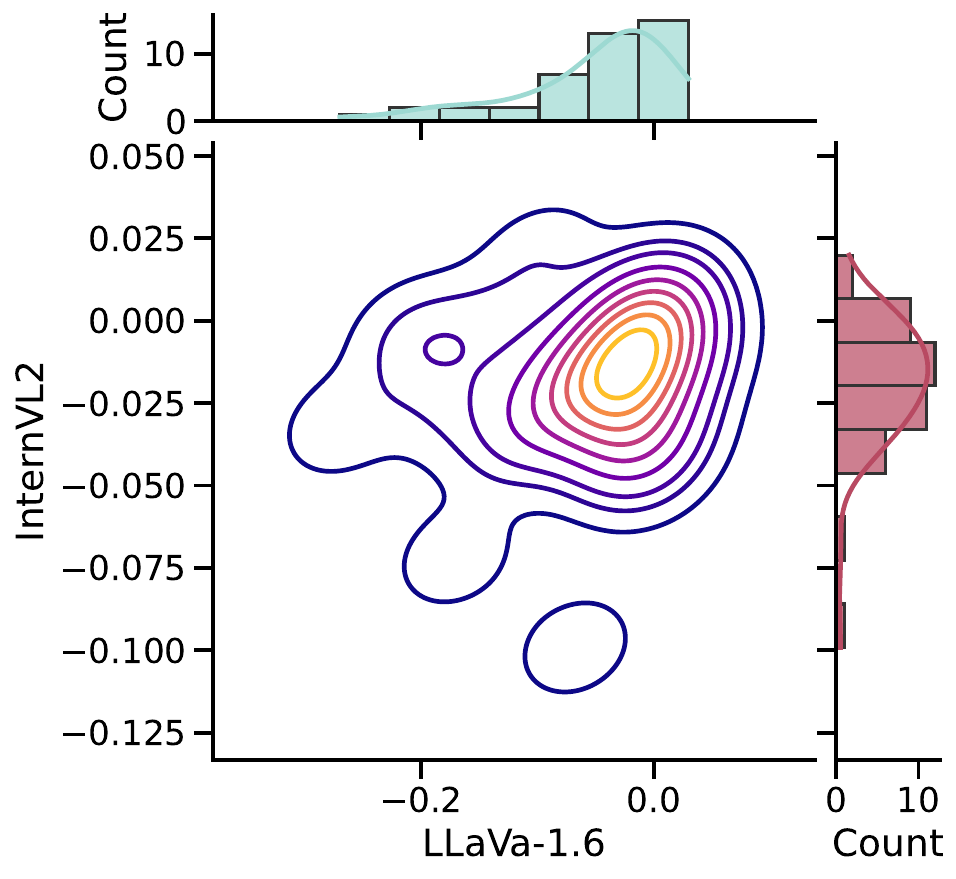}
    \end{subfigure}
    \hfill
    \begin{subfigure}[b]{0.245\textwidth}
        \centering
        \includegraphics[width=0.98\textwidth]{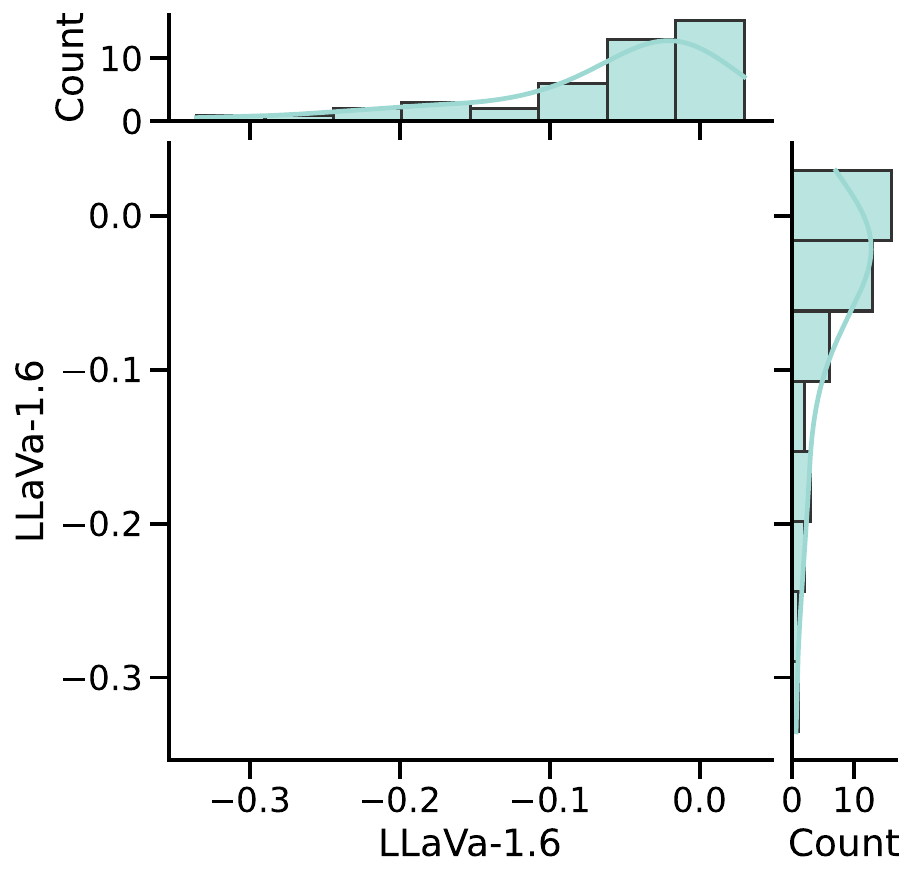}
    \end{subfigure}
    \hfill
    \begin{subfigure}[b]{0.245\textwidth}
        \centering
        \includegraphics[width=0.98\textwidth]{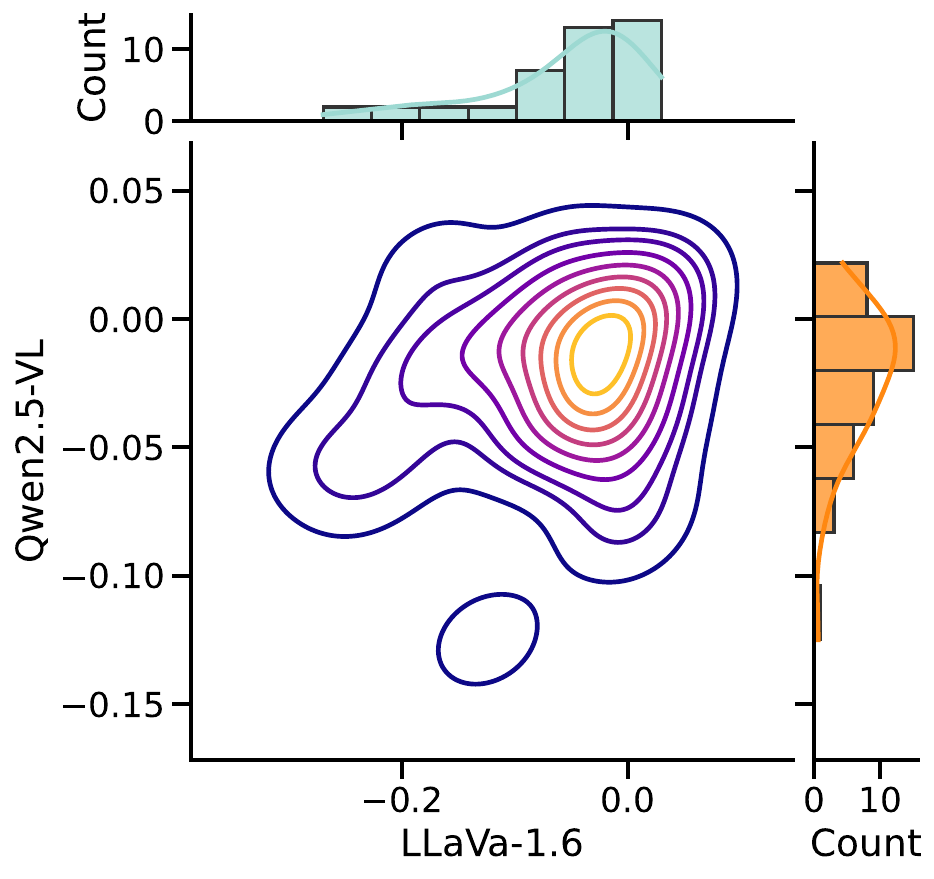}
    \end{subfigure}
    \hfill
    \begin{subfigure}[b]{0.245\textwidth}
        \centering
        \includegraphics[width=0.98\textwidth]{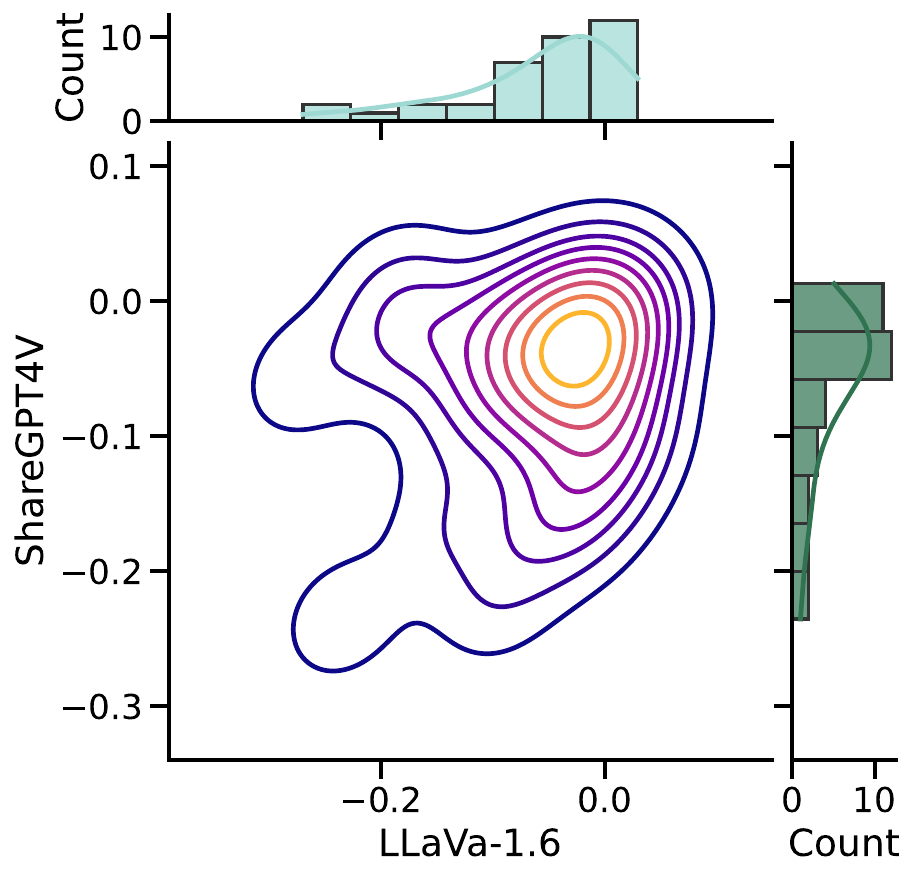}
    \end{subfigure}
    % new line
    \begin{subfigure}[b]{0.245\textwidth}
        \centering
        \includegraphics[width=0.98\textwidth]{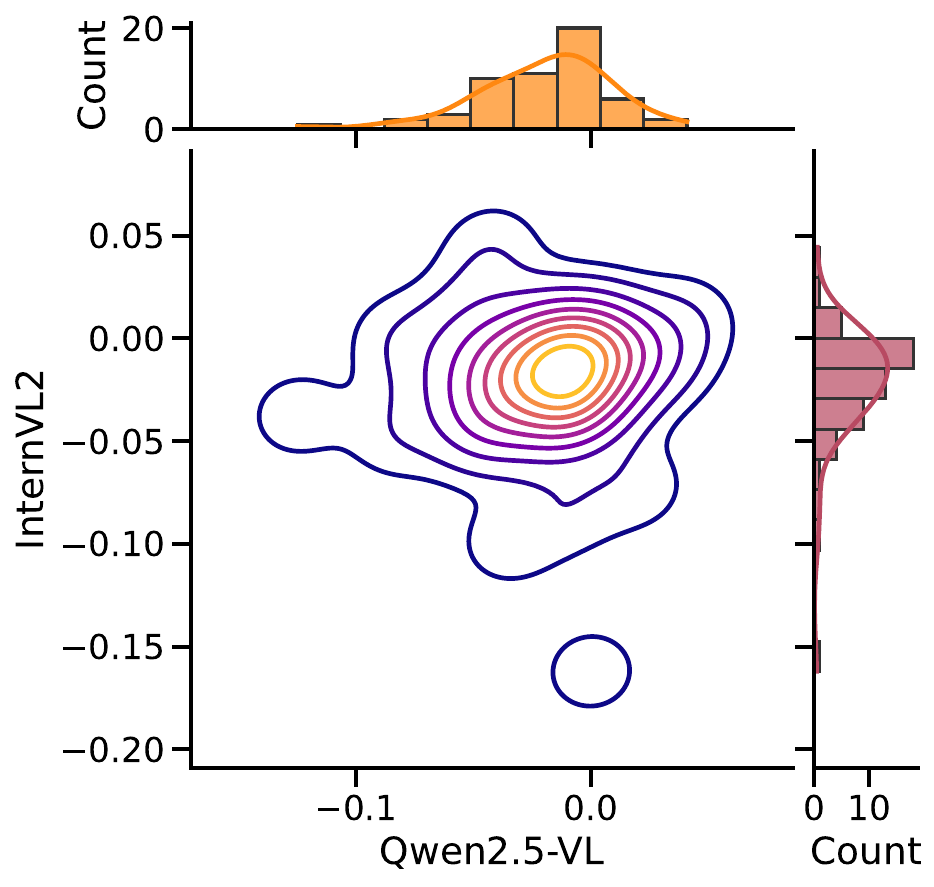}
    \end{subfigure}
    \hfill
    \begin{subfigure}[b]{0.245\textwidth}
        \centering
        \includegraphics[width=0.98\textwidth]{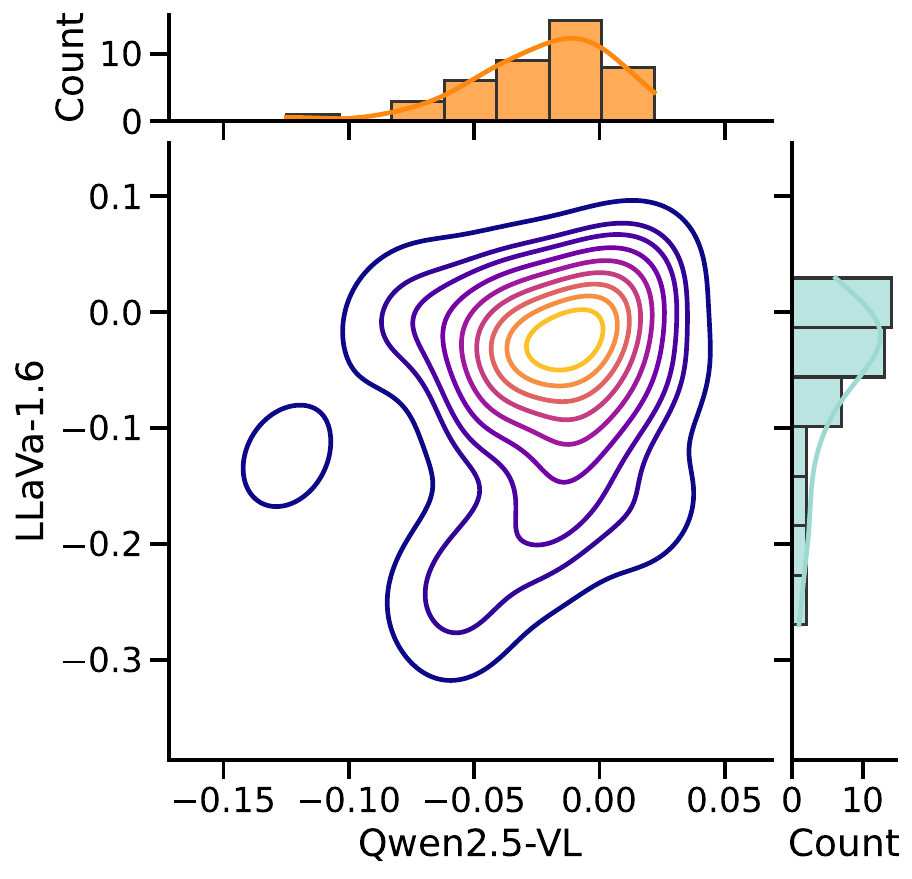}
    \end{subfigure}
    \hfill
    \begin{subfigure}[b]{0.245\textwidth}
        \centering
        \includegraphics[width=0.98\textwidth]{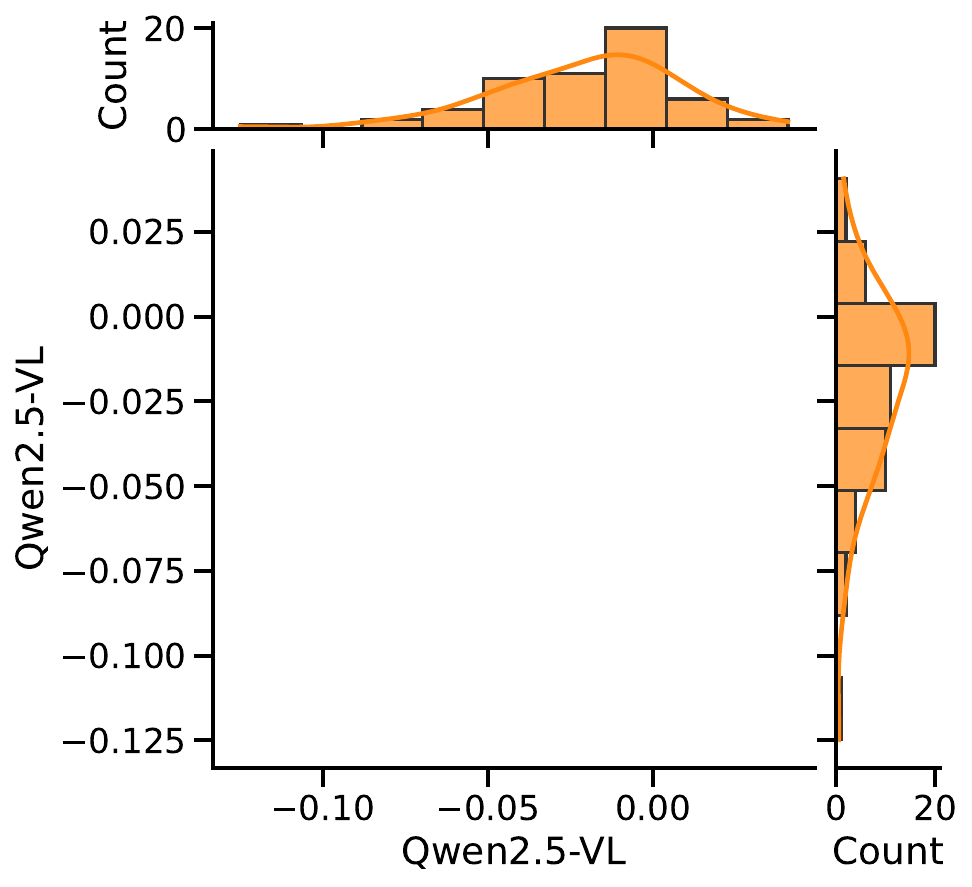}
    \end{subfigure}
    \hfill
    \begin{subfigure}[b]{0.245\textwidth}
        \centering
        \includegraphics[width=0.98\textwidth]{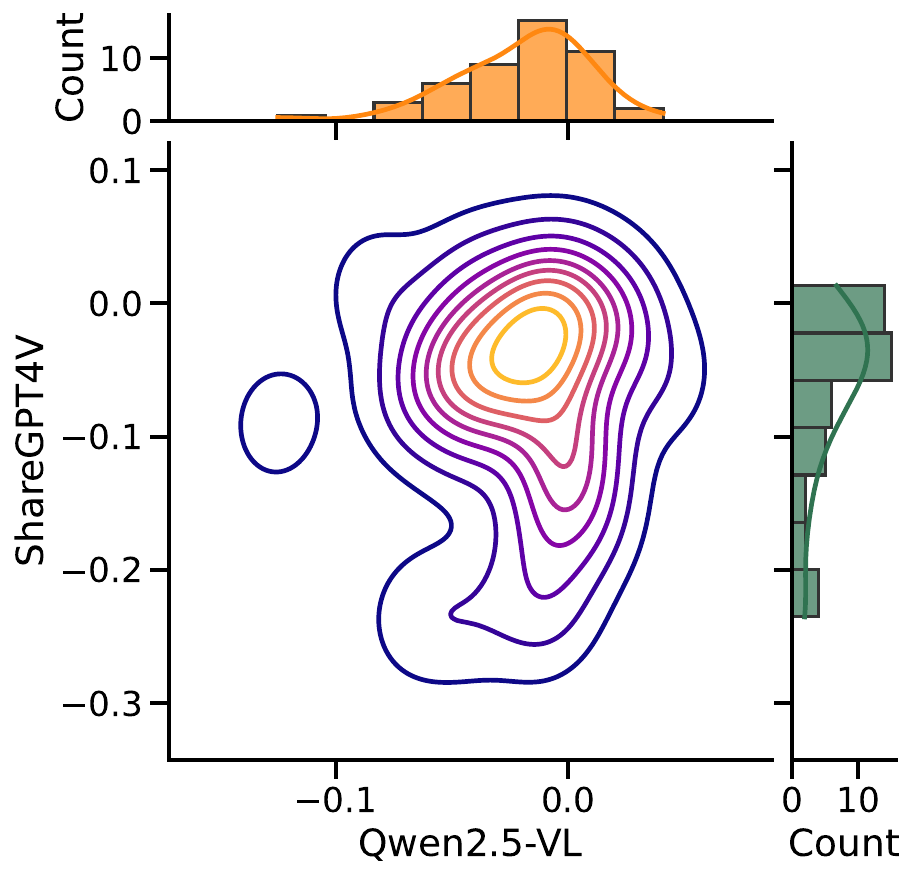}
    \end{subfigure}
    % new line
    \begin{subfigure}[b]{0.245\textwidth}
        \centering
        \includegraphics[width=0.98\textwidth]{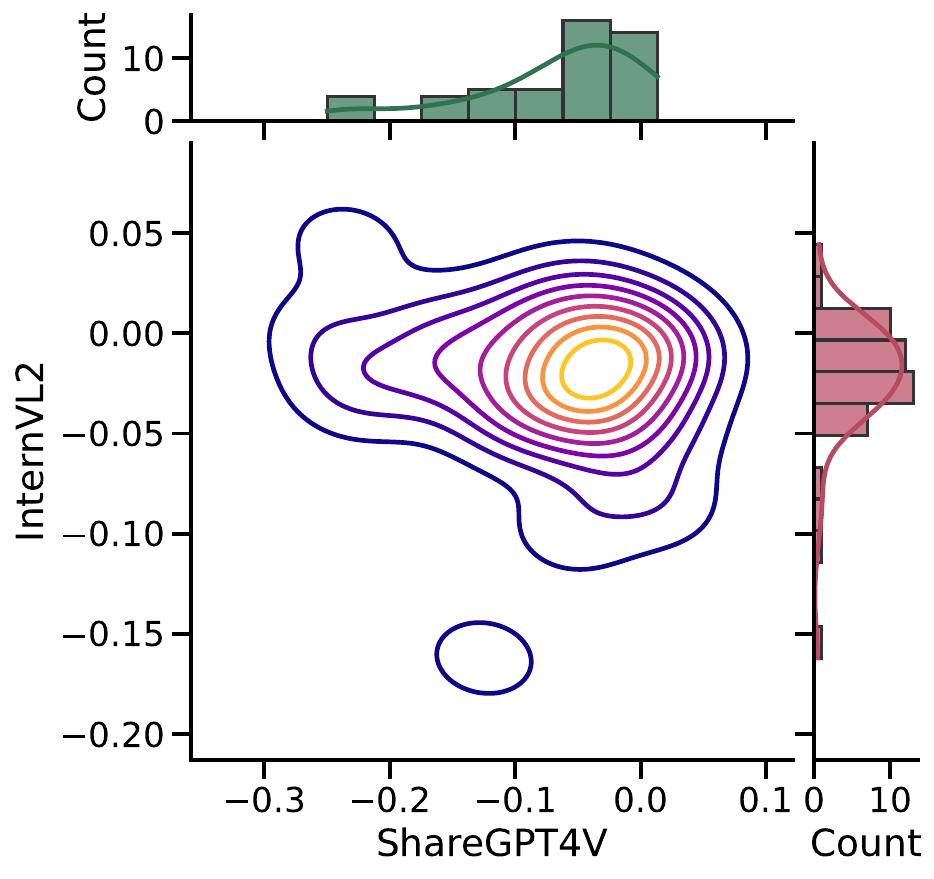}
    \end{subfigure}
    \hfill
    \begin{subfigure}[b]{0.245\textwidth}
        \centering
        \includegraphics[width=0.98\textwidth]{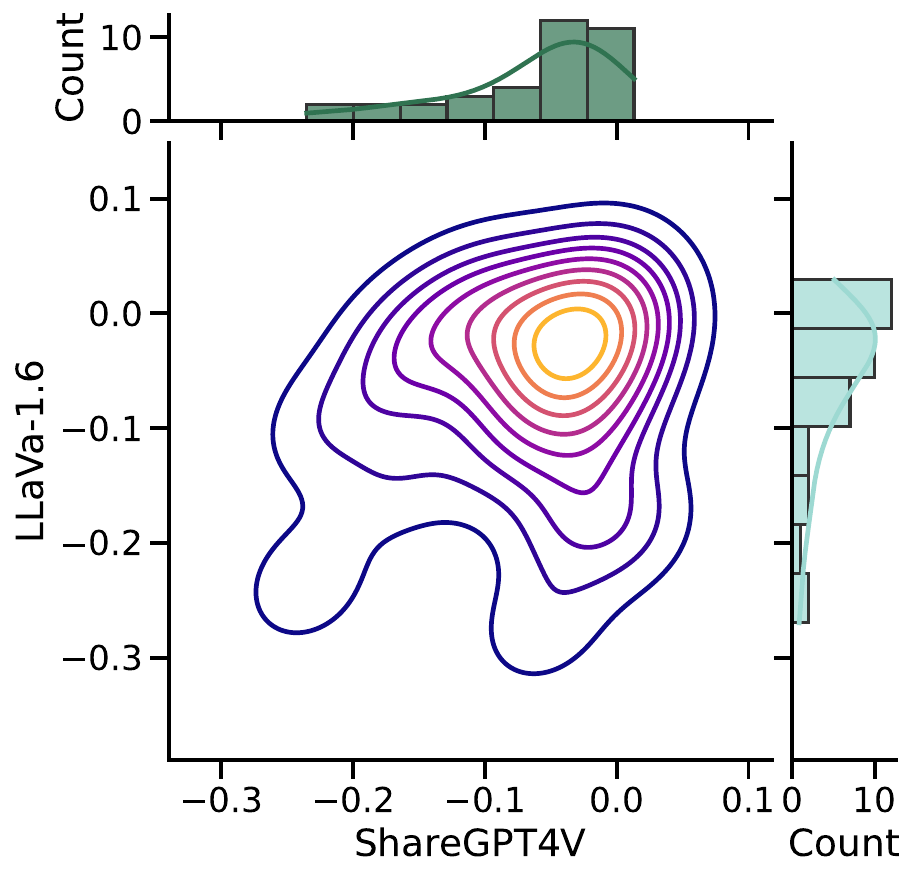}
    \end{subfigure}
    \hfill
    \begin{subfigure}[b]{0.245\textwidth}
        \centering
        \includegraphics[width=0.98\textwidth]{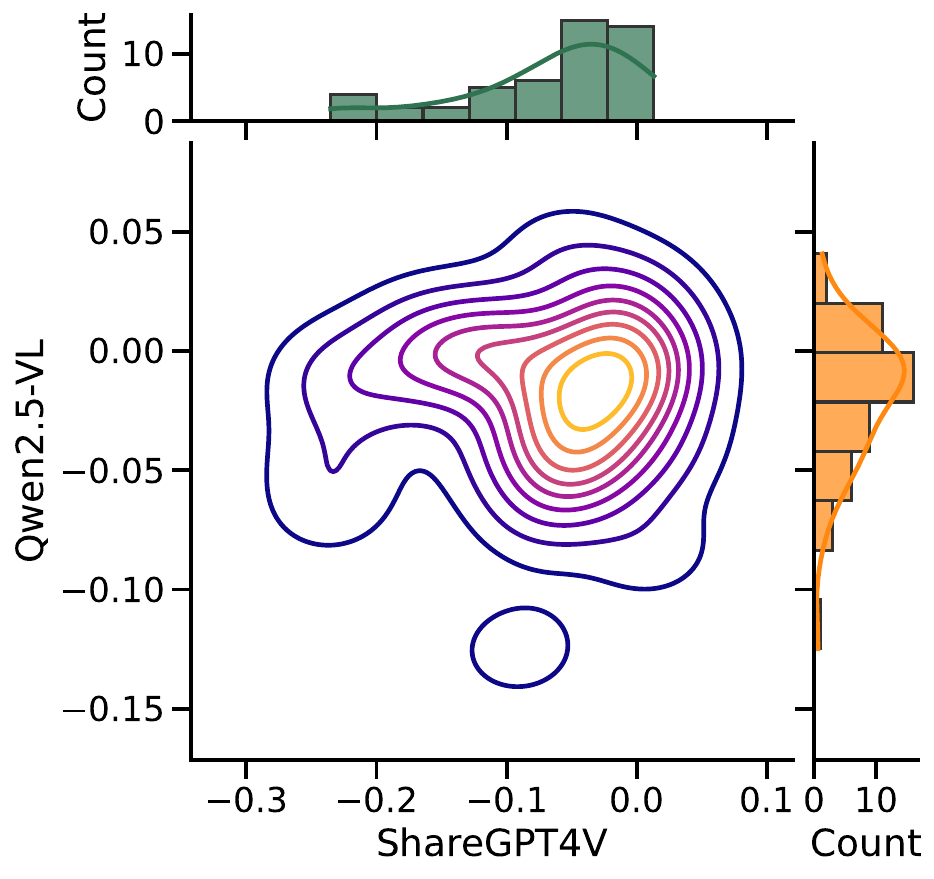}
    \end{subfigure}
    \hfill
    \begin{subfigure}[b]{0.245\textwidth}
        \centering
        \includegraphics[width=0.98\textwidth]{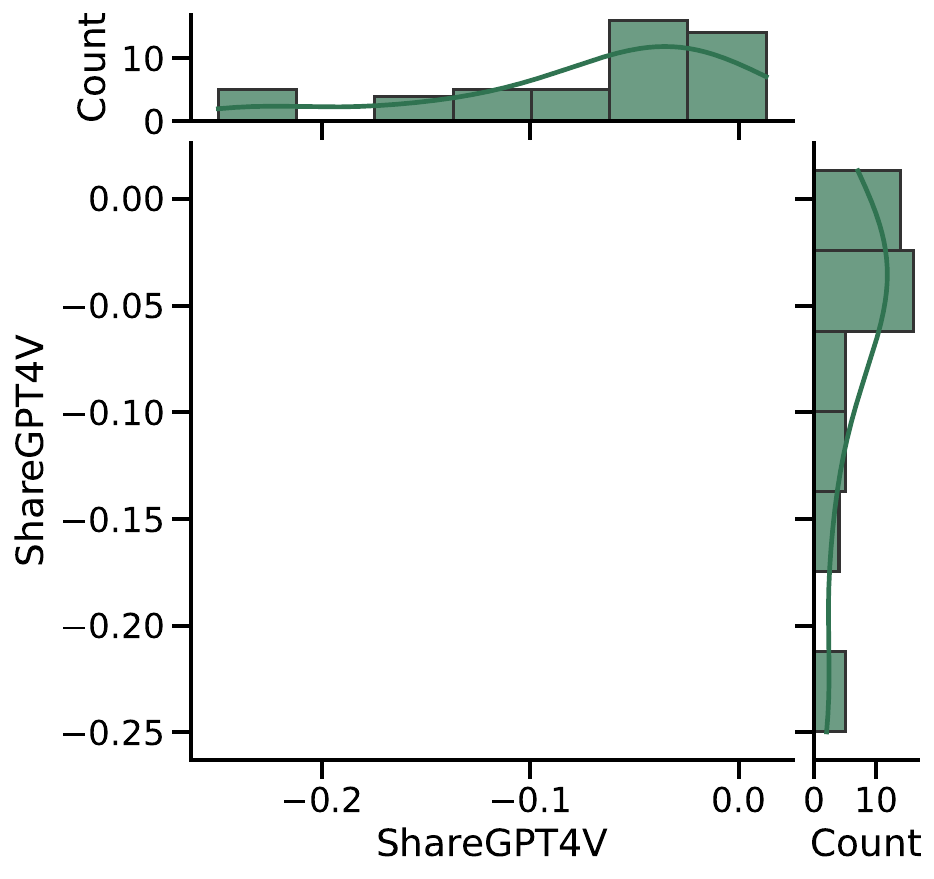}
    \end{subfigure}
    \caption{Model-pairwise distributions of estimated causal effects on LLaVa-Bench \citep{Liu2023LLaVa}.}
    \label{fig:effects_llava_bench}
\end{figure*}

\begin{figure*}[t]
    \centering
    \begin{subfigure}[b]{0.245\textwidth}
        \centering
        \includegraphics[width=0.98\textwidth]{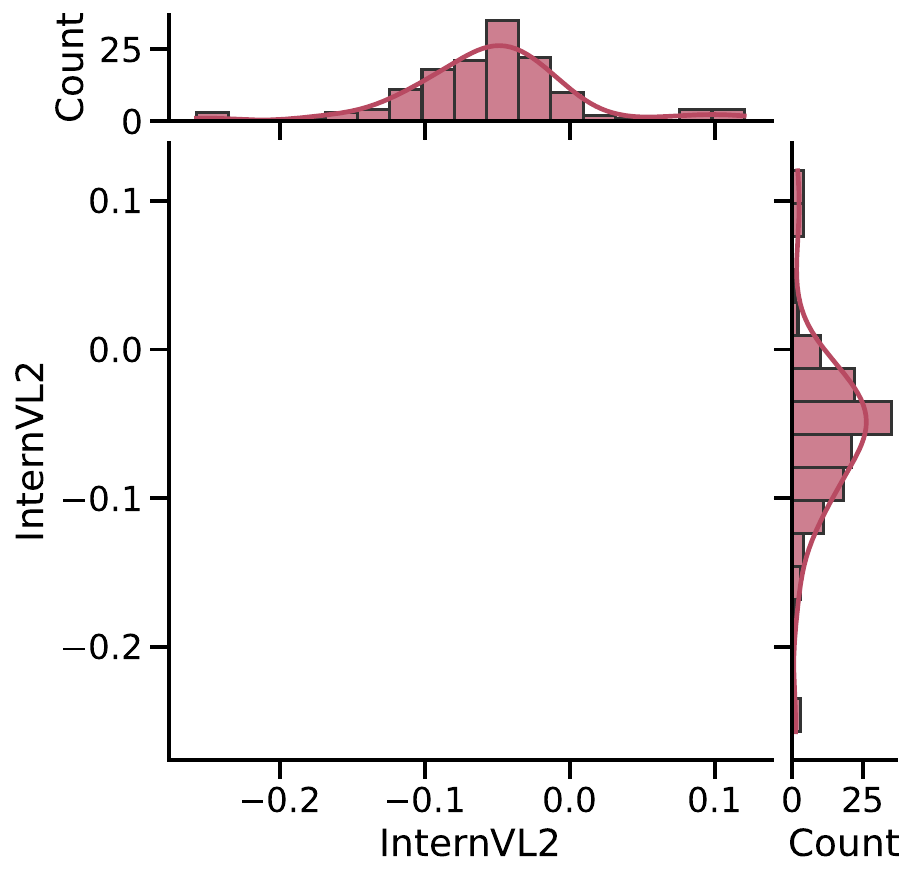}
    \end{subfigure}
    \hfill
    \begin{subfigure}[b]{0.245\textwidth}
        \centering
        \includegraphics[width=0.98\textwidth]{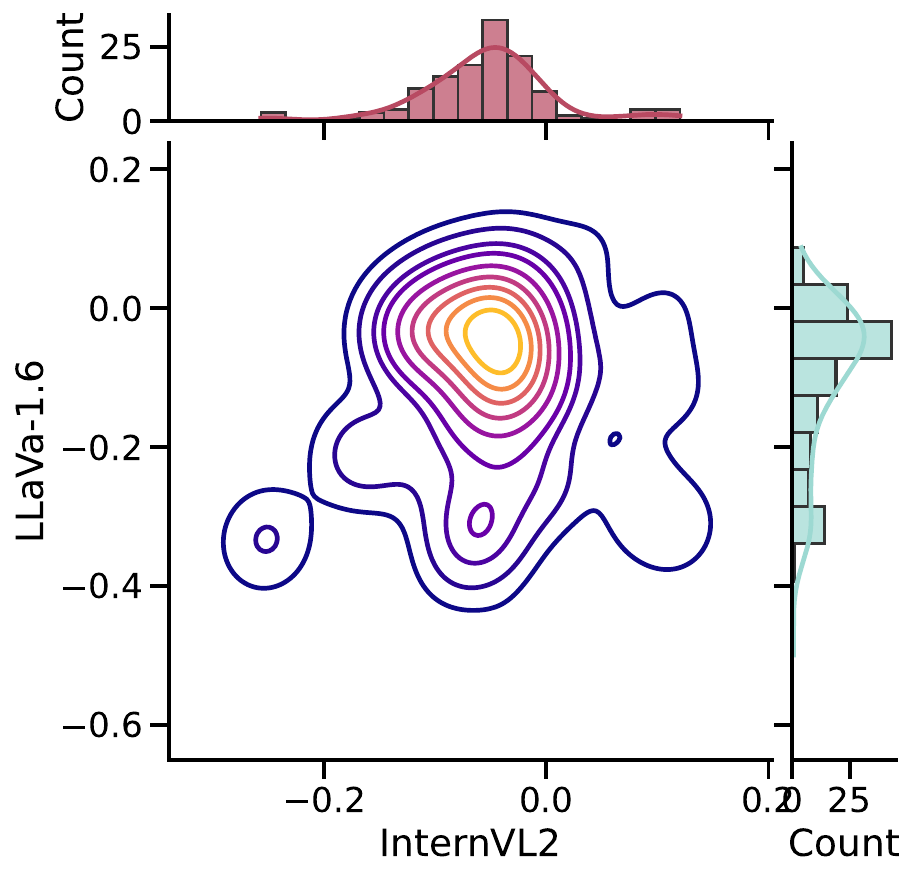}
    \end{subfigure}
    \hfill
    \begin{subfigure}[b]{0.245\textwidth}
        \centering
        \includegraphics[width=0.98\textwidth]{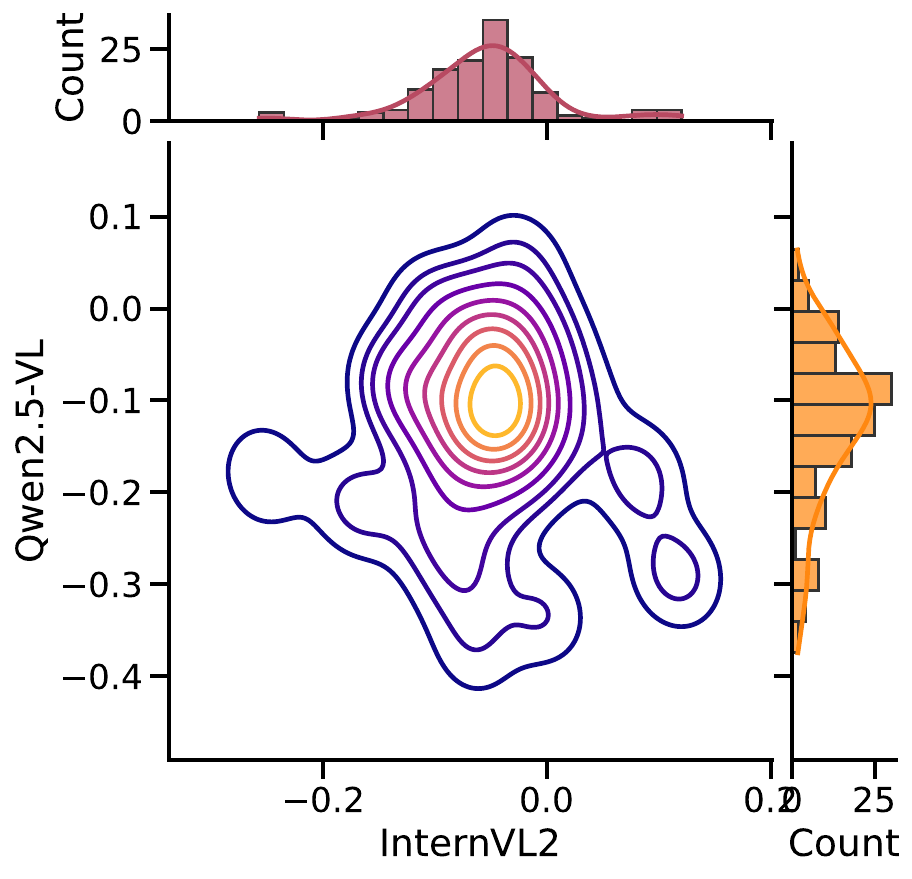}
    \end{subfigure}
    \hfill
    \begin{subfigure}[b]{0.245\textwidth}
        \centering
        \includegraphics[width=0.98\textwidth]{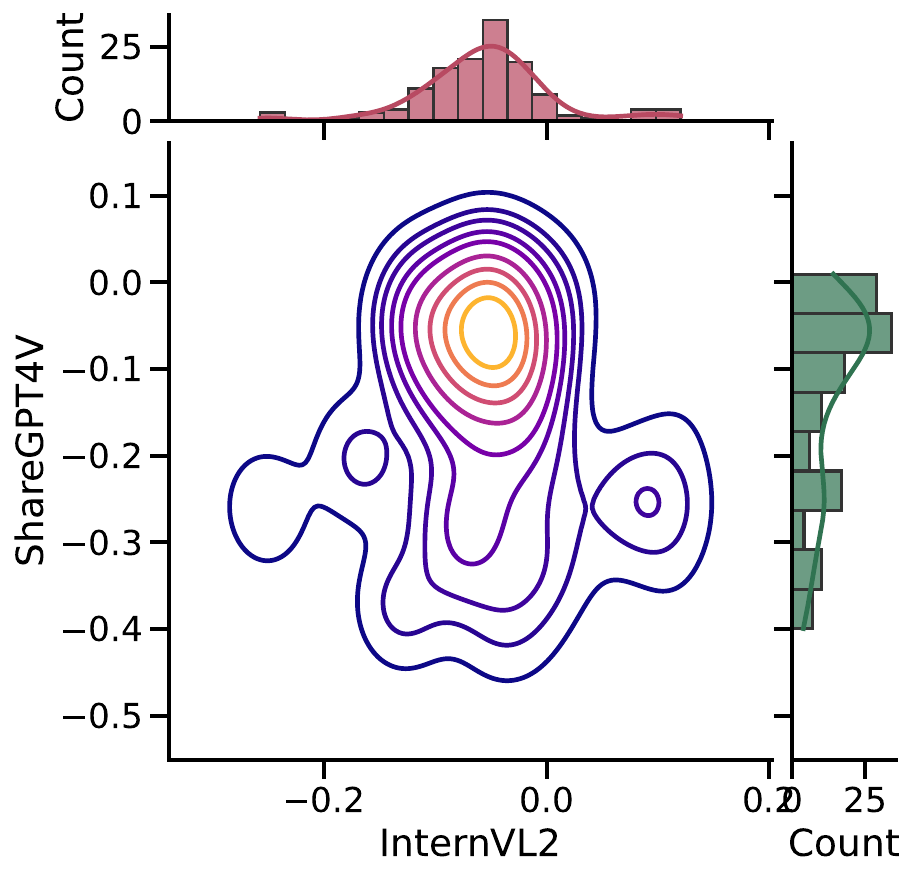}
    \end{subfigure}
    % new line
    \begin{subfigure}[b]{0.245\textwidth}
        \centering
        \includegraphics[width=0.98\textwidth]{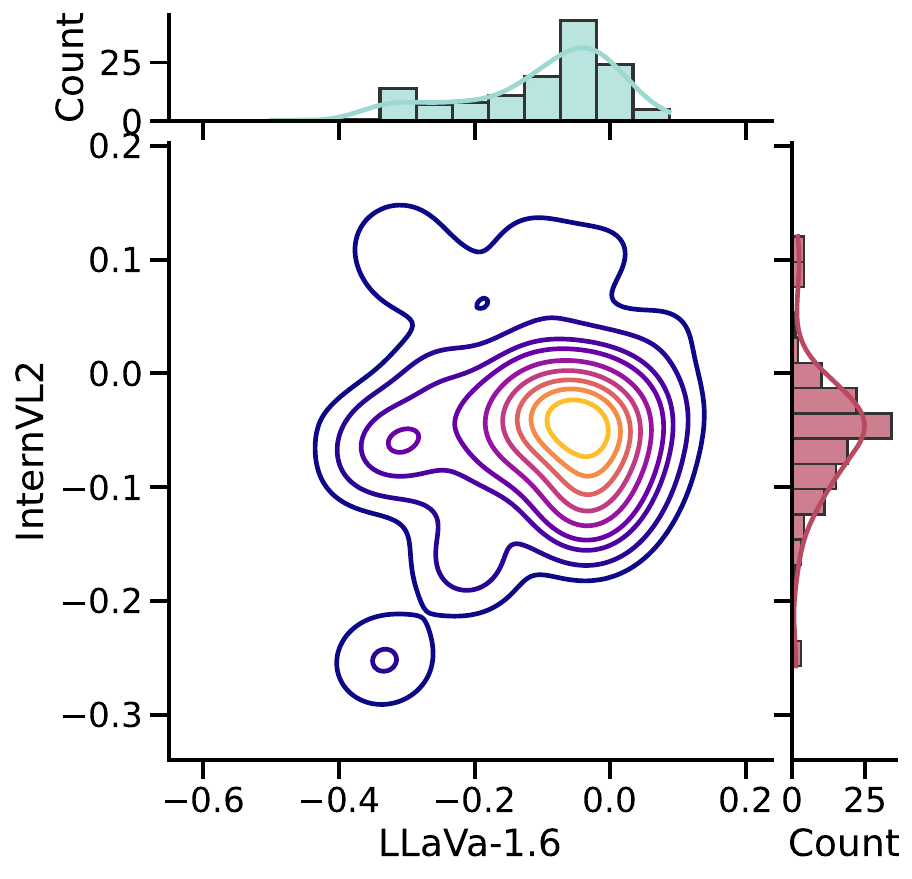}
    \end{subfigure}
    \hfill
    \begin{subfigure}[b]{0.245\textwidth}
        \centering
        \includegraphics[width=0.98\textwidth]{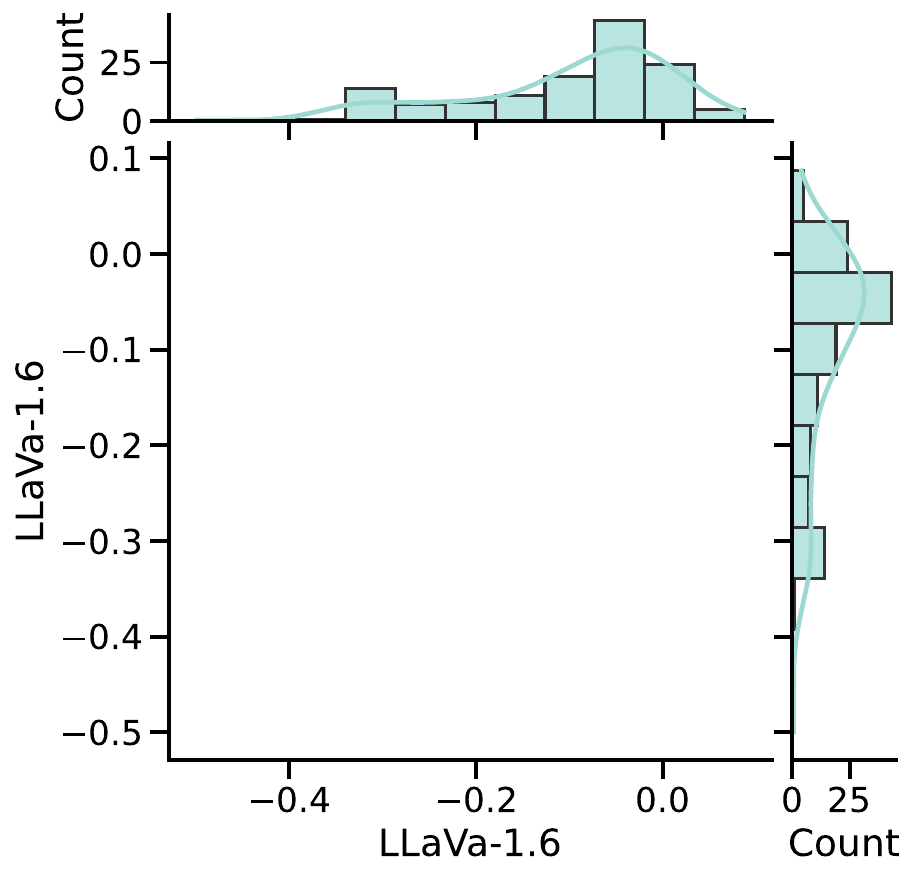}
    \end{subfigure}
    \hfill
    \begin{subfigure}[b]{0.245\textwidth}
        \centering
        \includegraphics[width=0.98\textwidth]{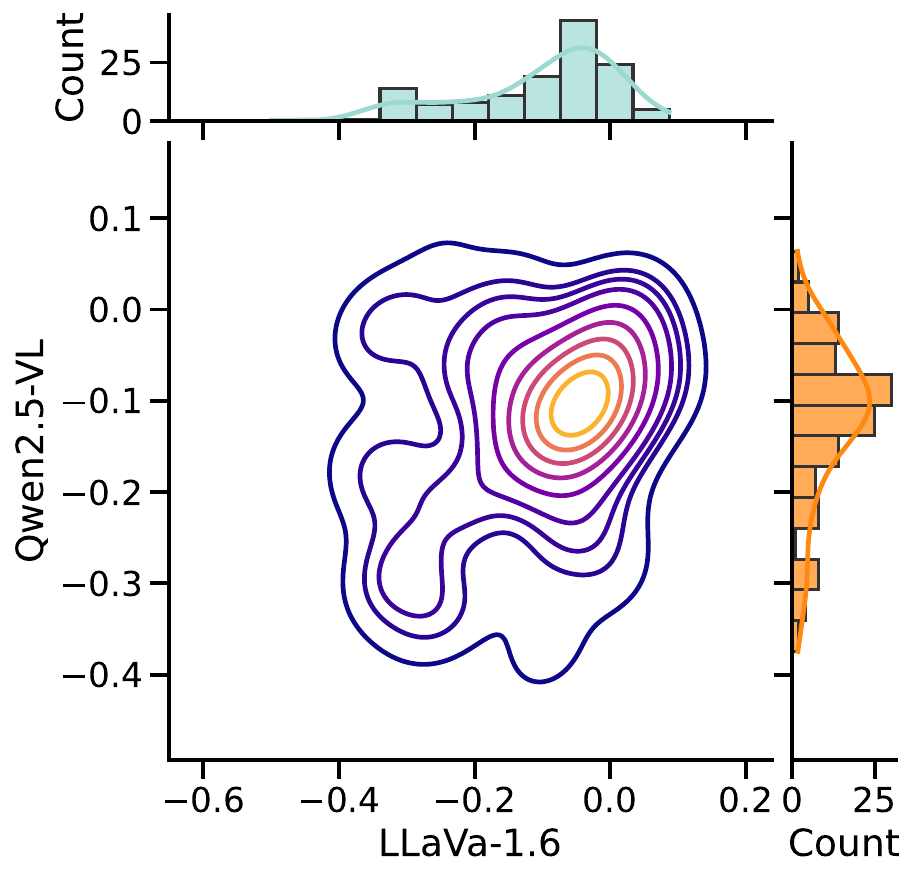}
    \end{subfigure}
    \hfill
    \begin{subfigure}[b]{0.245\textwidth}
        \centering
        \includegraphics[width=0.98\textwidth]{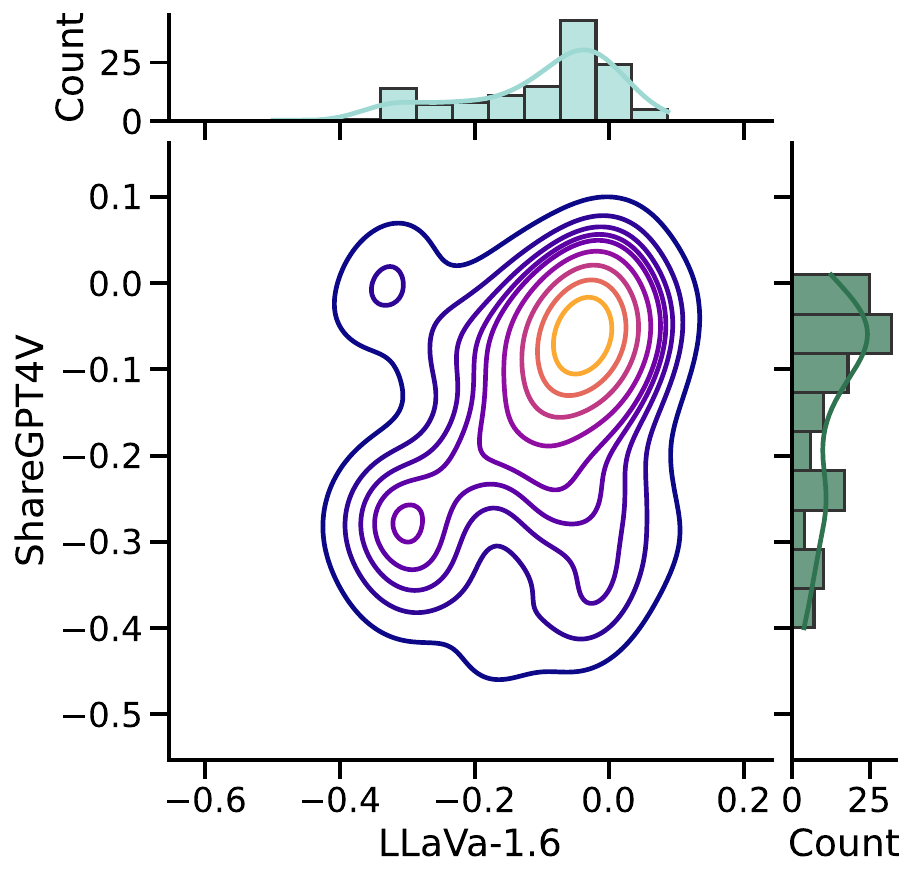}
    \end{subfigure}
    % new line
    \begin{subfigure}[b]{0.245\textwidth}
        \centering
        \includegraphics[width=0.98\textwidth]{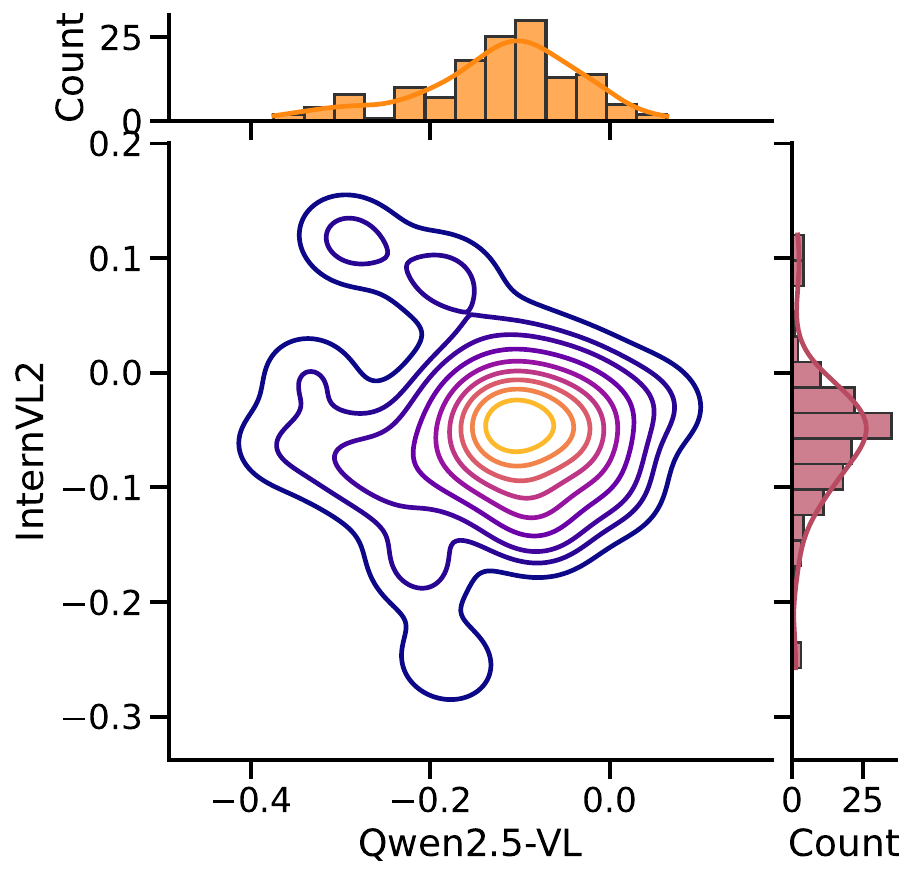}
    \end{subfigure}
    \hfill
    \begin{subfigure}[b]{0.245\textwidth}
        \centering
        \includegraphics[width=0.98\textwidth]{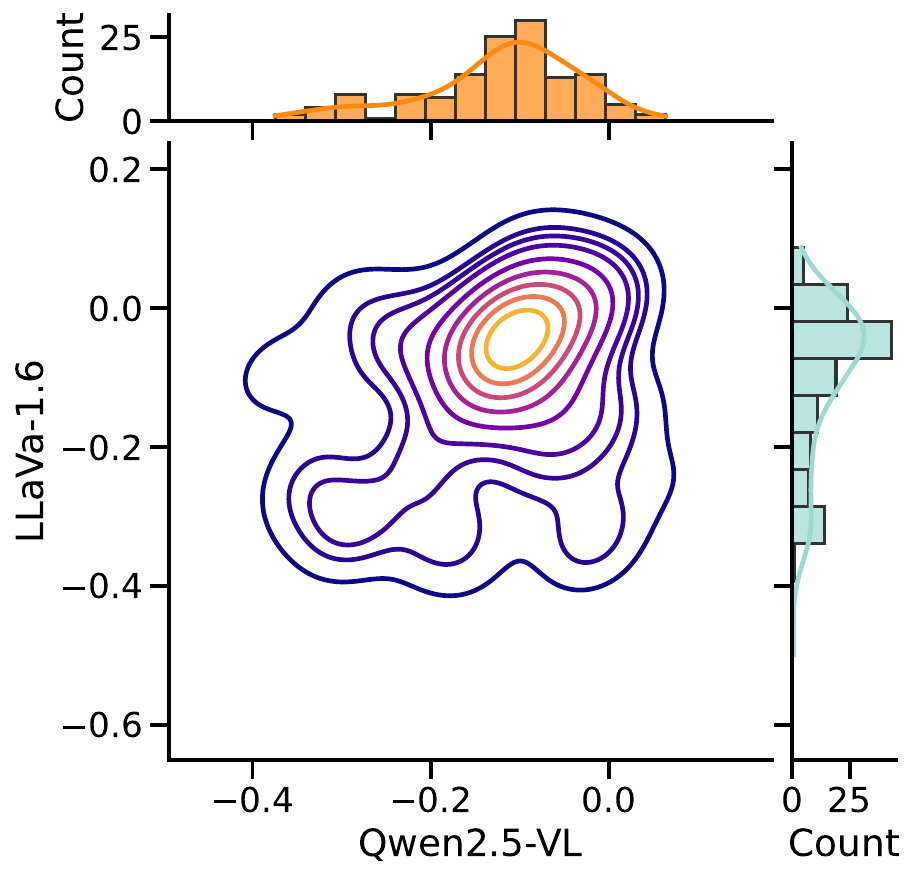}
    \end{subfigure}
    \hfill
    \begin{subfigure}[b]{0.245\textwidth}
        \centering
        \includegraphics[width=0.98\textwidth]{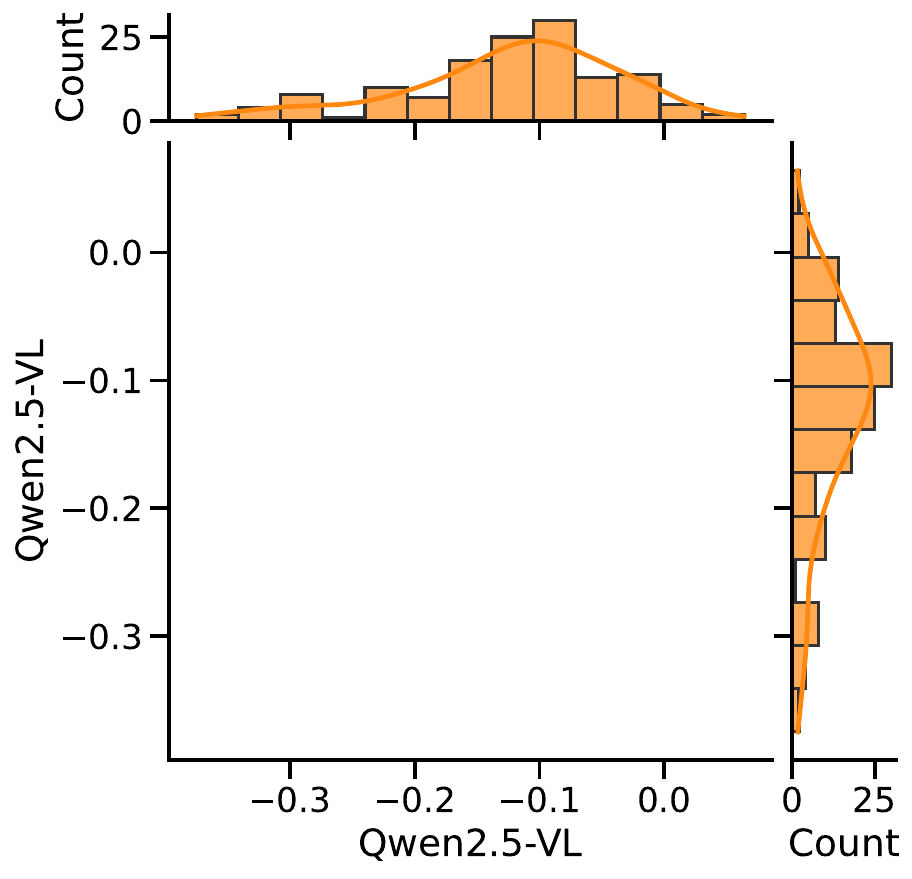}
    \end{subfigure}
    \hfill
    \begin{subfigure}[b]{0.245\textwidth}
        \centering
        \includegraphics[width=0.98\textwidth]{imgs/distr_effects/mmbench/qwen2_5_vl_sharegpt4v.pdf}
    \end{subfigure}
    % new line
    \begin{subfigure}[b]{0.245\textwidth}
        \centering
        \includegraphics[width=0.98\textwidth]{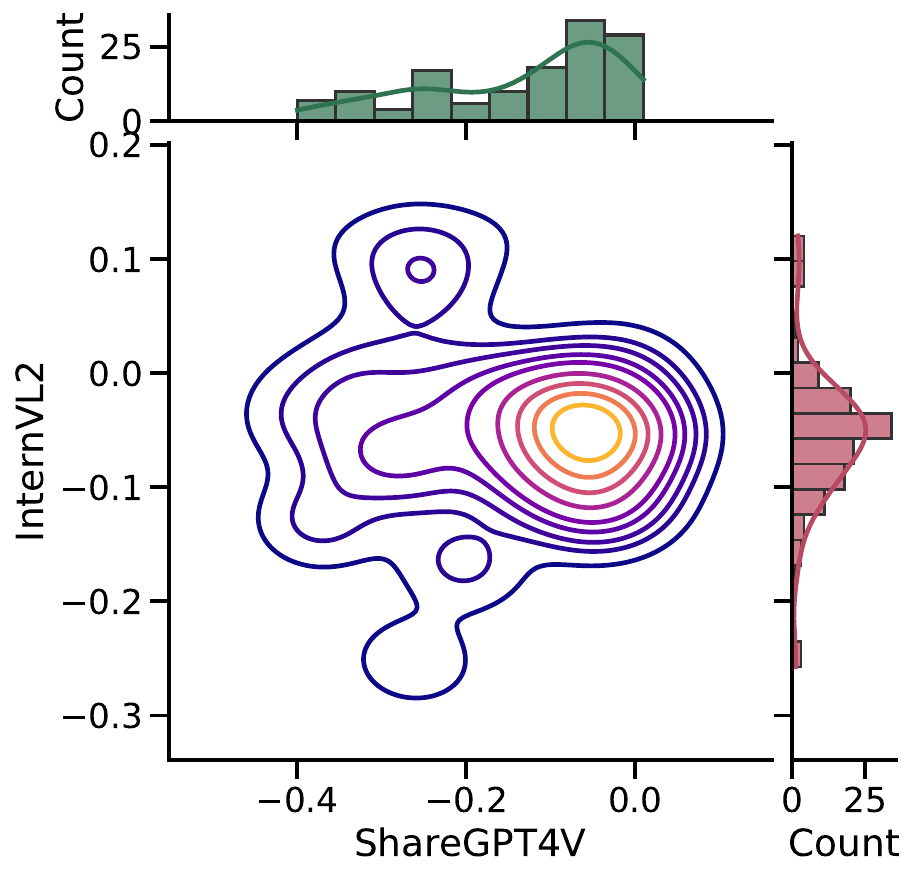}
    \end{subfigure}
    \hfill
    \begin{subfigure}[b]{0.245\textwidth}
        \centering
        \includegraphics[width=0.98\textwidth]{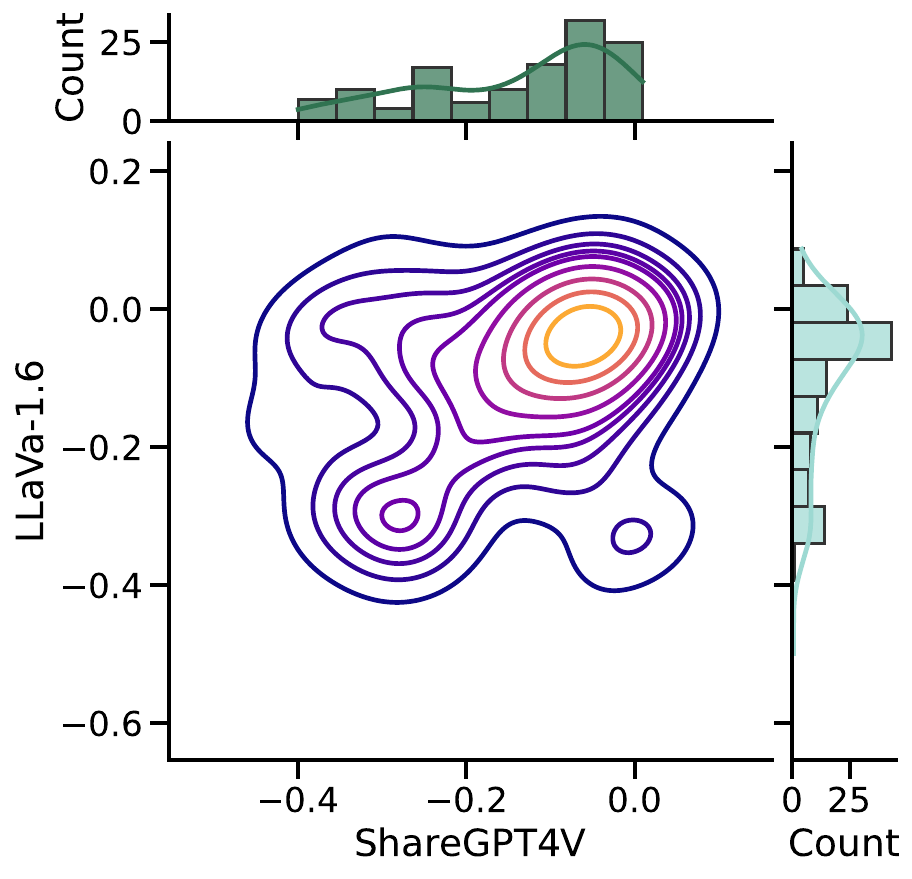}
    \end{subfigure}
    \hfill
    \begin{subfigure}[b]{0.245\textwidth}
        \centering
        \includegraphics[width=0.98\textwidth]{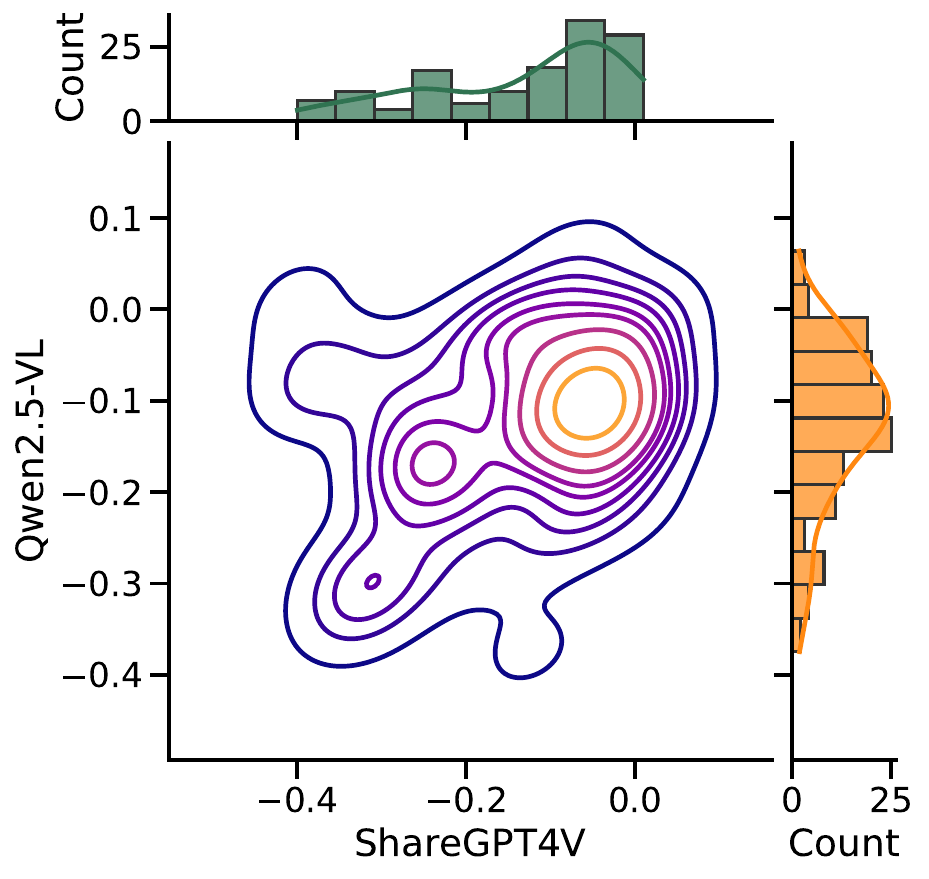}
    \end{subfigure}
    \hfill
    \begin{subfigure}[b]{0.245\textwidth}
        \centering
        \includegraphics[width=0.98\textwidth]{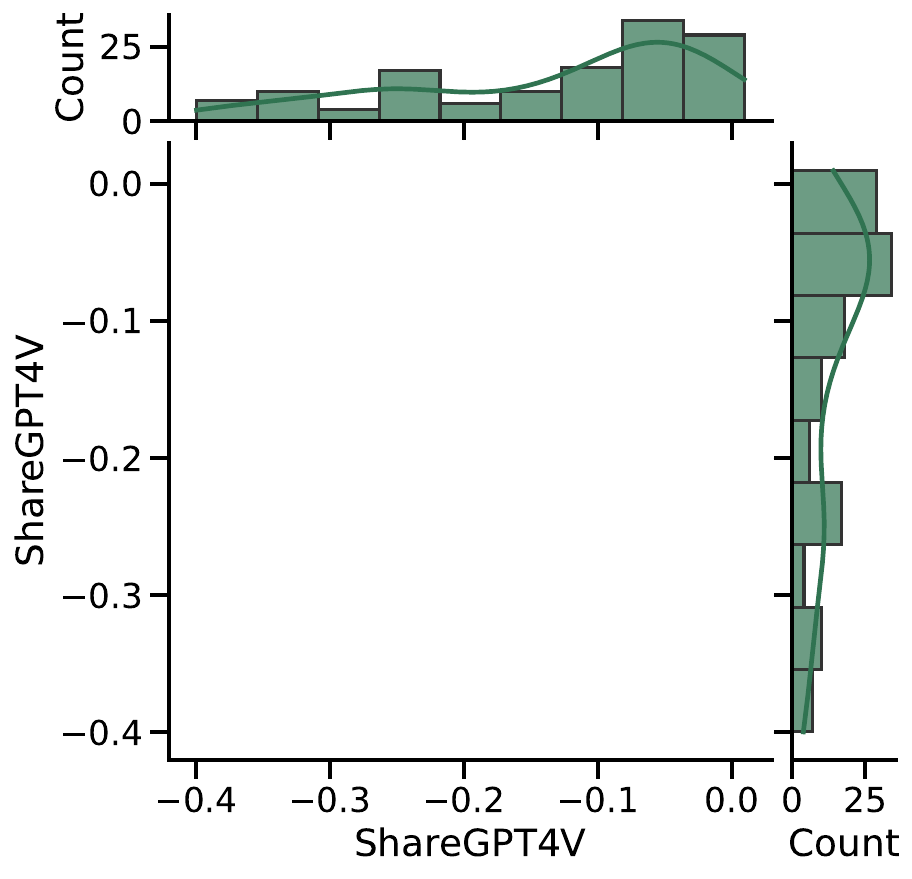}
    \end{subfigure}
    \caption{Model-pairwise distributions of estimated causal effects on MMBench \citep{Liu2024MMBench}.}
    \label{fig:effects_mmbench}
\end{figure*}

\begin{figure*}[t]
    \centering
    \begin{subfigure}[b]{0.245\textwidth}
        \centering
        \includegraphics[width=0.98\textwidth]{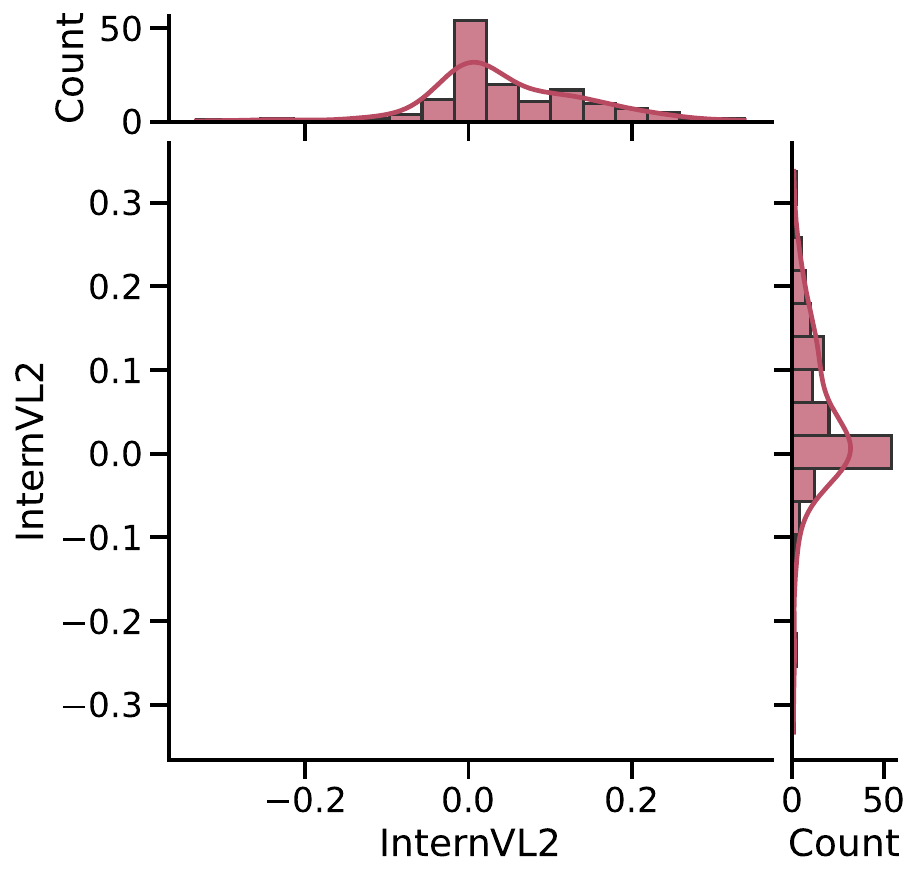}
    \end{subfigure}
    \hfill
    \begin{subfigure}[b]{0.245\textwidth}
        \centering
        \includegraphics[width=0.98\textwidth]{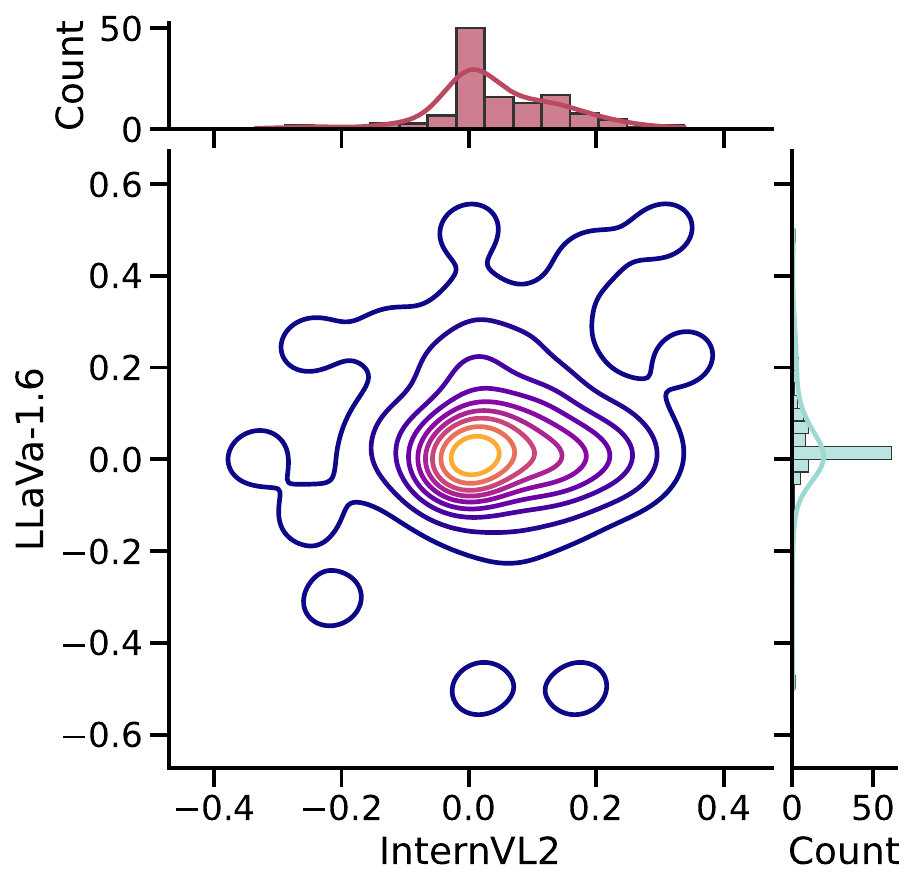}
    \end{subfigure}
    \hfill
    \begin{subfigure}[b]{0.245\textwidth}
        \centering
        \includegraphics[width=0.98\textwidth]{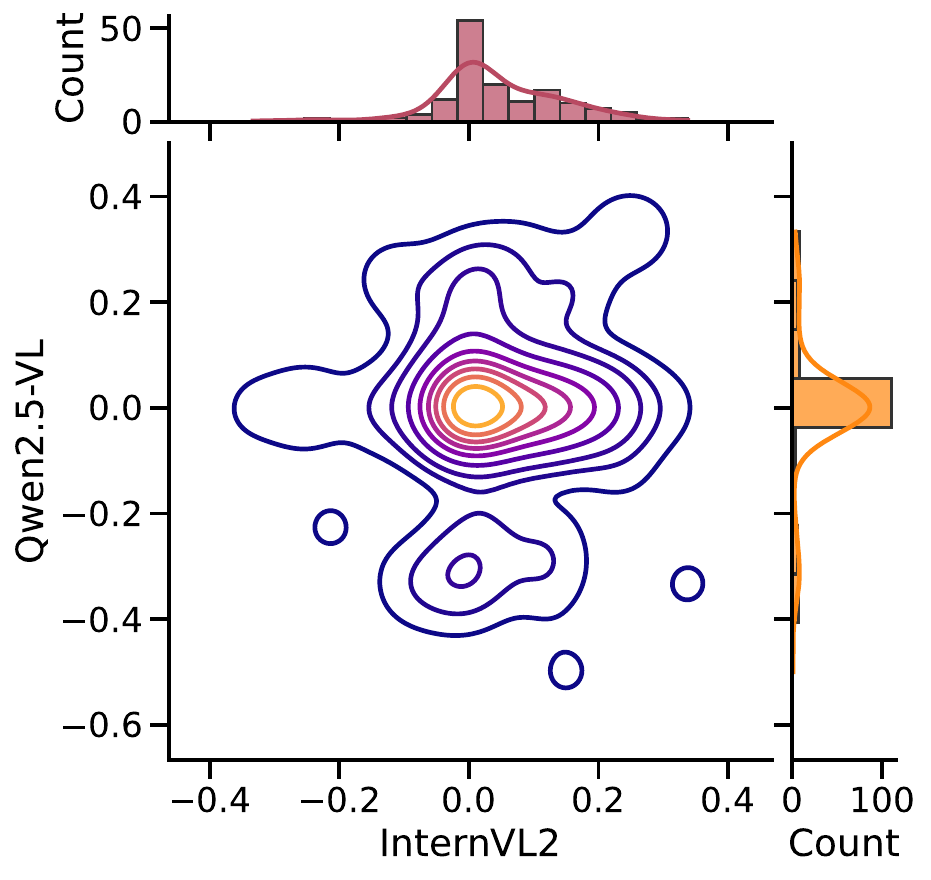}
    \end{subfigure}
    \hfill
    \begin{subfigure}[b]{0.245\textwidth}
        \centering
        \includegraphics[width=0.98\textwidth]{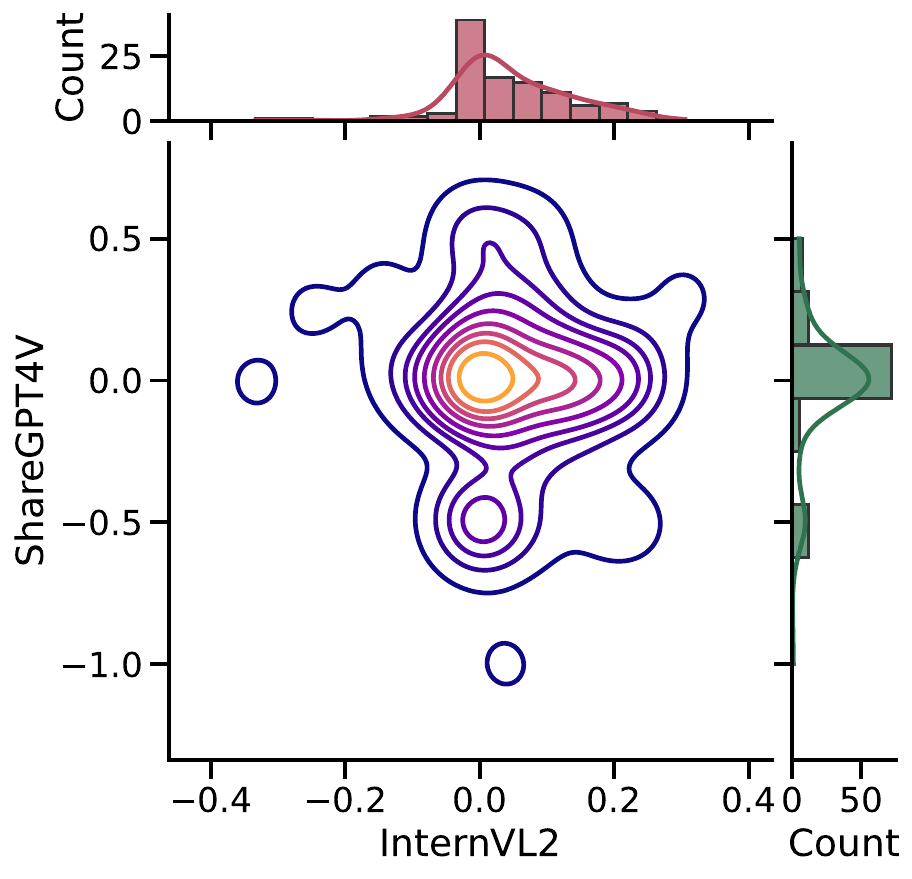}
    \end{subfigure}
    % new line
    \begin{subfigure}[b]{0.245\textwidth}
        \centering
        \includegraphics[width=0.98\textwidth]{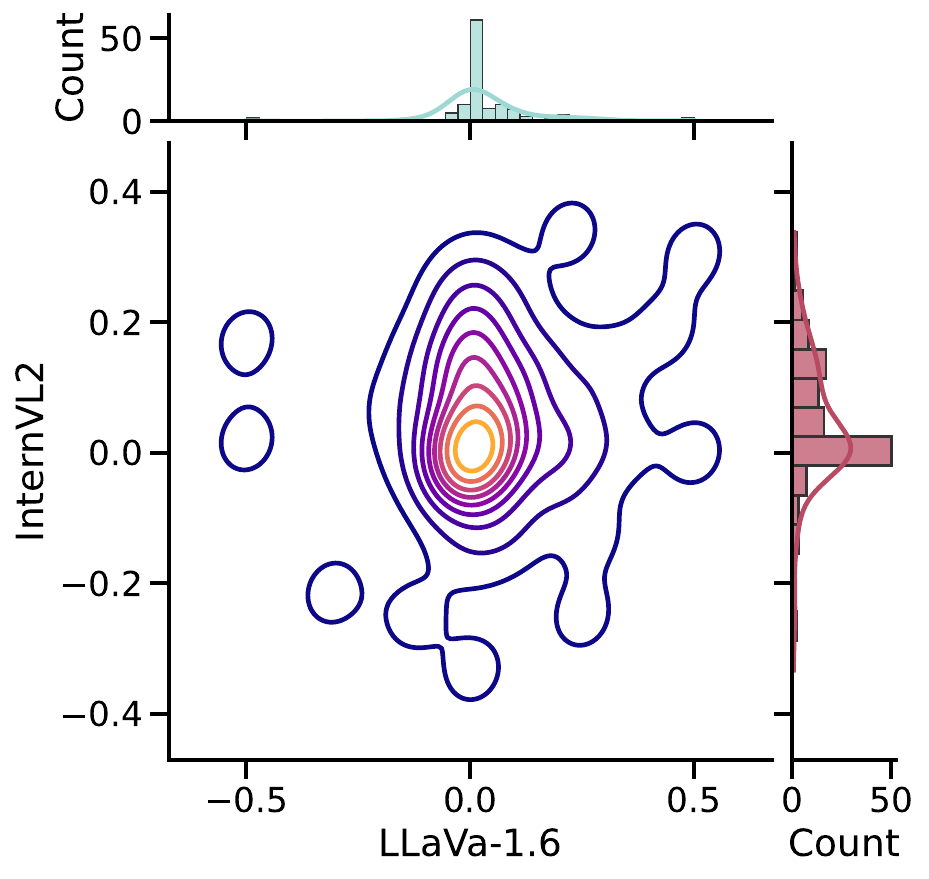}
    \end{subfigure}
    \hfill
    \begin{subfigure}[b]{0.245\textwidth}
        \centering
        \includegraphics[width=0.98\textwidth]{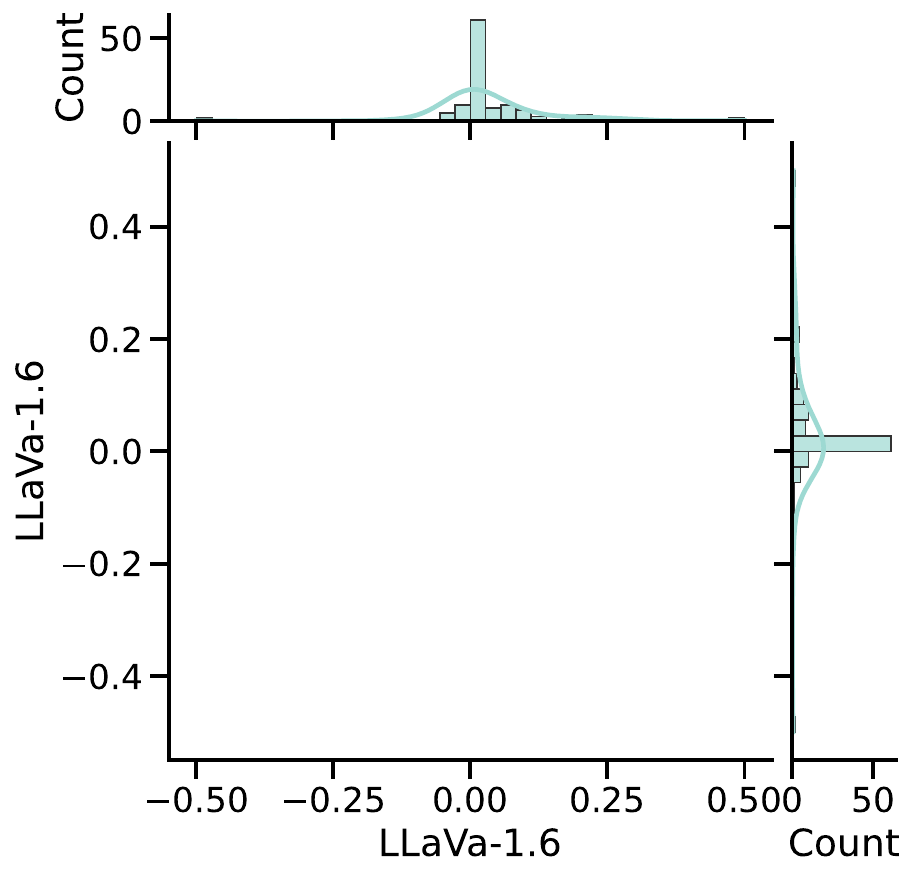}
    \end{subfigure}
    \hfill
    \begin{subfigure}[b]{0.245\textwidth}
        \centering
        \includegraphics[width=0.98\textwidth]{imgs/distr_effects/seed/llava-1.6_qwen2_5_vl.pdf}
    \end{subfigure}
    \hfill
    \begin{subfigure}[b]{0.245\textwidth}
        \centering
        \includegraphics[width=0.98\textwidth]{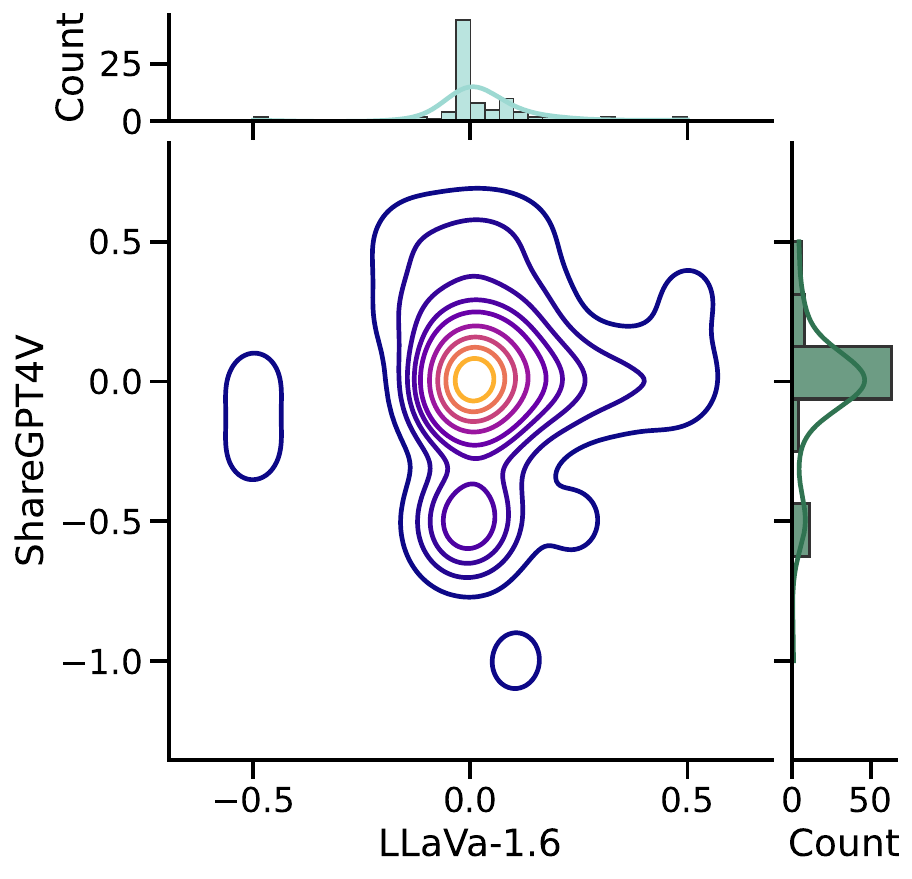}
    \end{subfigure}
    % new line
    \begin{subfigure}[b]{0.245\textwidth}
        \centering
        \includegraphics[width=0.98\textwidth]{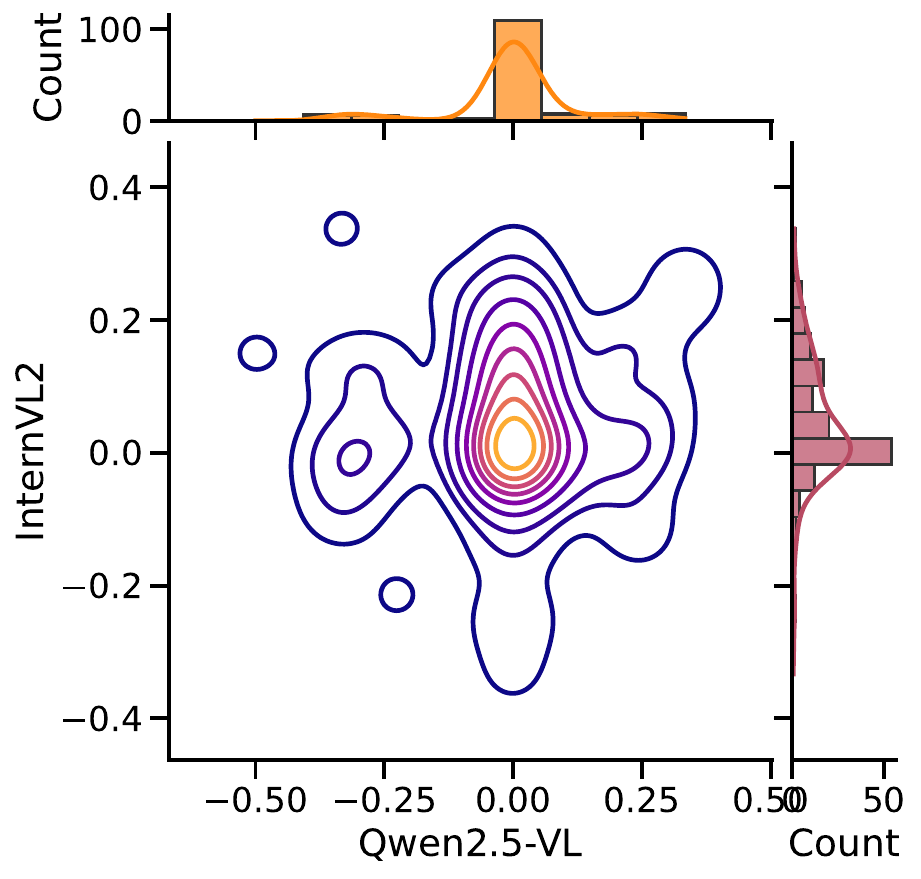}
    \end{subfigure}
    \hfill
    \begin{subfigure}[b]{0.245\textwidth}
        \centering
        \includegraphics[width=0.98\textwidth]{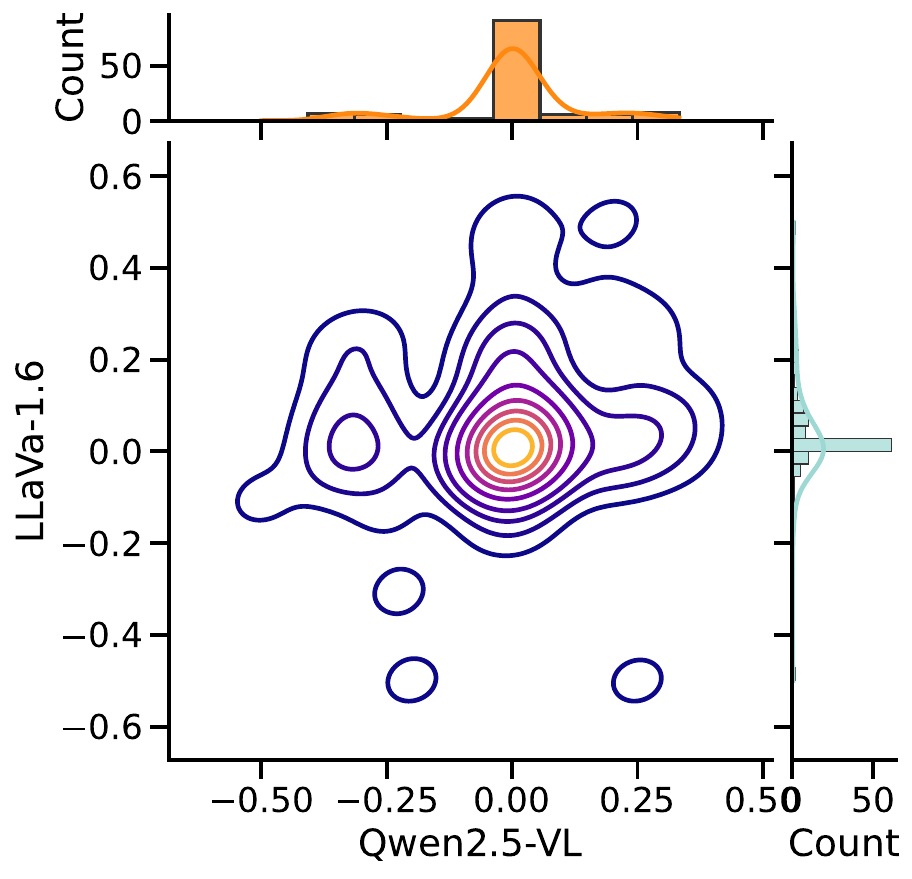}
    \end{subfigure}
    \hfill
    \begin{subfigure}[b]{0.245\textwidth}
        \centering
        \includegraphics[width=0.98\textwidth]{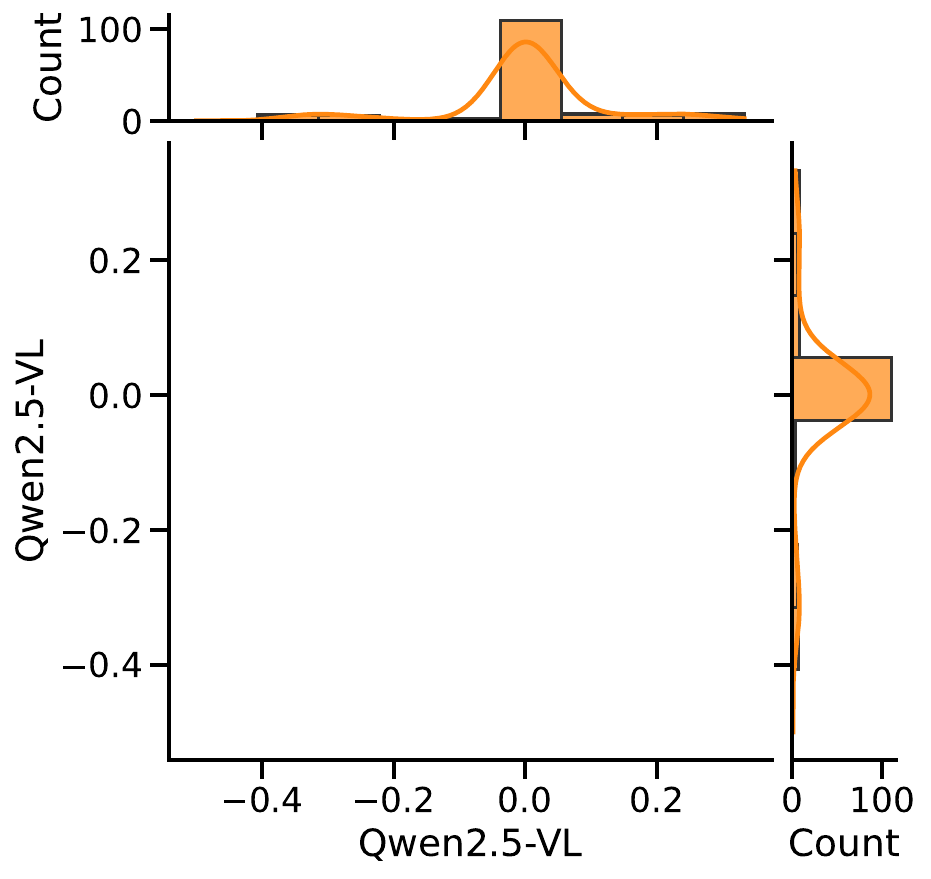}
    \end{subfigure}
    \hfill
    \begin{subfigure}[b]{0.245\textwidth}
        \centering
        \includegraphics[width=0.98\textwidth]{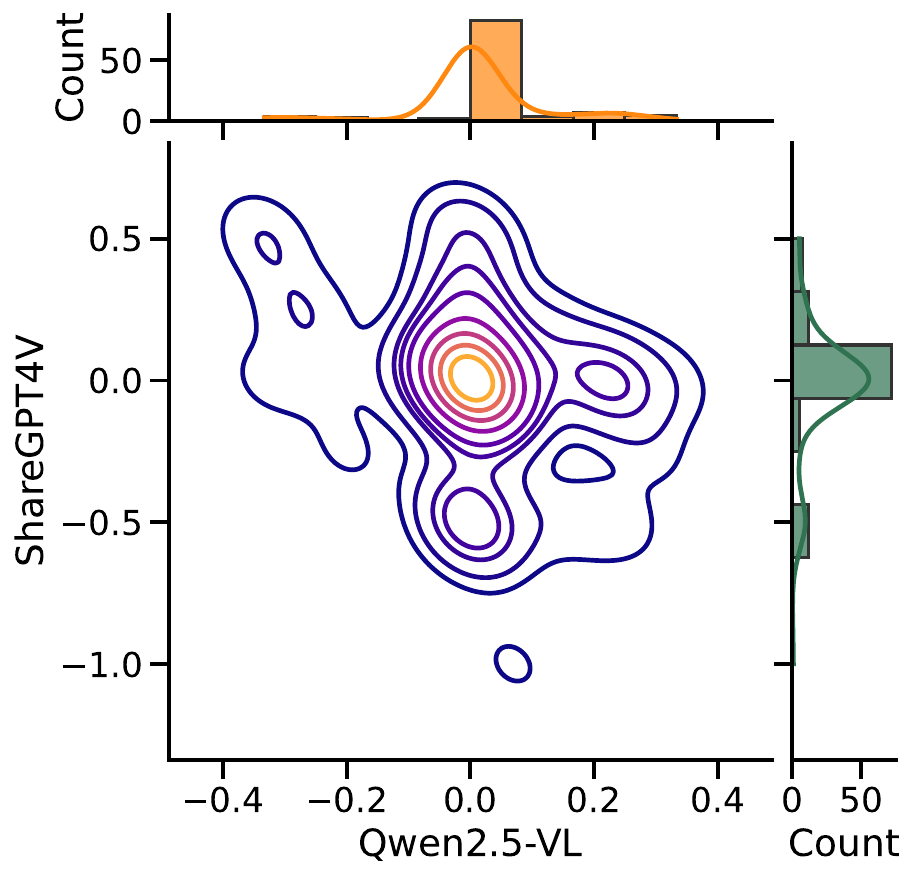}
    \end{subfigure}
    % new line
    \begin{subfigure}[b]{0.245\textwidth}
        \centering
        \includegraphics[width=0.98\textwidth]{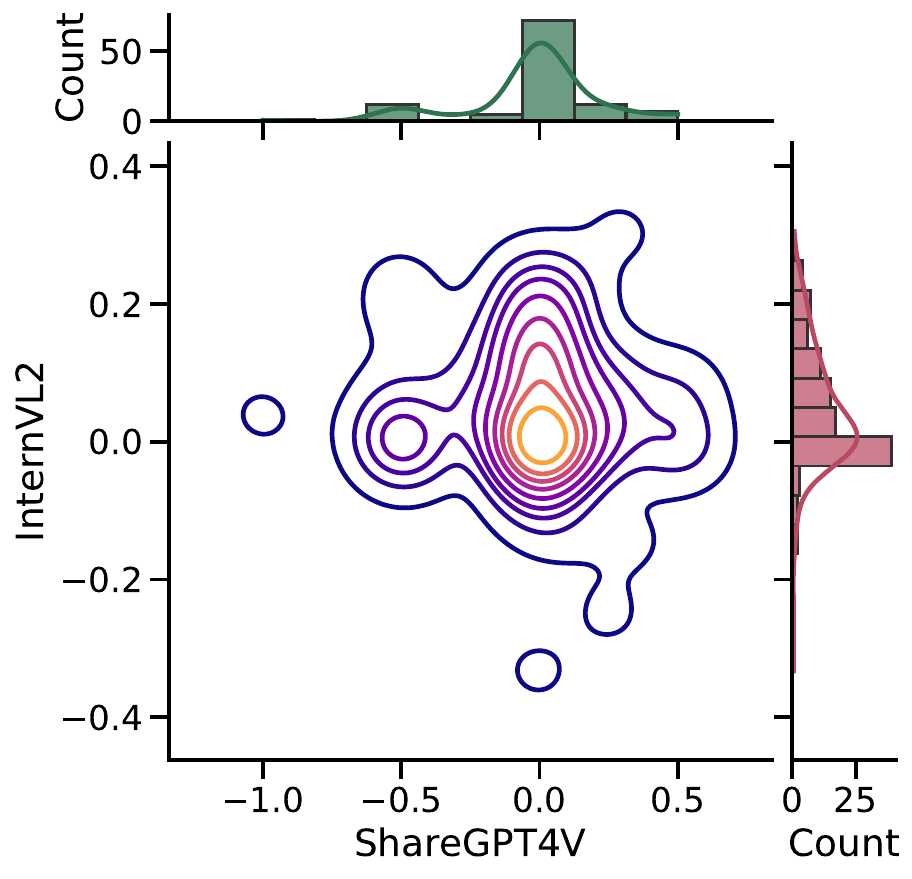}
    \end{subfigure}
    \hfill
    \begin{subfigure}[b]{0.245\textwidth}
        \centering
        \includegraphics[width=0.98\textwidth]{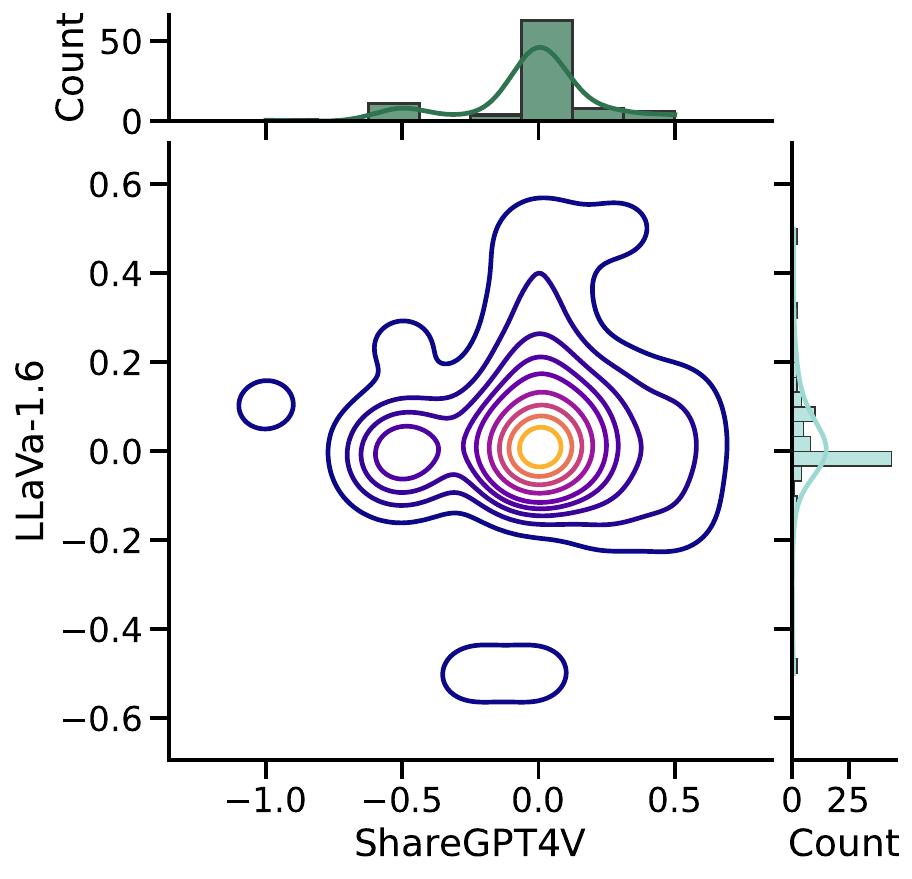}
    \end{subfigure}
    \hfill
    \begin{subfigure}[b]{0.245\textwidth}
        \centering
        \includegraphics[width=0.98\textwidth]{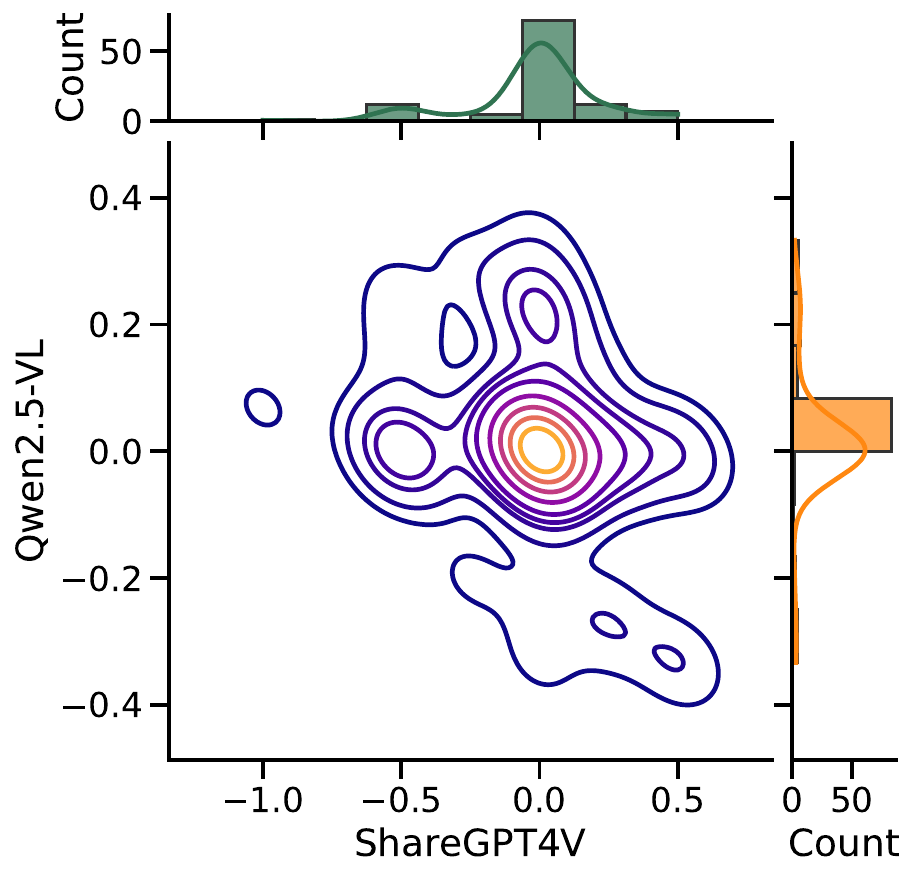}
    \end{subfigure}
    \hfill
    \begin{subfigure}[b]{0.245\textwidth}
        \centering
        \includegraphics[width=0.98\textwidth]{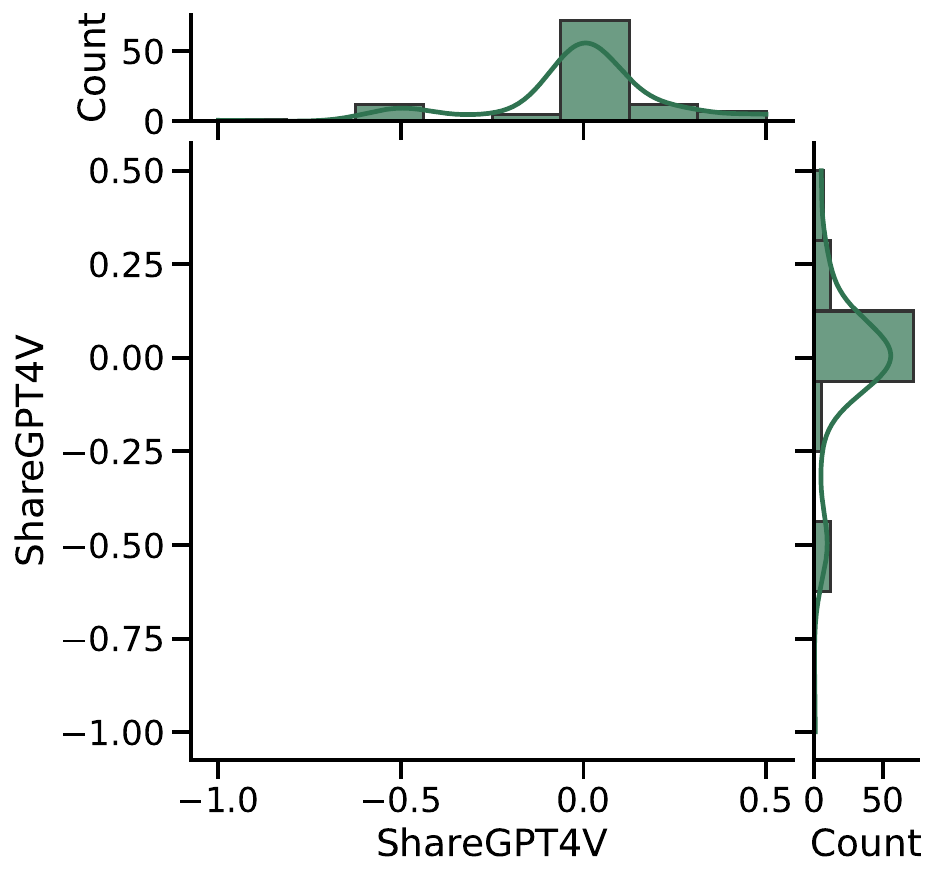}
    \end{subfigure}
    \caption{Model-pairwise distributions of estimated causal effects on SEED-Bench-2 \citep{Li2023SEEDBench}.}
    \label{fig:effects_seed}
\end{figure*}

\begin{figure*}[t]
    \centering
    \begin{subfigure}[b]{0.245\textwidth}
        \centering
        \includegraphics[width=0.97\textwidth]{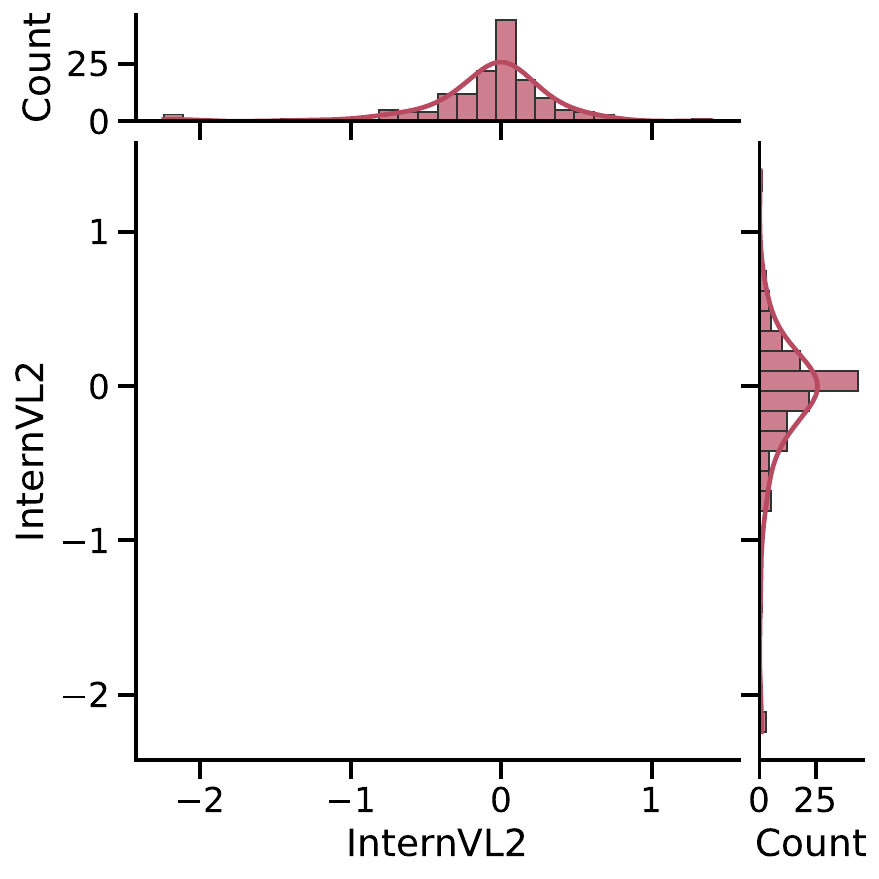}
    \end{subfigure}
    \hfill
    \begin{subfigure}[b]{0.245\textwidth}
        \centering
        \includegraphics[width=0.97\textwidth]{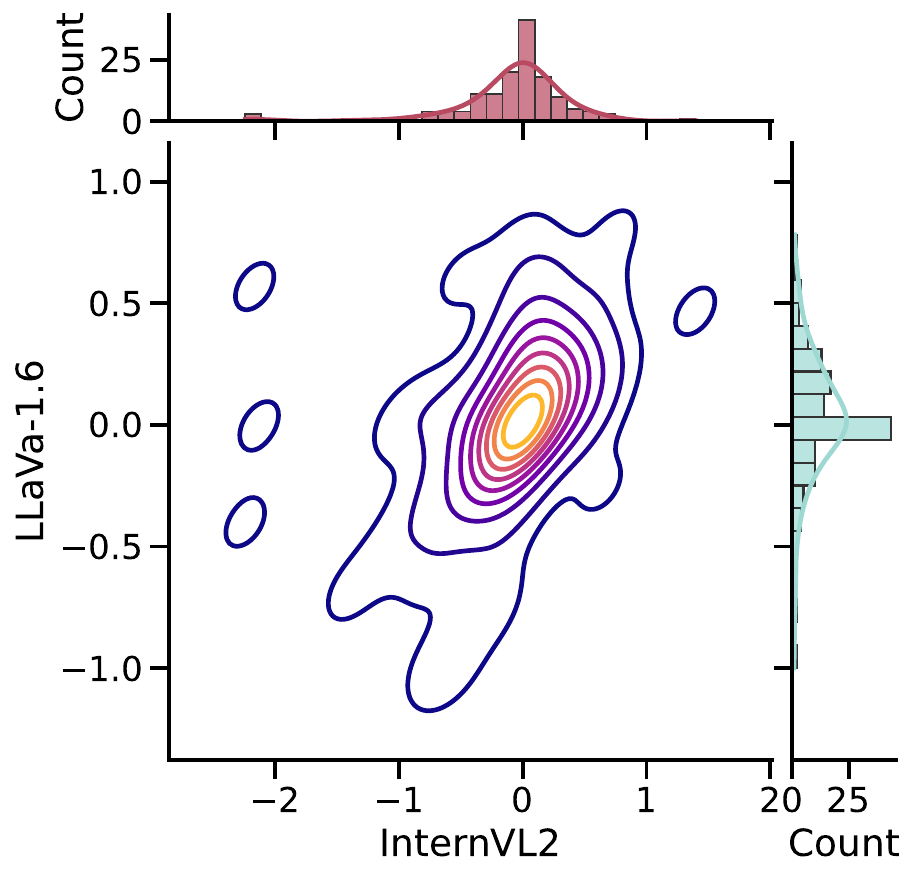}
    \end{subfigure}
    \hfill
    \begin{subfigure}[b]{0.245\textwidth}
        \centering
        \includegraphics[width=0.97\textwidth]{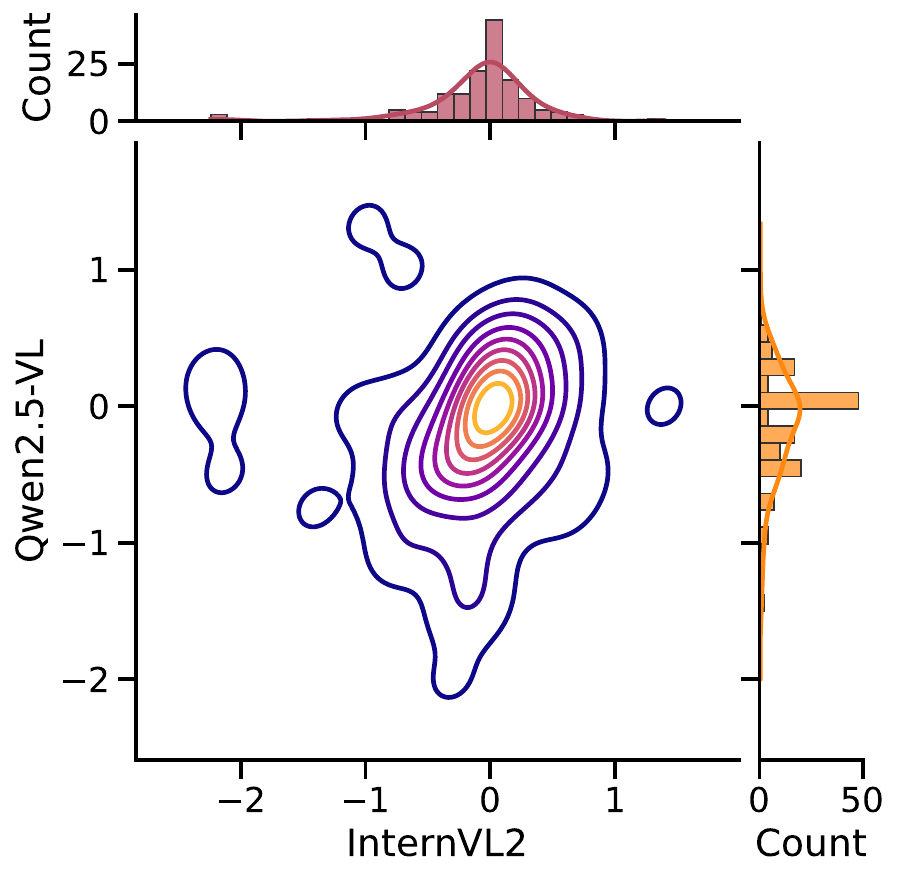}
    \end{subfigure}
    \hfill
    \begin{subfigure}[b]{0.245\textwidth}
        \centering
        \includegraphics[width=0.97\textwidth]{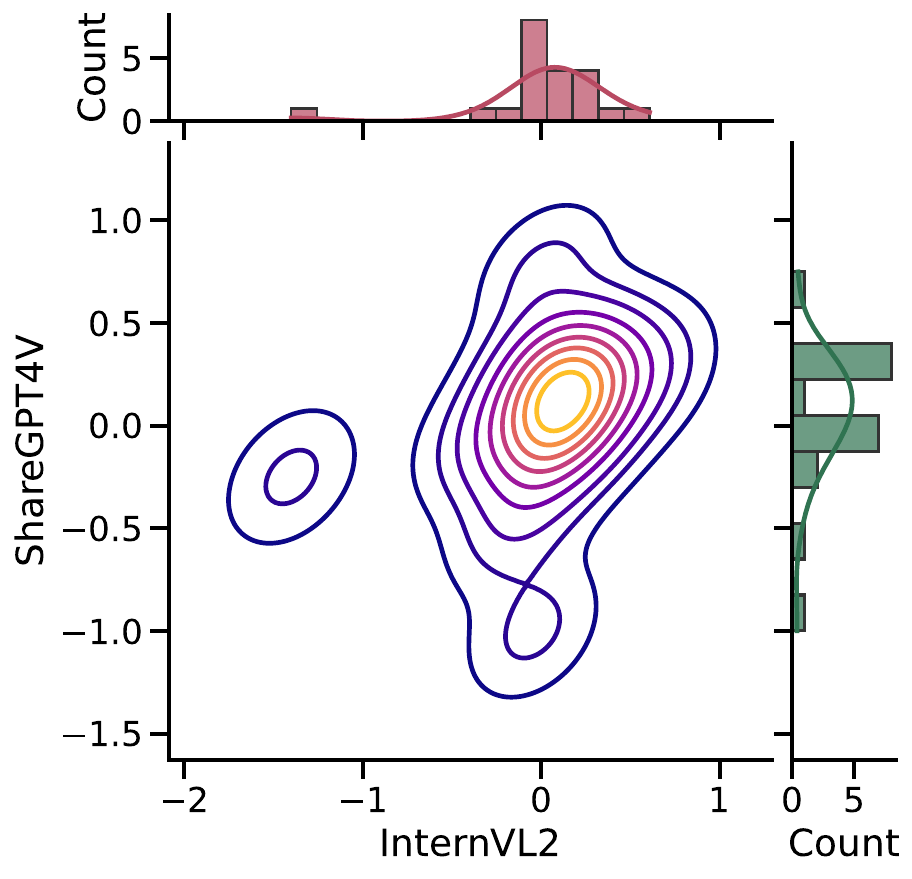}
    \end{subfigure}
    % new line
    \begin{subfigure}[b]{0.245\textwidth}
        \centering
        \includegraphics[width=0.97\textwidth]{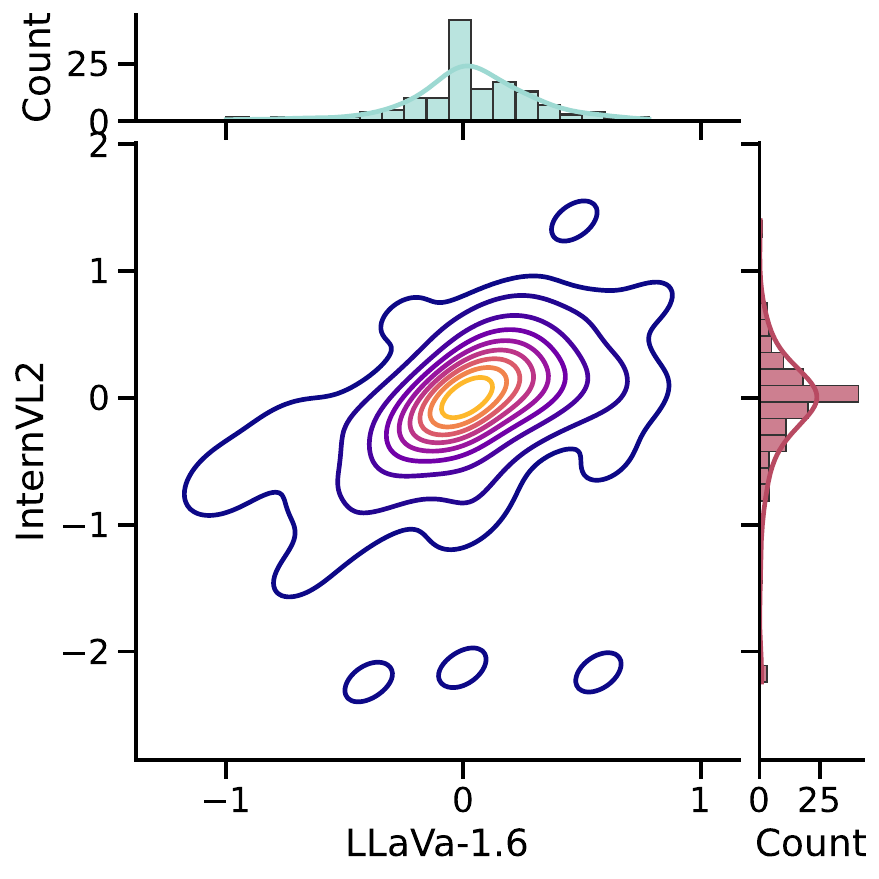}
    \end{subfigure}
    \hfill
    \begin{subfigure}[b]{0.245\textwidth}
        \centering
        \includegraphics[width=0.97\textwidth]{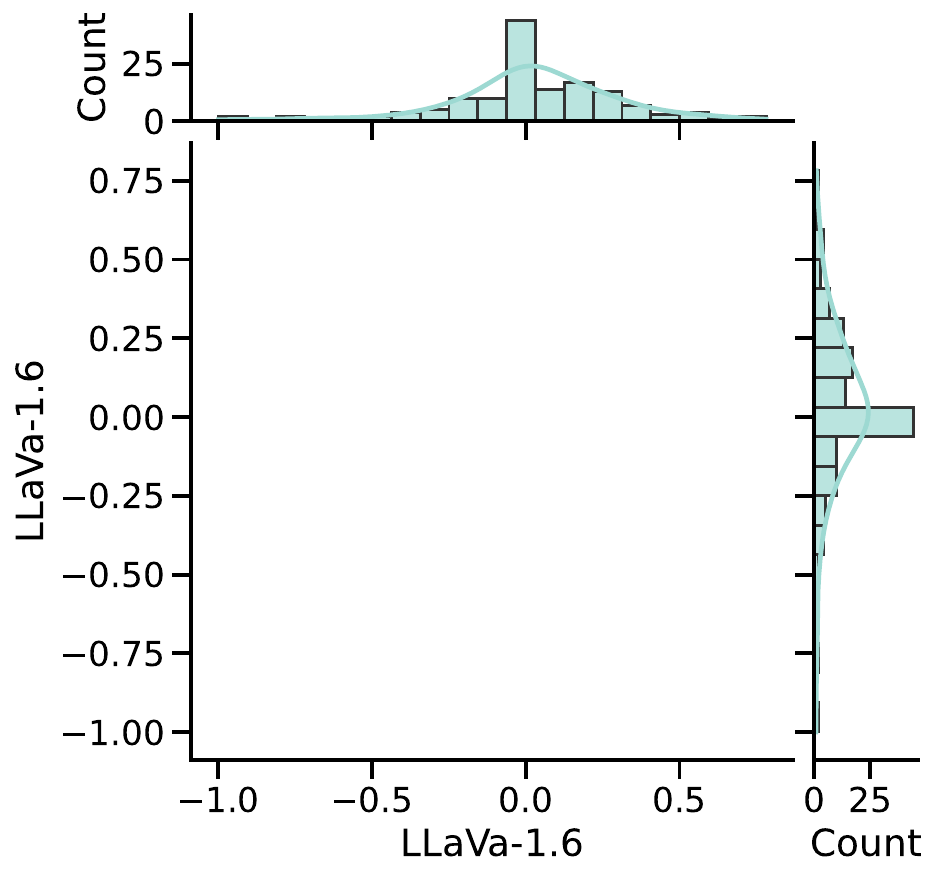}
    \end{subfigure}
    \hfill
    \begin{subfigure}[b]{0.245\textwidth}
        \centering
        \includegraphics[width=0.97\textwidth]{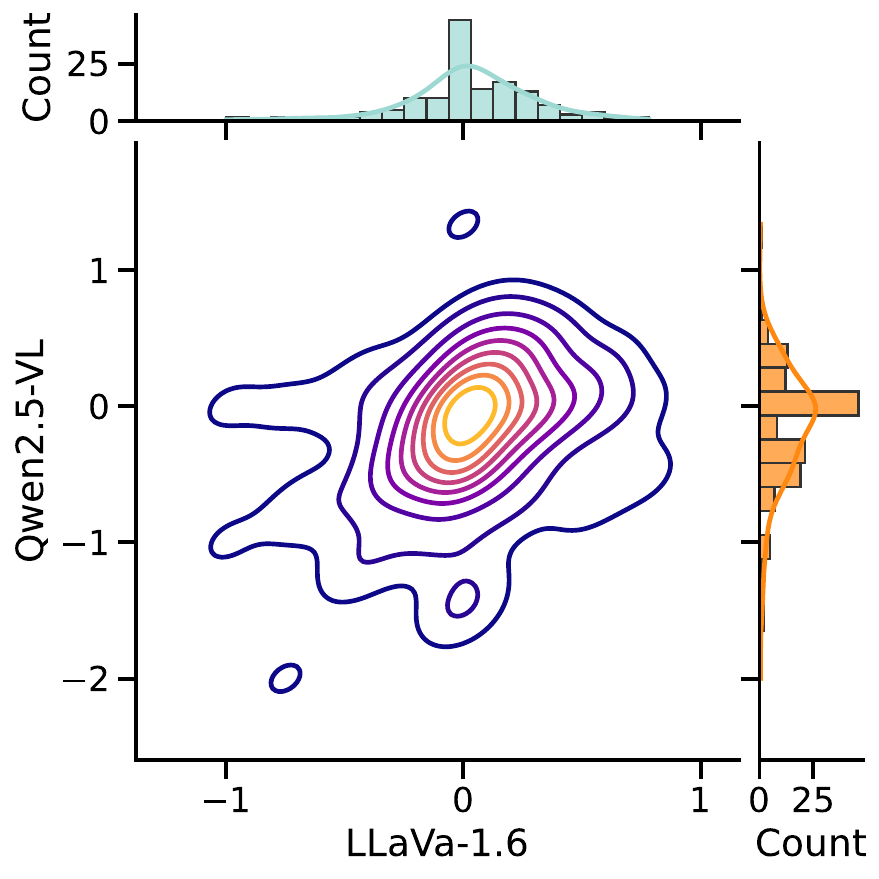}
    \end{subfigure}
    \hfill
    \begin{subfigure}[b]{0.245\textwidth}
        \centering
        \includegraphics[width=0.97\textwidth]{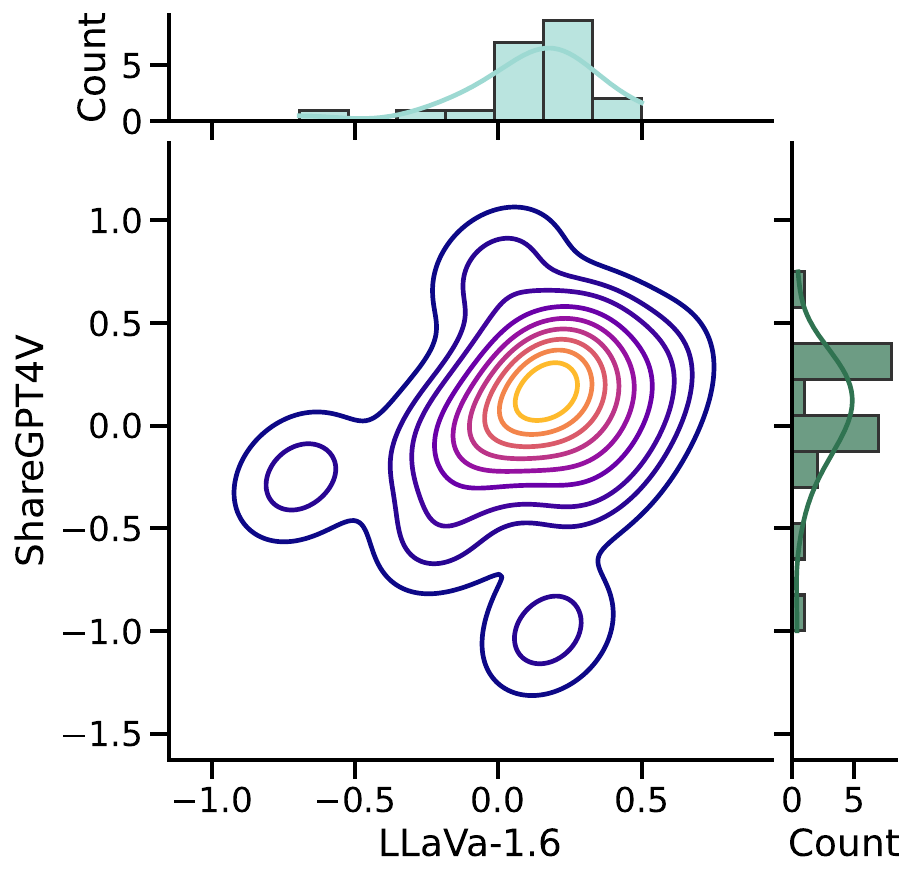}
    \end{subfigure}
    % new line
    \begin{subfigure}[b]{0.245\textwidth}
        \centering
        \includegraphics[width=0.97\textwidth]{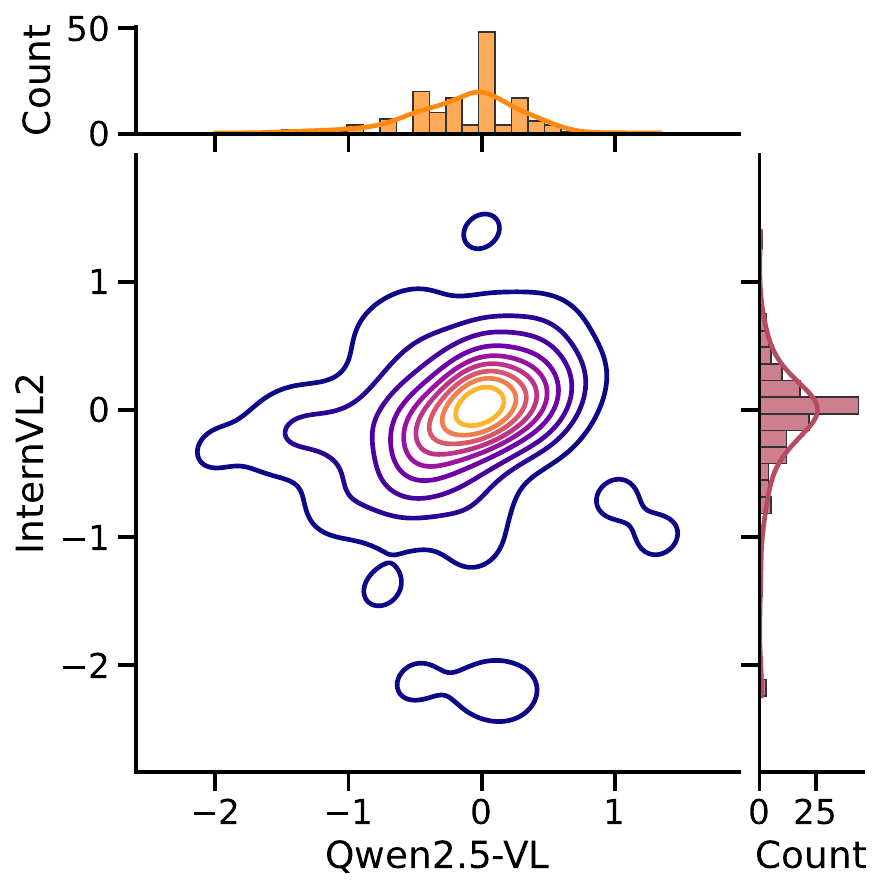}
    \end{subfigure}
    \hfill
    \begin{subfigure}[b]{0.245\textwidth}
        \centering
        \includegraphics[width=0.97\textwidth]{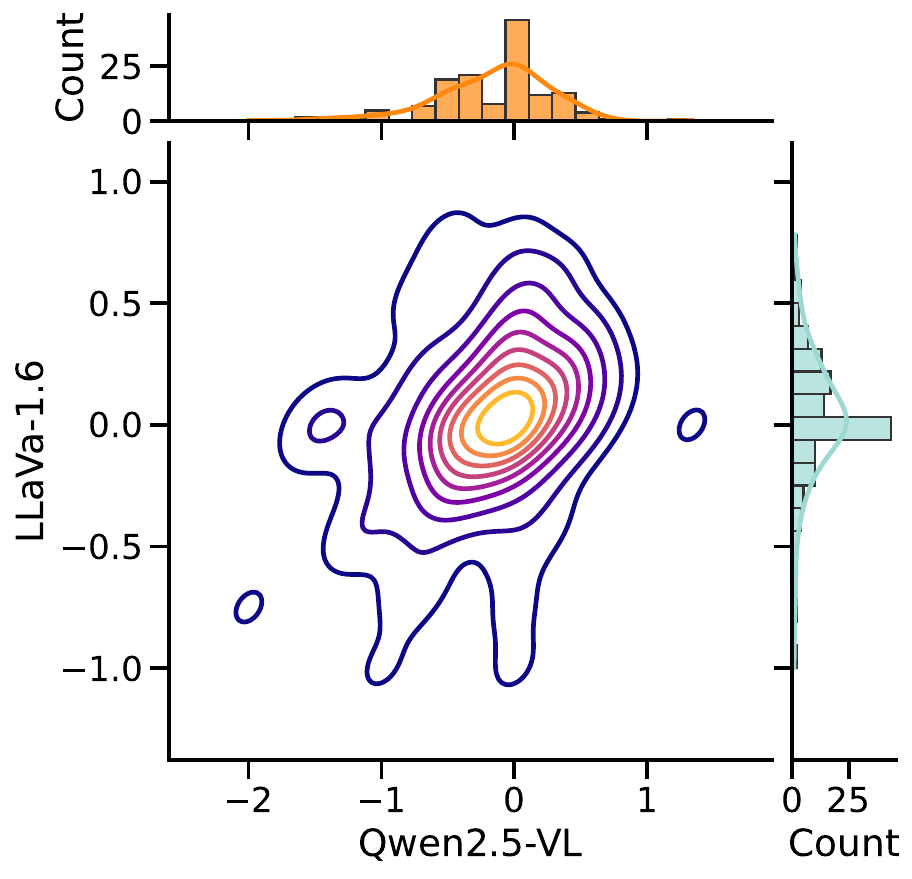}
    \end{subfigure}
    \hfill
    \begin{subfigure}[b]{0.245\textwidth}
        \centering
        \includegraphics[width=0.97\textwidth]{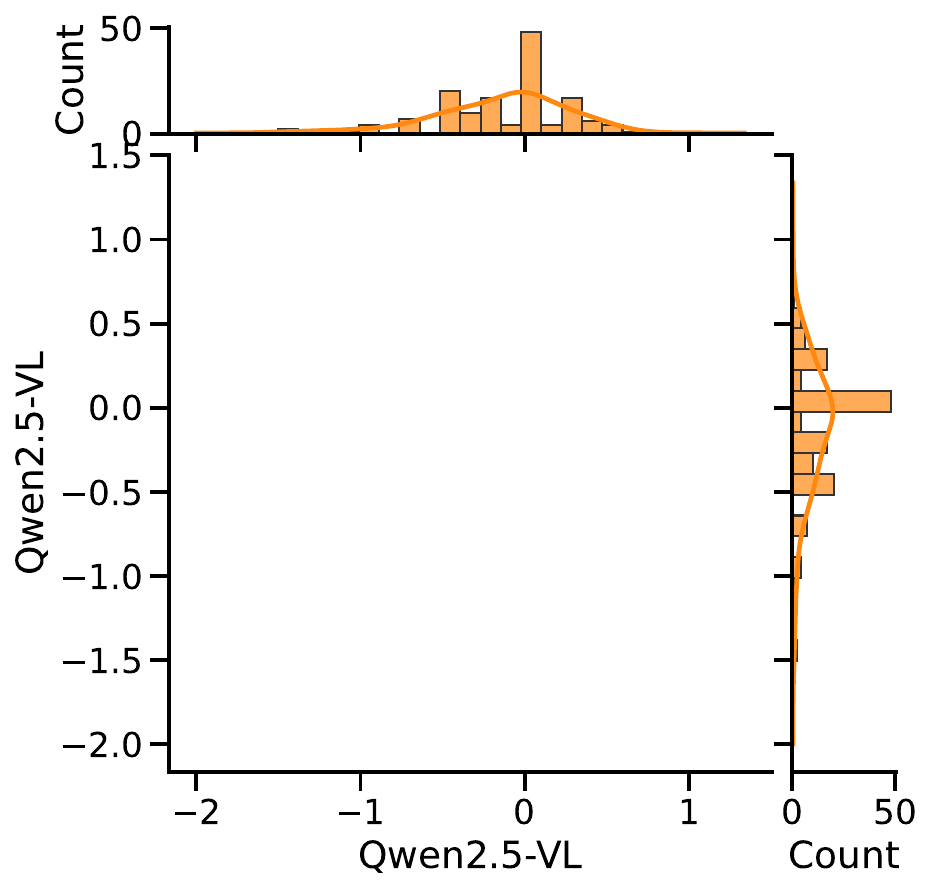}
    \end{subfigure}
    \hfill
    \begin{subfigure}[b]{0.245\textwidth}
        \centering
        \includegraphics[width=0.97\textwidth]{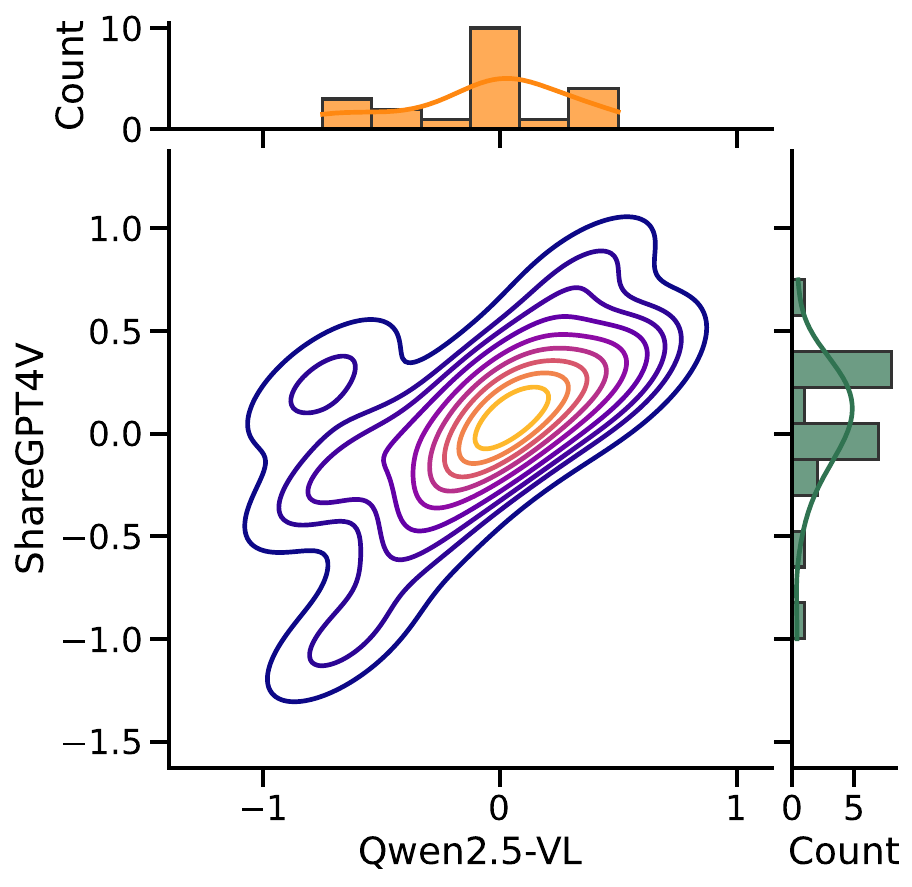}
    \end{subfigure}
    % new line
    \begin{subfigure}[b]{0.245\textwidth}
        \centering
        \includegraphics[width=0.97\textwidth]{imgs/distr_effects/vqav2/sharegpt4v_internvl2.pdf}
    \end{subfigure}
    \hfill
    \begin{subfigure}[b]{0.245\textwidth}
        \centering
        \includegraphics[width=0.97\textwidth]{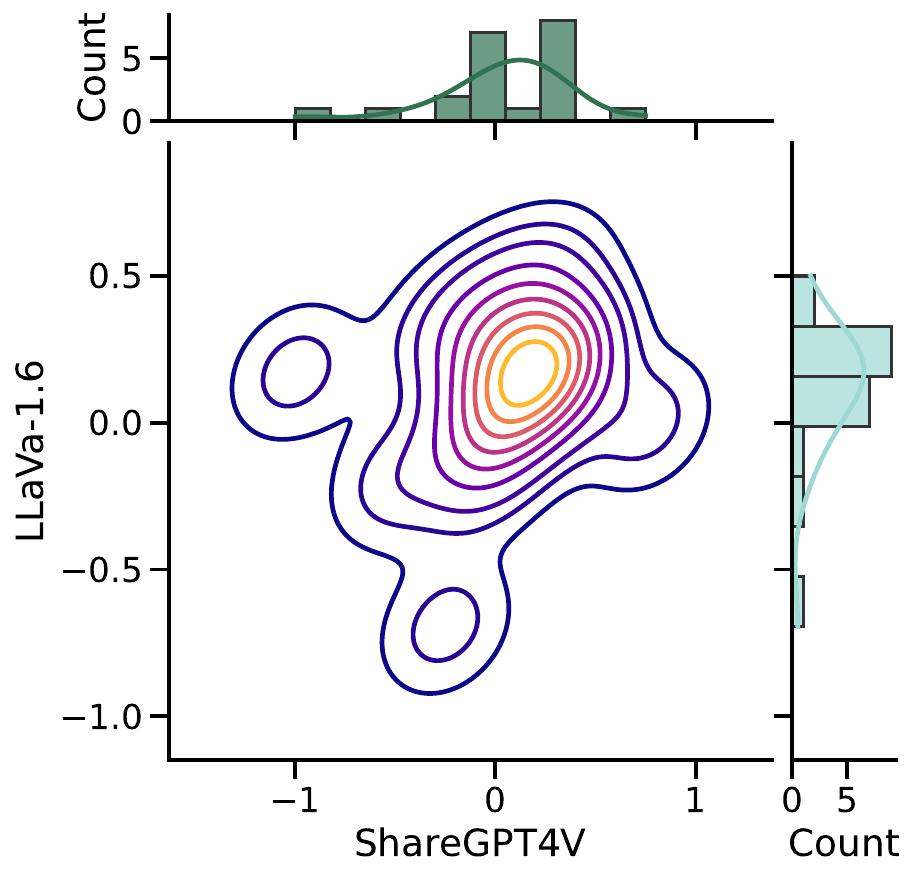}
    \end{subfigure}
    \hfill
    \begin{subfigure}[b]{0.245\textwidth}
        \centering
        \includegraphics[width=0.97\textwidth]{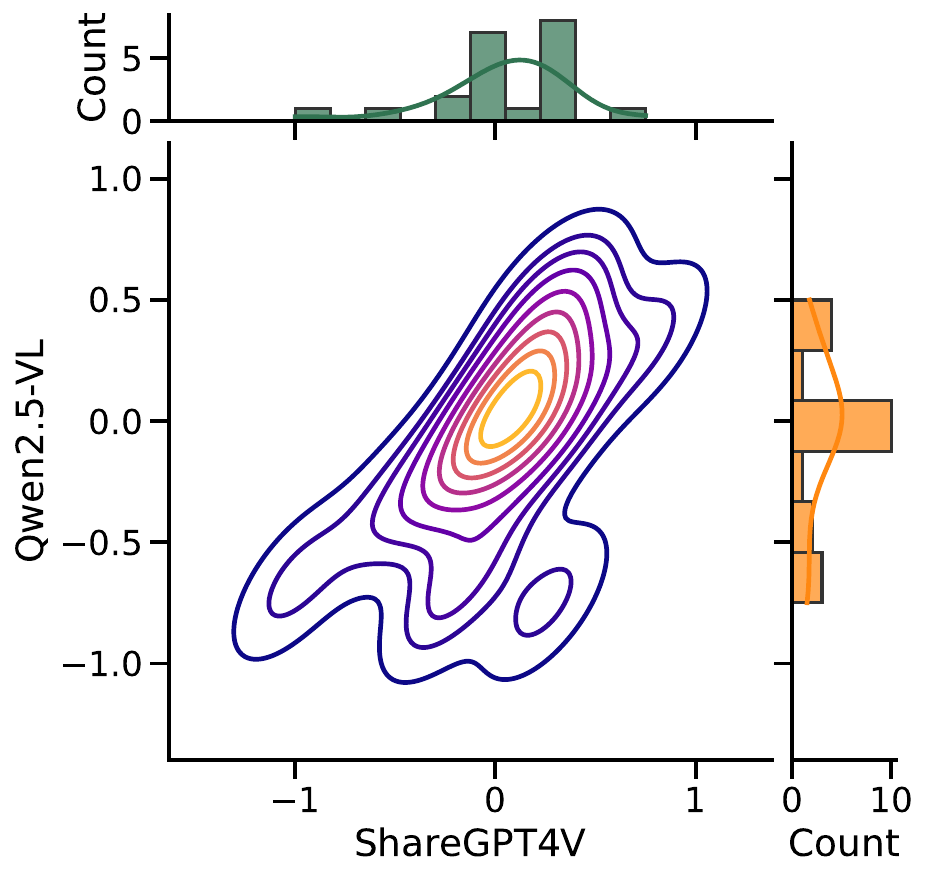}
    \end{subfigure}
    \hfill
    \begin{subfigure}[b]{0.245\textwidth}
        \centering
        \includegraphics[width=0.97\textwidth]{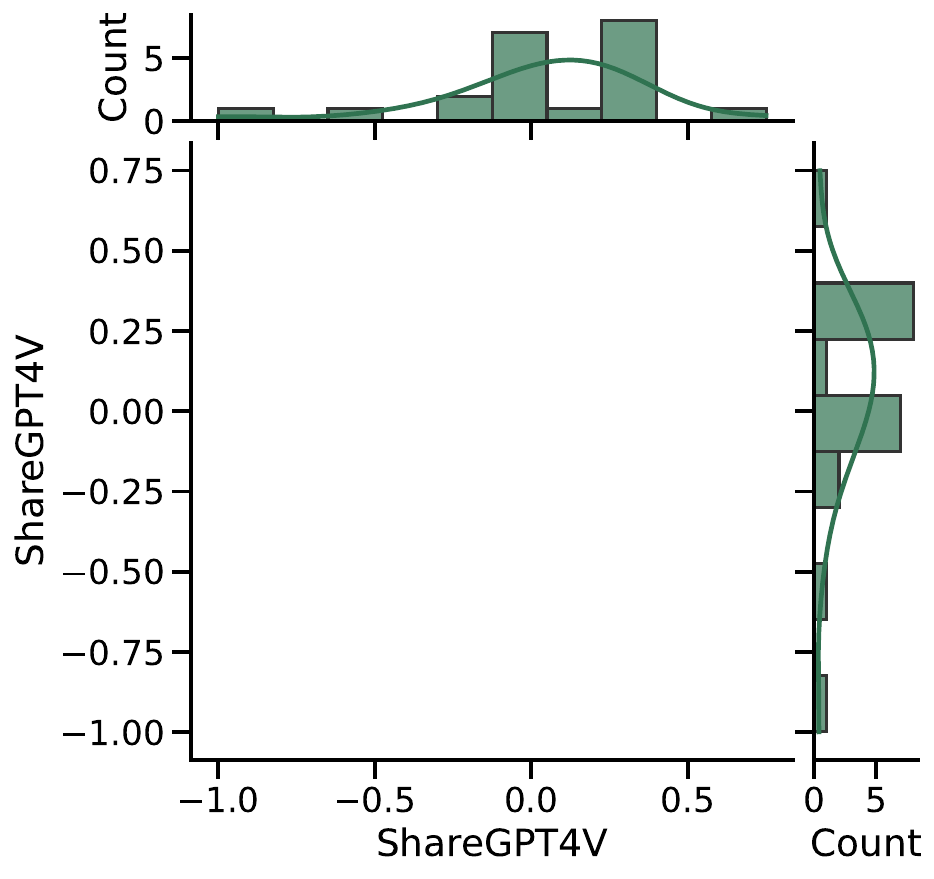}
    \end{subfigure}
    \caption{Model-pairwise distributions of estimated causal effects on VQA v2 \citep{Goyal2017VQAv2}.}
    \label{fig:effects_vqav2}
\end{figure*}

\begin{itemize}[leftmargin=*]
    \item Distributions on LLaVa-Bench: \autoref{fig:effects_llava_bench}
    \item Distributions on MMBench: \autoref{fig:effects_mmbench}
    \item Distributions on SEED-Bench-2: \autoref{fig:effects_seed}
    \item Distributions on VQA v2: \autoref{fig:effects_vqav2}
\end{itemize}

% LLaVa-Bench

% MMBench

% SEED-Bench-2

% VQA v2

\end{document}